\documentclass[sigconf]{acmart}

\AtBeginDocument{%
  \providecommand\BibTeX{{%
    \normalfont B\kern-0.5em{\scshape i\kern-0.25em b}\kern-0.8em\TeX}}}

\acmConference[Technique Report]{MobiHoc '21: ACM International Symposium on Theory, Algorithmic Foundations, and Protocol Design for Mobile Networks and Mobile Computing}{June}{2021}
\acmBooktitle{.}
\acmPrice{15.00}
\acmISBN{978-1-4503-XXXX-X/18/06}



\usepackage{algorithmic}
\usepackage{graphicx}
\usepackage{textcomp}

\usepackage{url}            
\usepackage{booktabs}       
\usepackage{amsfonts}       
\usepackage{nicefrac}       
\usepackage{microtype}      

\usepackage{algorithm}
\usepackage{algorithmic}
\usepackage{graphics} 
\usepackage{epsfig} 

\usepackage{enumitem}
\usepackage{multirow}
\usepackage{array}
\usepackage{bbding}
\usepackage{threeparttable}
\usepackage{thmtools}
\usepackage{thm-restate}

\begin{document}

\title{CFedAvg: Achieving Efficient Communication and Fast Convergence in Non-IID Federated Learning}


\author{Haibo Yang}
\affiliation{%
  \institution{Department of ECE \\
  The Ohio State University}
  \streetaddress{2015 Neil Ave}
  \city{Columbus, OH}
  \country{U.S.A}}
\email{yang.5952@osu.edu}

\author{Jia Liu}
\affiliation{%
  \institution{Department of ECE \\
  The Ohio State University}
  \streetaddress{2015 Neil Ave}
  \city{Columbus, OH}
  \country{U.S.A}}
\email{liu@ece.osu.edu}

\author{Elizabeth S. Bentley}
\affiliation{%
  \institution{Information Directorate \\
  Air Force Research Laboratory}
  \streetaddress{2015 Neil Ave}
  \city{Rome, NY}
  \country{U.S.A}}
\email{elizabeth.bentley.3@us.af.mil}



\newcommand{\sign}{\mathbf{sign}}

\newcommand{\A}{\mathbf{A}}
\renewcommand{\a}{\mathbf{a}}
\newcommand{\B}{\mathbf{B}}
\newcommand{\Bt}{\mathbf{B}}
\renewcommand{\b}{\mathbf{b}}
\newcommand{\bt}{\mathbf{b}}
\renewcommand{\c}{\mathbf{c}}
\newcommand{\ct}{\widetilde{\mathbf{c}}}
\newcommand{\Ch}{\widehat{\mathbf{C}}}
\newcommand{\D}{\mathbf{D}}
\newcommand{\Dh}{\widehat{\mathbf{D}}}
\renewcommand{\d}{\mathbf{d}}
\newcommand{\E}{\mathbf{E}}
\newcommand{\e}{\mathbf{e}}
\newcommand{\et}{\widetilde{\mathbf{e}}}
\newcommand{\f}{\mathbf{f}}
\newcommand{\F}{\mathbf{F}}
\newcommand{\g}{\mathbf{g}}
\newcommand{\I}{\mathbf{I}}
\newcommand{\J}{\mathbf{J}}
\newcommand{\K}{\mathbf{K}}
\newcommand{\M}{\mathbf{M}}
\newcommand{\m}{\mathbf{m}}
\newcommand{\mh}{\widehat{\mathbf{m}}}
\newcommand{\Mh}{\widehat{\mathbf{M}}}
\newcommand{\Mt}{\widetilde{\mathbf{M}}}
\newcommand{\Mb}{\overline{\mathbf{M}}}
\newcommand{\N}{\mathbf{N}}
\newcommand{\n}{\mathbf{n}}
\newcommand{\p}{\mathbf{p}}
\renewcommand{\P}{\mathbf{P}}
\newcommand{\Q}{\mathbf{Q}}
\newcommand{\q}{\mathbf{q}}
\newcommand{\R}{\mathbf{R}}
\renewcommand{\r}{\mathbf{r}}
\renewcommand{\S}{\mathbf{S}}
\newcommand{\Sh}{\widehat{\mathbf{S}}}
\newcommand{\s}{\mathbf{s}}
\renewcommand{\t}{\mathbf{t}}
\newcommand{\T}{\mathbf{T}}
\newcommand{\Th}{\widehat{\mathbf{T}}}
\renewcommand{\u}{\mathbf{u}}
\newcommand{\V}{\mathbf{V}}
\renewcommand{\v}{\mathbf{v}}
\newcommand{\W}{\mathbf{W}}
\newcommand{\w}{\mathbf{w}}
\newcommand{\wt}{\widetilde{\mathbf{w}}}
\newcommand{\wts}{\widetilde{w}}
\newcommand{\X}{\mathbf{X}}
\newcommand{\Xh}{\widehat{\mathbf{X}}}
\newcommand{\x}{\mathbf{x}}
\newcommand{\xt}{\widetilde{\mathbf{x}}}
\newcommand{\xh}{\widehat{\mathbf{x}}}
\newcommand{\Y}{\mathbf{Y}}
\newcommand{\y}{\mathbf{y}}
\newcommand{\z}{\mathbf{z}}
\newcommand{\Z}{\mathbf{Z}}
\newcommand{\yt}{\widetilde{\mathbf{y}}}
\newcommand{\rb}{\mathbf{b}}
\newcommand{\rvec}{\mathbf{r}}
\newcommand{\0}{\mathbf{0}}
\newcommand{\1}{\mathbf{1}}
\newcommand{\Hc}{\mathbf{H}}
\newcommand{\tHc}{\widetilde{\mathbf{H}}}
\newcommand{\hHc}{\overline{\mathbf{H}}}
\newcommand{\DL}{\mathbf{DL}}
\newcommand{\mLambda}{\mathbf{\Lambda}}
\newcommand{\mOmega}{\mathbf{\Omega}}
\newcommand{\mOmegab}{\overline{\mathbf{\Omega}}}
\newcommand{\mGamma}{\mathbf{\Gamma}}
\newcommand{\mPi}{\mathbf{\Pi}}
\newcommand{\mPsi}{\mathbf{\Psi}}
\newcommand{\mPsib}{\overline{\mathbf{\Psi}}}
\newcommand{\valpha}{\mathbf{\alpha}}
\newcommand{\vpi}{\boldsymbol \pi}
\newcommand{\Sone}{\mathcal{S}}
\newcommand{\tr}{\mathrm{Tr}}
\newcommand{\new}{\mathrm{new}}
\newcommand{\net}{\mathrm{net}}
\newcommand{\link}{\mathrm{link-phy}}
\newcommand{\LL}{\mathbf{L}}
\newcommand{\Out}[1]{\mathcal{O}\left( #1 \right)}
\newcommand{\In}[1]{\mathcal{I}\left( #1 \right)}
\newcommand{\Int}[1]{\mathcal{F}\left( #1 \right)}
\newcommand{\sInt}[1]{\hat{\mathcal{F}}\left( #1 \right)}
\renewcommand{\Re}[1]{\mbox{$\mathfrak{Re} \left( #1 \right)$}}
\renewcommand{\Im}[1]{\mbox{$\mathfrak{Im} \left( #1 \right)$}}
\newcommand{\Resq}[1]{\mbox{$\mathfrak{Re}^{2} \left( #1 \right)$}}
\newcommand{\Imsq}[1]{\mbox{$\mathfrak{Im}^{2} \left( #1 \right)$}}
\newcommand{\LTM}{\langle L \rangle}
\newcommand{\con}[1]{\overline{#1}}
\newcommand{\abs}[1]{\mbox{$\lvert #1 \rvert$}}
\newcommand{\norm}[1]{\mbox{$\left\lVert #1 \right\rVert$}}
\newcommand{\dual}{$\mathbf{D}^{\mathrm{CRPBA}}$}
\newcommand{\dnet}{$\mathbf{D}^{\mathrm{CRPBA}}_{\net}$}
\newcommand{\dlink}{$\mathbf{D}^{\mathrm{CRPBA}}_{\link}$}
\newcommand{\Pwrcon}{\Omega_{+}(n)}
\newcommand{\src}[1]{\mathrm{Src}(#1)}
\newcommand{\dst}[1]{\mathrm{Dst}(#1)}
\newcommand{\rx}[1]{\mathrm{Rx}(#1)}
\newcommand{\tx}[1]{\mathrm{Tx}(#1)}
\newcommand{\Dy}{\Delta \y}
\newcommand{\Dyt}{\Delta \widetilde{\y}}

\newtheorem{thm}{Theorem}
\newtheorem{cor}[thm]{Corollary}
\newtheorem{lem}{Lemma}
\newtheorem{claim}{Claim}
\newtheorem{prop}[thm]{Proposition}
\newtheorem{ex}{Example}
\newtheorem{defn}{Definition}
\newtheorem{finalremark}[thm]{Final Remark}
\newtheorem{rem}{Remark}
\newtheorem{sol}{Solution}
\newtheorem{assum}{Assumption}

\newcommand{\po}{\mbox{$ \mbox{P}_{\rm out}$}}
\newcommand{\ga}{\bar{\gamma}}
\newcommand{\mtrx}[1]{\mbox{$\left[\begin{array} #1 \end{array}\right]$}}
\newcommand{\mtrxt}[1]{\mbox{$\begin{array} #1 \end{array}$}}
\newcommand{\diag}[1]{\mathrm{Diag}\left\{ #1 \right\}}
\newcommand{\diaginv}[1]{\mathrm{Diag}^{-1}\left\{ #1 \right\}}
\newcommand{\diagtran}[1]{\mathrm{Diag}^{T}\left\{ #1 \right\}}
\newcommand{\diagv}[1]{\mathrm{diag}\left\{ #1 \right\}}
\newcommand{\etal}{{\em et al. }}
\newcommand{\ind}{\mathbbm{1}}
\newcommand{\vbeta}{\boldsymbol \beta}

\begin{abstract}
Federated learning (FL) is a prevailing distributed learning paradigm, where a large number of workers jointly learn a model without sharing their training data. However, high communication costs could arise in FL due to large-scale (deep) learning models and bandwidth-constrained connections. In this paper, we introduce a communication-efficient algorithmic framework called CFedAvg for FL with non-i.i.d. datasets, which works with general (biased or unbiased) SNR-constrained compressors. We analyze the convergence rate of CFedAvg for non-convex functions with constant and decaying learning rates. The CFedAvg algorithm can achieve an $\mathcal{O}(1 / \sqrt{mKT} + 1 / T)$ convergence rate with a constant learning rate, implying a linear speedup for convergence as the number of workers increases, where $K$ is the number of local steps, $T$ is the number of total communication rounds, and $m$ is the total worker number. This matches the convergence rate of distributed/federated learning without compression, thus achieving high communication efficiency while not sacrificing learning accuracy in FL. 
Furthermore, we extend CFedAvg to cases with heterogeneous local steps, which allows different workers to perform a different number of local steps to better adapt to their own circumstances.
The interesting observation in general is that the noise/variance introduced by compressors does not affect the overall convergence rate order for non-i.i.d. FL. We verify the effectiveness of our CFedAvg algorithm on three datasets with two gradient compression schemes of different compression ratios.
\end{abstract}



\maketitle


\section{Introduction} \label{sec: intro}

In recent years, advances in machine learning (ML) have sparked many new and emerging applications that transform our society, which include a collection of widely used deep learning models for computer vision, text prediction, and many others \cite{shi2016edge,li2018learning}.
Traditionally, the training of ML applications often relies on cloud-based large data-centers to collect and process a vast amount of data.
However, with the rise of Internet of Things (IoT), data and demands for ML are increasingly being generated from mobile devices in wireless edge networks. 
Due to high latency, low bandwidth, and privacy/security concerns, aggregating all data to the cloud for ML training may no longer be desirable or may even be infeasible. 
In these circumstances, Federated Learning (FL) has emerged as a prevailing ML paradigm, thanks to the rapidly growing computation capability of modern mobile devices.
Generally speaking, FL is a network-based ML architecture, under which a large number of local devices (often referred to as workers/nodes) collaboratively train a model based on their local datasets and coordinated by a central parameter server.
In FL, the parameter server is responsible for aggregating and updating model parameters without requiring the knowledge of the data located at each worker.
Workers process their computational tasks independently with decentralized data, and communicate to the parameter server to update the model.
By doing so, not only can FL significantly alleviate the risk of exposing data privacy, it also fully utilizes the idle computation resources at the workers.
This constitutes a win-win situation that have led to a variety of successful real-world applications (see \cite{kairouz2019advances} for a comprehensive survey). 

Despite the aforementioned advantages of FL, a number of technical challenges also arise due to the unique characteristics of FL.
One key challenge in FL is the high communication cost due to the every-increasing sizes of learning models and datasets~\cite{kairouz2019advances}.
For example, modern models in deep learning (e.g., ResNet~\cite{he2016deep}, VGG~\cite{simonyan2014very}, etc.) typically contain millions of parameters, which implies a large amount of data being injected into the network that supports FL.
The problem of a high communication load in FL is further exacerbated by the fact that, in many wireless edge networks, the communication links are often bandwidth-constrained and their link capacities are highly dynamic due to stochastic channel fading effects.
As a result, information exchanges between the parameter server and workers could be highly inefficient, rendering a major bottleneck in FL \cite{kairouz2019advances}.
If this problem is not handled appropriately, FL could perform far worse than its centralized counterparts in scenarios where the communication-to-computation cost ratio is high and/or communication resources are constrained.
Given the rapidly growing demands of FL and the restricted communication environments in reality, there is a compelling need to significantly reduce the communication cost in FL without substantially decaying the learning performance.

Generally speaking, the total communication cost during the training process is determined by two factors: the number of communication rounds and the size (or amount) of the update parameters in each communication round.
There are some algorithms in FL (e.g., federated averaging (FedAvg) \cite{mcmahan2016communication}) that take more local gradient steps at each node and communicate infrequently with the parameter server, thus decreasing the total number of communication rounds and in turn reducing the total communication cost.
However, this does not completely solve the problem since the communicated parameter/gradient vectors could still be high-dimensional.
On the other hand, in the literature, there exist a variety of gradient compression techniques (e.g., signSGD \cite{bernstein2018signsgd}, gradient dropping \cite{aji2017sparse}, TernGrad \cite{wen2017terngrad}, etc.) that were originally proposed for centralized/distributed learning and shown to be effective in reducing the size of exchanged parameters in each communication round.
For example, Lin {\em et al.} \cite{lin2017deep} numerically demonstrated that $99.9\%$ of the gradient exchange in distributed learning are redundant and proposed Deep Gradient Compression (DGC) to largely reduce the communication bandwidth requirement.
Similar compression ideas have also been extended to decentralized learning over networks {\em without} dedicated parameter servers.
In \cite{zhang2020private,zhang2020communication}, Zhang {\em et al.} developed a series of decentralized learning algorithms with differential-coded compressions.
Koloskova {\em et al.} \cite{koloskova2019decentralized} proposed an algorithm that could achieve a linear speedup with respect to the number of workers for convergence in decentralized learning with arbitrary gradient compression.


Given the above encouraging results of information compression in distributed/decentralized learning, an interesting question naturally arises: 
{\em Could we combine compression with infrequent communication to further reduce the communication cost of FL?}
However, answering this question turns out to be highly non-trivial.
One key challenge stems from the heterogeneity of the local datasets among different workers.
In the traditional distributed learning literature, the dataset at each node is usually well-shuffled and hence can be assumed to be independent and identically distributed (i.i.d.). 
However, the dataset at each worker in FL could be generated based on the local environment and cannot be shuffled with other workers due to privacy protection.
Thus, the i.i.d. assumption often fails to hold.
In some circumstances, the dataset distributions at different workers could vary dramatically due to factors such as geographic location differences, time window gaps, among many others.
Upon integrating compression with FL, the already-complicated non-i.i.d. dataset problem is further worsened by the significant loss of information due to the use of compression operators.
In addition, under infrequent communication, 
the multiple local steps in each worker introduce further ``model drift'' to the non-i.i.d. datasets.
It has been shown that this ``model drift'' results in extra variances that may lead to deterioration or even failures of training \cite{li2018federated}.
Due to these complex randomness couplings between compression, local steps, and non-i.i.d. datasets, results on non-i.i.d. compressed FL remain limited.
This motivates us to fill this gap and rigorously investigate the algorithmic design that integrates compression in non-i.i.d. FL.

Moreover, workers in FL system vary tremendously in terms of computation capabilities and resources limits (e.g., memory, battery capacity).
Hence, using a predefined constant number of local steps for all workers (assumed in most existing work in FL) may not be a good design strategy, which may result in the faster workers idling and slower workers causing straggler problems.
As a result, another important question in FL emerges: 
{\em Could we use heterogeneous local steps for workers to further improve flexibility and efficiency in FL?}

In this paper, we answer the above open questions by proposing a communication-efficient algorithm called {\em CFedAvg} (compressed FedAvg) with error-feedback.
Our CFedAvg algorithm reduces both the communication rounds and link capacity requirement in each communication round.
It also allows the use of heterogeneous local steps (i.e., different workers perform different local steps to better adapt to their own computing environments). 
Our main contributions and results are summarized as follows:

\begin{list}{\labelitemi}{\leftmargin=1em \itemindent=0em \itemsep=.2em}
\item 
We show that, under general signal-to-noise-ratio (SNR) constrained compressors, the convergence rate of our CFedAvg algorithm is $\mathcal{O}(1 / \sqrt{mKT} + 1 / T)$ and $\mathcal{\tilde{O}}(1 / \sqrt{mKT}) + \mathcal{O}(1 / \sqrt{T})$ with constant and decaying learning rates, respectively, for general non-convex functions and non-i.i.d. datasets in FL, where $K$ is the number of local steps, $T$ is the number of total communication rounds, and $m$ is the total number of workers.
For a sufficiently large $T$, this implies that CFedAvg achieves an $\mathcal{O}(1 / \sqrt{mKT})$ convergence rate with a constant learning rate and enjoys the linear speedup effect as the number of workers increases.\footnote{To attain an $\epsilon$ accuracy for an algorithm, it takes $\mathcal{O}(1/\epsilon^2)$ steps with a convergence rate $\mathcal{O}(1/\sqrt{T})$.
In contrast, it takes $\mathcal{O}(1/m \epsilon^2)$ steps if the convergence rate is $\mathcal{O}(1/\sqrt{mT})$ (the hidden constant in Big-O is the same). In this sense, it is a {\em linear speedup} with respect to the number of workers $m$.} 
Note that this matches the convergence rate of uncompressed distributed/federated learning algorithm orderly \cite{dekel2012optimal,karimireddy2019scaffold,yang2021achieving}, thus achieving high communication efficiency while not sacrificing learning accuracy in FL.

\item
We extend CFedAvg to heterogeneous local steps among workers, which allows each worker performs different local steps based on its own computation capability and other conditions.
To our knowledge, our paper is the first to show that FL can still offer theoretical performance guarantee without requiring the same predefined constant number of local steps among all workers.

\item
We show that the use of SNR-constrained compressors in CFedAvg only slightly increases the local variance constant and does not affect the overall convergence rate order for non-i.i.d. FL with infrequent communication. 
Moreover, we show that the convergence results of CFedAvg hold for either unbiased or biased SNR-constrained compressors, which is far more flexible than previous works that require unbiased compressors.


\item
We verify the effectiveness of CFedAvg on MNIST, FMNIST and CIFAR-10 datasets with different SNR-constrained compressors.
We find that CFedAvg can reduce up to $99\%$ of information exchange with minimal impacts on learning accuracy, which confirms the communication-efficiency advantages of training large learning models in non-i.i.d. FL with information compression.
\end{list}

The rest of the paper is organized as follows.
In Section~\ref{sec:related_work}, we review the literature to put our work in comparative perspectives.
In Section~\ref{sec:model}, we introduce the system model, problem formulation, and preliminaries on general SNR-constrained compressors.
In Section~\ref{sec:algorithm}, we present the details of our algorithm design and the convergence analysis.
Numerical results are provided in Section~\ref{sec:exp} and Section~\ref{sec:con} concludes this paper.


\section{Related work} \label{sec:related_work}

For distributed/federated learning in communication-constrained environments, a variety of communication-efficient algorithms have been proposed.
We categorize the existing work into two classes: one is to use infrequent communication to reduce the communication rounds and the other is to compress the information to reduce the size of parameters transmitted from each worker to the parameter server in each communication round. 

\smallskip
\textbf{Infrequent-Communication Approaches:}
One notable algorithm of FL is the federated averaging (FedAvg) algorithm, which was first proposed by McMahan {\em et al.} \cite{mcmahan2016communication} as a heuristic to improve both communication efficiency and data privacy.
In FedAvg, every worker performs multiple SGD steps independently to update the model locally before communicating with the parameter server, which is different from traditional distributed learning with only one local step.
It has been shown that the number of local steps can be up to $100$ without significantly affecting the convergence speed in i.i.d. datasets for various convolution and recurrent neural network models.
Since then, this work has sparked many follow-ups that focus on FL with i.i.d. datasets (referred to as LocalSGD) \cite{stich2018local,yu2019parallel,wang2018cooperative,stich2019error,lin2018don,khaled2019better,zhou2017convergence} and non-i.i.d. datasets \cite{sattler2019robust,zhao2018federated,li2018federated,wang2019slowmo,karimireddy2019scaffold,huang2018loadaboost,jeong2018communication,lin2018don,yang2021achieving}.
These studies heuristically demonstrated the effectiveness of FedAvg and its variants  on reducing communication cost.
Also, researchers have theoretically shown that FedAvg and its variants can achieve the same convergence rate order as the traditional distributed learning (see, e.g., \cite{karimireddy2019scaffold,li2019convergence,kairouz2019advances,yang2021achieving}).

\smallskip
\textbf{Compression-Based Approaches:}
Although FedAvg and its variants save communication costs by utilizing multiple local steps to reduce the total number of communication rounds, it has to transmit the full amount of model parameters in each communication round at every worker.
Thus, it could still induce high latency and communication overhead in networks with low connection speeds or large channel variations.
To address this challenge, a natural idea is to compress the parameters to reduce the amount of transmitted data from each worker to the parameter server.
Compression-based approaches have attracted increasing attention in recent years in distributed and decentralized learning \cite{lin2017deep,koloskova2019decentralized}, which have enabled the training of large-size models over networks with low-speed connections.
Broadly speaking, compression-based approaches can be classified into the following two main categories:
\begin{list}{\labelitemi}{\leftmargin=1em \itemindent=0em \itemsep=.2em}
\item \textbf{Quantization:}
The basic idea of quantization is to project a vector from a high-dimensional space to a low-dimensional subspace, so that the projected vector can be represented by a fewer number of bits.
Notable examples of quantization-based algorithms in the learning literature include, e.g., signSGD \cite{bernstein2018signsgd,yang2020adaptive}, QSGD \cite{alistarh2017qsgd}.
Note that these commonly seen quantization schemes could be either unbiased \cite{alistarh2017qsgd,wen2017terngrad} or biased \cite{bernstein2018signsgd}.


\item \textbf{Sparsification:}
Given a high-dimensional vector, the basic idea of sparsification is to select only a part of its components to transmit.
The component selection could be based on a predefined threshold.
For example, Strom \cite{strom2015scalable} proposed to only send components in the  vector that are larger than a predefined constant, while Aji {\em et al.} \cite{aji2017sparse} chose to send a fixed proportion of components.
The component selection could also be randomized \cite{wangni2018gradient,aji2017sparse}.
Other variants include, e.g., adaptive threshold \cite{dryden2016communication} and unbiased random dropping \cite{wangni2018gradient}.
\end{list}

In fact, these two approaches are closely related, and there are works that combine them to achieve better compression results \cite{sattler2019robust,alistarh2017qsgd,wen2017terngrad}.
Qsparse-local-SGD proposed by Basu {\em et al.} ~\cite{basu2020qsparse} is the most related work to this paper, which combines quantization, sparsification and local steps to be more communication-efficient.
However, our algorithm has a better convergence rate with more relaxed assumptions. We also propose an algorithm design that allows more flexible heterogeneous local steps. Please see Section~\ref{sec:main_results} for further details.

\smallskip
{\bf Error Feedback:} 
With the information loss of model parameters transmitted from the workers to server due to gradient compression, training accuracy 
could be significantly affected.
To address this problem, the error feedback technique has been proposed~\cite{stich2019error,karimireddy2019error,lin2017deep,tang2019texttt} for both distributed and decentralized learning.
It has been shown that error feedback improves the convergence performance for cases with high compression ratios.
It has also been theoretically shown that distributed and decentralized learning with gradient compression and error feedback could achieve the same convergence rate as that of the classical distributed SGD \cite{tang2019texttt,stich2019error,karimireddy2019error,koloskova2019decentralized} and enjoy the linear speedup effect.
 
 So far, however, it remains unknown whether the same convergence rate (with linear speedup) could be achieved in FL with compression, particularly under the asynchrony due to non-i.i.d. dataset and the use of heterogeneous local steps at each worker.
Answering this question constitutes the rest of this paper.
 


\section{System Model, Problem Formulation, and Preliminaries} \label{sec:model}



\subsection{System Model and Problem Formulation} \label{sec:FL}
Consider an FL system with $m$ workers who collaboratively learn a model with decentralized data and under the coordination of a central parameter server.
The goal of the FL system is to solve the following optimization problem:
\begin{align} \label{eqn:formulation}
\min_{\x \in \mathbb{R}^d} f(\x) := \frac{1}{m} \sum_{i=1}^{m} F_i(\x),
\end{align}
where $F_i(\x) \triangleq \mathbb{E}_{\xi_i \sim D_i}[F_i(\x, \xi_i)]$ denotes the local (non-convex) loss function, which evaluates the average discrepancy between the learning model's output and the ground truth corresponding to a random training sample $\xi_{i}$ that follows a local data distribution $D_i$.
In \eqref{eqn:formulation}, the parameter $d$ represents the dimensionality of the training model.
For the i.i.d. setting, each local dataset is assumed to sample from some common latent distribution, i.e., $D_i = D, \forall i \in [m]$.
In practice, however, the local dataset at each worker in FL could be generated based on its local environment and thus being non-i.i.d., i.e., $D_i \neq D_j$ if $i \neq j$.
Note that the i.i.d. setting can be viewed as a special case of the non-i.i.d. setting.
Hence, our results for the non-i.i.d. setting are directly applicable to the i.i.d. setting.

\subsection{General SNR-Constrained Compressors} \label{sec:compressor}

To facilitate the discussions of our CFedAvg algorithm, 
we will first formally define the notion of {\em general SNR (signal-to-noise ratio)-constrained compressors}, 
which has been used in the literature (e.g., \cite{karimireddy2019error,stich2018sparsified}):

\begin{defn} (General SNR-Constrained Compressor) \label{def_compression}
An operator $\mathcal{C}(\cdot) : \mathbb{R}^d \rightarrow \mathbb{R}^d$ is said to be constrained by an SNR threshold $\gamma \geq 1$ if it satisfies:
\begin{align*}
\mathbb{E_{\mathcal{C}}} \| \mathcal{C}(\x) - \x \|^2 \leq (1/\gamma) \| \x \|^2, \quad \forall \x \in \mathbb{R}^d.
\end{align*}
\end{defn}

It is clear from Definition~\ref{def_compression} that, for a given compressor, $\gamma$ is its lowest SNR guarantee yielded by its largest compression noise
power $\| \mathcal{C}(\x) - \x \|^2$.
The $\gamma$-threshold can be viewed as a proxy of compression rate of the compressor.
For $\gamma = \infty$, we have $\mathcal{C}(\x) = \x$, which means no compression and zero information loss.
On the other hand, $\gamma \rightarrow 1$ implies that the compression rate is arbitrarily high and the output contains no information of $\x$.
It is worth pointing out that, in Definition~\ref{def_compression}, the SNR-constrained compressor is not assumed to be unbiased, hence the term ``general''.
Definition~\ref{def_compression} covers a large class of compression schemes, e.g., the Top-$k$ compressor \cite{aji2017sparse,lin2017deep} that selects $k$ coordinates with the largest absolute values, and the random sparsifier~\cite{wangni2018gradient} that randomly selects components. 

\section{Compressed FedAvg (CFedAvg) for Non-IID Federated Learning} \label{sec:algorithm}

In this section, we will first introduce our CFedAvg (compressed FedAvg) algorithm in Section~\ref{sec:cfedavg}.
Then, we will present the main theoretical result and their key insights/interpretations in Section~\ref{sec:main_results}.
Due to space limitation, we provide proof sketches for the main results in Section~\ref{sec:proofs}
 and relegate the full proofs of all theoretical results in the appendix.

\subsection{The CFedAvg Algorithmic Framework} \label{sec:cfedavg} 
The general CFedAvg algorithmic framework is stated in Algorithm~\ref{alg:FedAvg_compression}.
We aim to not only reduce total communication rounds, but also compress the gradients transmitted in each communication round.
The algorithm contains four key stages:
\begin{list}{\labelitemi}{\leftmargin=0.1em \itemindent=0em \itemsep=.2em}
\item[1.] {\em Local Computation:}
Line~6 says that, in each communication round, each worker runs $K_i$ local updates before communicating with the server.
As shown in Line~7, each local update step takes an unbiased gradient estimator (e.g., vanilla SGD).
A local learning rate $\eta_{L,t}$ is adopted for each local step (Line~8).

\item[2.] {\em Gradient Compression:}
We compress the model changes $\g_t^i$ instead of the last model $\x_{t, K_i}^i$ in each worker, where $\g_t^i = \x_{t, K}^i - \x_t$ for homogeneous local step ($K_i = K, \forall i \in [m]$) in Line~10 or $\g_t^i = \frac{1}{K_i} (\x_{t, K_i}^i - \x_t)$ for heterogeneous local step (different local steps $K_i, \forall i \in [m]$) in Line~11.
Before compressing the parameters, we add the error term to compensate the parameter in each worker in Line~12, i.e., $\p_t^i = \g_t^i + \e_t^i, \forall i \in [m]$.
Then, we compress the parameter $\p_t^i$ and send the result $\tilde{\Delta}_t^i = \mathcal{C}(\p_t^i)$ to the server, where $\mathcal{C}(\cdot)$ denotes a general SNR-constrained compressor.

\item[3.] {\em Error Feedback:}
We update error term after gradient compression in each communication round in Line~15,
representing the information loss due to compression.
This would be used later to compensate the parameters in the next communication round to ensure not too much parameter information is lost.

\item[4.] {\em Global Update:}
Upon the reception of all returned parameters, the server updates the parameters using a global learning rate $\eta$ and broadcasts the new model parameters to all workers.
\end{list}



\begin{algorithm}[ht!]
\caption{The General CFedAvg Algorithmic Framework.} \label{alg:FedAvg_compression} 
\begin{algorithmic}[1]
\STATE 
Initialize $\x_0$. 
\FOR{$t = 0, \cdots, T - 1$} 
\STATE {Initialize $\e_0^i=0, i \in [m]$ if $t = 0$.}
	\FOR{each worker $i \in [m]$ in parallel} 
		\STATE {$\x_{t, 0}^i = \x_t$}
		\FOR{$k = 0, \cdots, K_i - 1$}
			\STATE {Compute an unbiased stochastic gradient estimate $\g_{t, k}^{i} = \nabla F_i(\x_{t, k}^i, \xi_{t, k}^i)$ of $\nabla F_i(\x_{t, k}^i)$}.
			\STATE {Local update: $\x_{t, k+1}^i = \x_{t, k}^i - \eta_{L,t} \g_{t, k}^{i}$}.
		\ENDFOR \\
		\STATE For Homogeneous Local Steps: $\g_t^i = \x_{t, K}^i - \x_t$. \\
		\STATE For Heterogeneous Local Steps: $\g_t^i = \frac{1}{K_i} (\x_{t, K_i}^i - \x_t)$. \\
		\STATE $\p_t^i = \g_t^i + \e_t^i$. \\
		\STATE $\tilde{\Delta}_t^i = \mathcal{C}(\p_t^i)$ ($\mathcal{C}(\cdot)$ is an SNR-constrained compressor). \\
		\STATE Send $\tilde{\Delta}_t^i$ to server. \\
		\STATE $\e_{t+1}^i = \p_t^i - \tilde{\Delta}_t^i$.
	\ENDFOR \\
	{\bf At Parameter Server:} 
		\STATE \quad {Receive $\tilde{\Delta}_t^i, i \in [m]$.} \\
		\STATE \quad $\tilde{\Delta}_t = \frac{1}{m} \sum_{i \in [m]} \tilde{\Delta}_t^i$. \\
		\STATE \quad {Server Update: $\x_{t+1} = \x_t + \eta \tilde{\Delta}_t$}.
		\STATE \quad {Broadcast $\x_{t+1}$ to each worker}.
\ENDFOR
\end{algorithmic}
\end{algorithm}
\setlength{\textfloatsep}{0pt}

\subsection{Main Theoretical Results} \label{sec:main_results}
In this subsection, we will establish the convergence results of our proposed CFedAvg algorithmic framework. 
Our convergence results are proved under the following mild assumptions:

\begin{assum}(Lipschitz Smoothness) \label{a_smooth}
	There exists a constant $L \!>\! 0$, s.t. $ \| \nabla F_i(\x) \!\!-\!\! \nabla F_i(\y) \| \!\!\leq\! L \| \x \!-\! \y \|$, $\forall \x, \y \in \mathbb{R}^d, \forall i \in [m]$.
\end{assum}

\begin{assum}(Unbiased Gradient Estimator) \label{a_unbias}
	Let $\xi_t^i$ be a random local sample in the $t$-th round at worker $i$.
	The gradient estimator is unbiased, i.e.,
	$\mathbb{E} [\nabla F_i(\x, \xi^i)] = \nabla F_i(\x)$, $\forall i \in [m].$
\end{assum}

\begin{assum}(Bounded Local/Global Variance) \label{a_variance}
	There exist two constants $\sigma_L \!\!\geq\!\! 0$ and $\sigma_G \!\!\geq\!\! 0$, such that  the variance of each local gradient estimator is bounded by
	$\mathbb{E} [|| \nabla F_i(\x, \xi_t^i) -  \nabla F_i(\x) ||^2] \leq \sigma_{L}^2$,
	and the global variability of the local gradient is bounded by
	$\| \nabla F_i(\x) - \nabla f(\x) \|^2 \leq \sigma_{G}^2$, $\forall i \in [m]$.
\end{assum}
The first two assumptions and the bounded local variance assumption in Assumption~\ref{a_variance} are standard assumptions in the convergence analysis of stochastic gradient-type algorithms in non-convex optimization (e.g., \cite{kairouz2019advances}
).
We use a universal bound $\sigma_G$ to quantify the heterogeneity of the {\em non-i.i.d. datasets} among different workers.
This assumption has also been used in other works for FL with non-i.i.d. datasets \cite{reddi2020adaptive} as well as in decentralized optimization \cite{kairouz2019advances}.
It is worth noting that we do {\em not} require a bounded gradient assumption, which is often used in FL optimization analysis \cite{kairouz2019advances}.
With the above assumptions, we are now in a position to present our main theoretical results.
First, we state a useful result in Lemma~\ref{lem: Bounded_error}:
\begin{restatable} {lemma} {lemmaError}
(Bounded Error). \label{lem: Bounded_error}
For any local learning rate satisfying $\eta_{L,t} \!\!\!\leq\!\!\! \frac{1}{8LK}$, the error term can be upper bounded by $\sum_{t=0}^{T-1} \| \e_t \|^2  \!\!\!\leq\!\!\!  \sum_{t=0}^{T-1} h(\gamma, \alpha_t) \| \Delta_{t} \|^2$, where $\e_t \!=\! \frac{1}{m} \sum_{i \in [m]} \e_t^i, \Delta_t \!=\! \frac{1}{m} \sum_{i \in [m]} \g_t^i$,
$h(\gamma, \alpha_t) = (1/ \gamma) (1 + 1/a) b \alpha_t, \alpha_t = \frac{ \mathbb{E} \| \bar{\Delta}_t \|^2}{\mathbb{E} \| \Delta_t \|^2}$
and $a$ and $b$ are constants such that $\gamma \epsilon - 1 \geq a$ and $\frac{1}{1 - \epsilon} \leq b$ for $\epsilon \in (0, 1)$.
\end{restatable}

Lemma~\ref{lem: Bounded_error} implies that the error term cannot grow arbitrarily large with a proper SNR threshold $\gamma$ (determined by compression rate) for which $h(\gamma, \alpha_t)$ does not go to infinity.
The error is upper bounded by the accumulated parameters $\Delta_{t}$.
This suggests that the total information loss due to compression is only a fraction of the total of the accumulated parameters.

\smallskip
{\bf 1)~CFedAvg with Constant Learning Rates (Homogeneous Local Steps):} 
As a first step, we consider a simpler case where CFedAvg uses constant learning rates (i.e., $\eta_{L,t} \equiv \eta_{L}$, $\forall t$) and homogeneous local steps (i.e., $K_i \equiv K$, $\forall i$).
In this case, by Lemma~\ref{lem: Bounded_error}, we can establish convergence result for CFedAvg as follows:
\begin{restatable} {theorem}{FedAvg}
(Convergence Rate of CFedAvg). \label{thm:FedAvg_compression}
Choose constant local and global learning rates $\eta_L$ and $\eta$ s.t. $\eta_L \leq \frac{1}{8LK}$, $\eta \eta_L < \frac{1}{K L}$ and $ \eta \eta_L K ( L^2 h(\gamma, \alpha_t) + 1 + L) \leq 1, \forall t \in [T]$.
Under Assumptions~\ref{a_smooth}--\ref{a_variance}, the sequence  $\{ \x_t \}$ generated by Algorithm~\ref{alg:FedAvg_compression} satisfies:
\begin{align} \label{eqn:convg_bound}
\min_{t \in [T]} \mathbb{E}\|\nabla f(\x_t)\|_2^2 \leq \frac{f_0 - f_{*}}{c \eta \eta_L K T} + \Phi,
\end{align}
where $\Phi \triangleq \frac{1}{c} [(\frac{1}{2} L^2 h(\gamma, \alpha) + \frac{1}{2} + \frac{L}{2}) \frac{\eta \eta_L}{m} \sigma_L^2 + \frac{5 K \eta_L^2 L^2}{2} (\sigma_L^2 + 6 K \sigma_G^2)]$, $c$ is a constant, $h(\gamma, \alpha)$ is defined by $h(\gamma, \alpha) = \frac{1}{T} \sum_{t=0}^{T-1} h(\gamma, \alpha_t)$, $h(\gamma, \alpha_t)$ is defined the same as that in Lemma~\ref{lem: Bounded_error}, $f_0 = f(\hat{\x}_{0})$ and $f_{*}$ is the optimal.
\end{restatable}

\begin{rem} [Decomposition of The Bound]
{\em 
The bound of the convergence rate in (\ref{eqn:convg_bound}) contains two terms: the first term is a vanishing term $\frac{f_0 - f_{*}}{c \eta \eta_L K T}$ as $T$ increases, and the second term is a constant $\Phi$ independent of $T$.
Note that $\Phi$ depends on three factors: local variance $\sigma_L$, global variance $\sigma_G$, and the number of local steps $K$.
We can further decompose the constant term $\Phi$ into two parts.
The first part of $\Phi$ is due to the local variance of the stochastic gradient in each local SGD step for each worker.
It shrinks at the rate $\frac{1}{m}$ with respect to the number of workers $m$, which favors large distributed systems.
This makes intuitive sense since more workers means more training samples in one communication round, thus decreasing the local variance due to stochastic gradients.
It can also be viewed as having a larger batch size to decrease the variance in SGD. 
The cumulative variance of $K$ local steps contributes to the second term of $\Phi$.
This term depends on the number of local steps $K$, local learning rate $\eta_L$, local variance $\sigma_{L}^{2}$ and global variance $\sigma_{G}^{2}$ (non-i.i.d. data), but independent of $m$.
}
\end{rem}

\begin{rem} [Comparison with FedAvg without Compression]
{\em
Compared to the results of generalized FedAvg without compression, i.e., $\Phi \triangleq \frac{1}{c} [\frac{L \eta \eta_L}{2m} \sigma_L^2 + \frac{5K \eta_L^2 L^2}{2} (\sigma_L^2 + 6K \sigma_G^2)]$ (cf. \cite{yang2021achieving}), we have 
two key observations.
First, the compressor, which significantly reduces the communication cost, only slightly increases the constant $\Phi$ by $\frac{1}{c} [(\frac{1}{2} L^2 \eta_L^2 h(\gamma, \alpha) + \frac{1}{2}) \frac{\eta_L}{m} \sigma_L^2]$ and does not change the convergence rate $O(1/T)$.
This extra variance comes from increased local variance due to more noisy stochastic gradients after compression, and is independent of the number of local steps $K$.
This insight means that one can {\em safely} use more local steps $K$ without worrying about any accumulative effect due to compression, which is a somewhat surprising and counter-intuitive insight.
On the other hand, this extra variance shrinks at rate $\frac{1}{m}$, which favors large FL systems with more workers.
This makes intuitive sense since the server could obtain more information from the model updates with more workers, although each worker's information is noisy due to  compression. 
In other words, as the number of worker $m$ increases, the extra variance due to compression becomes negligible.
Second, the extra variance due to compression is irrelevant to the global variance $\sigma_{G}^{2}$ from the non-i.i.d. datasets.
The global variance measures the heterogeneity among the loss functions of the workers with non-i.i.d. local datasets.
Intuitively speaking, the compressor only introduces extra noise to the model's information.
Thus, the compression operation only increases the local variance of the stochastic gradient and is likely irrelevant to non-i.i.d. datasets and local steps in FL.
}
\end{rem}

\begin{rem} [Choice of Compressor]
{
\em
Our analysis also shows that many compression methods (e.g.,~\cite{bernstein2018signsgd, alistarh2017qsgd, wangni2018gradient}) that work well in traditional distributed and decentralized learning can also be used with the CFedAvg algorithm in FL.
With error feedback, CFedAvg enjoys the same benefits as traditional distributed learning even with the use of local steps in FL and under non-i.i.d. datasets.
Moreover, we do not restrict the choice of compressors.
Thus, CFedAvg works with both biased and unbiased compressors, as long as they satisfy Definition~\ref{def_compression}.
However, care must still be taken when one chooses a compressor in CFedAvg since it does not mean any compression methods with arbitrary compression rate could work.
Consider a compressor with arbitrary compression rate such that $\gamma \!\!\rightarrow\!\! 1$ and then $\epsilon \!\!\rightarrow\!\! 1$, so $h(\gamma, \alpha) \!\rightarrow\! \infty$.
This compressed model is too noisy to be trained since no useful information is available for the server.
This will also be empirically verified in Section~\ref{sec:exp}, where significant performance degradation due to an overly aggressive compression rate can be observed.
}
\end{rem}

From Theorem~\ref{thm:FedAvg_compression}, we immediately have the following convergence rate with a proper choice of learning rates:
\begin{restatable} {corollary}{FedAvgC}
(Linear Speedup for Convergence). \label{cor:FedAvg_compression}
If $\eta_L = \frac{1}{\sqrt{T}KL}$ and $\eta = \sqrt{Km}$, the convergence rate of Algorithm~\ref{alg:FedAvg_compression} is:
$\mathcal{O}( \frac{h(\gamma, \alpha)}{\sqrt{mKT}} + \frac{1}{T})= \mathcal{O}( \frac{1}{\sqrt{mKT}} + \frac{1}{T})$.
\end{restatable}

\begin{rem}
{\em
With a proper SNR-threshold $\gamma$ such that $h(\gamma, \alpha) = \mathcal{O}(1)$, 
CFedAvg achieves a linear speedup for convergence for non-i.i.d. datasets, i.e., $\mathcal{O}(\frac{1}{\sqrt{mKT}})$ convergence rate for $T \geq mK$.
This matches the convergence rate in distributed learning and FL without compression \cite{kairouz2019advances,karimireddy2019scaffold,yu2019linear,yang2021achieving}, indicating CFedAvg achieves high communication efficiency while not sacrificing learning accuracy in FL.
When degenerating to i.i.d. case, CFedAvg still achieves the linear speedup effect, matching the results of previous work in distributed and decentralize learning \cite{tang2019texttt,koloskova2019decentralized}.
The most related work to this paper is Qsparse-local-SGD~\cite{basu2020qsparse}, which combines unbiased quantization, sparsification and local steps together and is able to recover or generalize other compression methods. 
It achieves $\mathcal{O}( \frac{1}{\sqrt{mKT}} + \frac{mK}{T})$ convergence, implying a linear speedup for $T \geq m^3K^3$.
However, for large systems ($m$) and large local steps ($K$), their $T$ will be very large.
Besides a better convergence rate in this paper, we have a weak assumption (no bounded gradient assumption).
Challenges from this relaxed assumption are addressed in Section~\ref{sec:proofs}.
}
\end{rem}

{\bf 2)~CFedAvg with Decaying Learning Rates (Homogeneous Local Steps):} 
We can see from Corollary~\ref{cor:FedAvg_compression} that the choice of constant learning rate requires knowledge of time horizon $T$ before running the algorithm, which may not be available in practice.
In other words, the constant-learning-rate version of CFedAvg is not an {\em ``anytime''} algorithm.
To address this limitation, we propose CFedAvg with decaying learning rate, which is an anytime algorithm.
%
\begin{restatable} {theorem}{FedAvgDecaying}
(Convergence with Decaying Learning Rate). \label{thm:FedAvg_decay}
Choose decaying local learning rate $\eta_{L, t}$ and constant global learning rates $\eta$ s.t. $\eta_{L, t} \!\leq\! \frac{1}{8LK}$, $\eta \eta_{L, t} \!<\! \frac{1}{K L}$ and $ \eta \eta_{L, t} K ( L^2 h(\gamma, \alpha_t) \!+\! 1 \!+\! L) \!\leq\! 1, \forall t \in [T]$.
Under Assumptions~\ref{a_smooth}--\ref{a_variance}, the sequence $\{ \x_t \}$ generated by Algorithm~\ref{alg:FedAvg_compression} satisfies:
\begin{align}
\mathbb{E}\|\nabla f(\z)\|_2^2 \leq \frac{f_0 - f_{*}}{c \eta K H_T} + \Phi,
\end{align}
where $\Phi \triangleq (\frac{1}{2} L^2 h(\gamma, \alpha) + \frac{1}{2} + \frac{L}{2}) \frac{\eta}{c m H_T} \sigma_L^2 \sum_{t=0}^{T-1} \eta_{L, t}^2 + \frac{5 K L^2}{2 c H_T} (\sigma_L^2 + 6 K \sigma_G^2) \sum_{t=0}^{T-1} \eta_{L, t}^3$, $H_T = \sum_{t=0}^{T-1} \eta_{L, t}$ and $\z$ is sampled from $\{ \x_t \}, \forall t \in [T]$ with probability $\mathbb{P}[\z = \x_t] = \frac{\eta_{L, t}}{H_T}$.
Other parameters are defined the same as Theorem~\ref{thm:FedAvg_compression}.
\end{restatable}

\begin{restatable} {corollary}{FedAvgDecayingC}
\label{cor:FedAvg_compression_decaying}
Let $\eta_L \!\!=\!\! \frac{1}{\sqrt{t+a}KL}$ 
(constant $a\!>\!0$)
 and $\eta \!=\! \sqrt{Km}$,
the convergence rate of Algorithm~\ref{alg:FedAvg_compression} is:
$\mathcal{\tilde{O}}( \frac{1}{\sqrt{mKT}}) + \mathcal{O}(\frac{1}{\sqrt{T}})$.
\end{restatable}

\begin{rem}
{\em
When $h(\gamma, \alpha)$ is a constant, our CFedAvg algorithm achieves $\mathcal{O}(\frac{1}{\sqrt{mKT}} \ln(T) + \frac{1}{\sqrt{T}})$ convergence rate, which is slower than that with constant learning rates.
However, it is still better than $\mathcal{O}(1/\ln(T))$ proved in Qsparse-local-SGD~\cite{basu2020qsparse}.
}
\end{rem}

{\bf 3)~CFedAvg with Heterogeneous Local Steps (Constant Learning Rates):} 
In practice, FL systems are often formed by heterogeneous devices with various computing capabilities and resource limits (e.g., computation speed, memory size).
Hence, fixing the same number of local steps at all workers results in: i) fast workers being idle after finishing computation in each round and thus wasting resources and ii) slow workers being stragglers in the FL system.
Next, we show that it is possible to perform heterogeneous local steps among workers in CFedAvg while still offering theoretical performance guarantees.
\begin{restatable} {theorem}{FedAvgDistinct}
(Convergence of Heterogeneous Local Steps). \label{thm:FedAvg_dynamic}
Choose constant local and global learning rates $\eta_L$ and $\eta$ s.t. $\eta_L \!\leq\! \frac{1}{8LK_i}$, $\eta \eta_L \!<\! \frac{1}{K_i L},  \forall i \in [m]$ and $ \eta \eta_L ( L^2 h(\gamma, \alpha_t) \!+\! 1 \!+\! L) \!\leq\! 1, \forall t in [T]$.
Under Assumptions~\ref{a_smooth}--\ref{a_variance}, the sequence $\{ \x_t \}$ generated by Algorithm~\ref{alg:FedAvg_compression} with heterogeneous local steps satisfies:
\begin{align}
	\min_{t \in [T]} \mathbb{E}\|\nabla f(\x_t)\|_2^2 \leq \frac{f_0 - f_{*}}{c \eta \eta_L T} + \Phi,
\end{align}
where $\Phi \triangleq \frac{1}{c} [(\frac{1}{2} L^2 h(\gamma, \alpha) + \frac{1}{2} + \frac{L}{2}) \frac{\eta \eta_L}{m^2} \sum_{i = 1}^{m} \frac{1}{K_i} \sigma_L^2 + \frac{5 \eta_L^2 L^2}{2} \frac{1}{m} \sum_{i=1}^{m} K_i \\ (\sigma_L^2 + 6 K_i \sigma_G^2)]$.
Other parameters are defined the same as Theorem~\ref{thm:FedAvg_compression}.
\end{restatable}

\begin{restatable} {corollary}{FedAvgDistinctC}
(Linear Speedup with Heterogeneous Local Steps). \label{cor:FedAvg_compression_dynamic}
Let $\eta_L = \frac{1}{\sqrt{T}L}$ and $\eta = \sqrt{K_{min}m}$.
The convergence rate of Algorithm~\ref{alg:FedAvg_compression} with heterogeneous local steps and constant learning rate is:
$\mathcal{O}( \frac{1}{\sqrt{mK_{min}T}}) + \mathcal{O}(\frac{K_{max}^2}{T}),$
where $K_{min} = \min_{i \ in [m]} \{ K_i \}$ and $K_{max} = \max_{i \in [m]} \{ K_i \}$.
\end{restatable}

\begin{rem}
{\em
Allowing heterogeneous local steps at different workers entails efficient FL implementations in practice.
Specifically, instead of waiting for all workers to finish the same number of local steps,
the server can broadcast a periodic ``time-out'' signal to all workers to interrupt their local updates.
All worker can simply submit their current computation results even if they are in different stages in their local updates.
We can see from Corollary~\ref{cor:FedAvg_compression_dynamic} that CFedAvg with heterogeneous local steps can still achieve the same linear speedup for convergence $\mathcal{O}( \frac{1}{\sqrt{mK_{min}T}})$ for $T \geq m K_{min} K^4_{max}$, while having the flexibility of choosing different number of local steps at each worker. 
Although this result is for CFedAvg, our theoretical analysis is general and can be applied to uncompressed FL algorithms (e.g., FedAvg) to allow heterogeneous $K^t_i, \forall i \in [m]$.

To our knowledge, our work is the first to show that performing heterogeneous local steps among workers still achieves theoretical performance guarantees for FL.
We note that, in asynchronous Qsparse-local-SGD~\cite{basu2020qsparse}, the workers synchronize with the server at different times based on the workers, but it is required that each worker follows the same rate and performs the same local steps.
Rizk et al.~\cite{rizk2020dynamic} proposed dynamic federated learning (without compression) to use heterogeneous local steps at each worker.
But they require the local steps to be known in advance in order to scale the gradient in each local step.
For our CFedAvg, $K_i$ can be set in an ad-hoc fashion without being known in advance. 
}
\end{rem}

\subsection{Proof of the Main Results} \label{sec:proofs}

\begin{proof} [Proof Sketch for Theorem~\ref{thm:FedAvg_compression}]
Due to space limitation, we only provide a proof sketch here.
For convenience, we define the following notation:
$\e_t \triangleq \frac{1}{m} \sum_{i=1}^{m} \e_t^i$, $\hat{\x}_{t} \triangleq \x_t + \eta \e_t$ and
$\Delta_t = \frac{1}{m} \sum_{i \in [m]} \Delta_t^i = \frac{1}{m} \sum_{i \in [m]} \g_t^i$.
Note that we do not assume bounded gradients, which poses two challenges.
One key difficulty is to bound the error feedback term $\e_t^i, \forall i \in [m]$ due to the compression.
$\e_t^i$ cannot be bounded in this fashion if the bounded gradient assumption is relaxed.
In our analysis, we manage to bound $\e_t$ rather than $\e_t^i$ individually.
By using
$
\| \e_t \|^2 = \big\| \frac{1}{m} \sum_{i=1}^{m} \e_t^i \big\|^2 \leq \frac{1}{m} \sum_{i=1}^{m} \big\| \e_t^i \big\|^2 = \big\| \bar{\e}_t \big\|^2 $
and 
the recursion 
$
\big\| \bar{\e}_t \big\|^2 \leq (1/\gamma) \frac{1}{\epsilon} \big\| \bar{\e}_{t-1} \big\|^2 + (1/\gamma)\frac{1}{1 - \epsilon} \big\| \bar{\Delta}_{t-1} \big\|^2, 
$
where $\| \bar{\Delta}_t \|^2 = \frac{1}{m} \sum_{i=1}^{m} \big\| \Delta_t^i \big\|^2$, $\sum_{t=0}^{T-1} \| \e_t \|^2$ can be bounded in Lemma~\ref{lem: Bounded_error}.
Second, the lack of bounded gradient assumption also results in difficulty in bounding the model drift stemming from the non-i.i.d. datasets and the increase of the local steps, i.e., $\| \x_i - \bar{\x} \|, \forall i \in [m]$, where $\bar{\x} = \frac{1}{m} \sum_{i= 1}^{m} \x_i$.
To address this challenge, we derive the recursive relation based on communication round instead of local steps.
Thanks to the virtual variable $\hat{\x}_t$, we have the recursive relation of $\hat{\x}_t$: $\hat{\x}_{t+1} \!=\! \hat{\x}_t \!+\! \eta \Delta_t$, where $t$ is the index of communication round.

After addressing these two challenges, Assumption~\ref{a_smooth} yields per-communication-round descent as follows:
\begin{align}
\hspace{-0.05in}
	\mathbb{E}_t f(\hat{\x}_{t+1}) \!-\! f(\hat{\x}_t)
	&\leq \underbrace{\mathbb{E}_t  \big< \nabla f(\hat{\x}_t), \eta \Delta_t \big> }_{A_1} + \frac{L \eta^2}{2} \underbrace{ \mathbb{E}_t \| \Delta_t \|^2 }_{A_2}. \label{ineq_compression_smooth}
\end{align}
%

For $A_1$, the first step is to transform variable $\hat{\x}_t$ to $\x_t$ since only $\x_t$ is involved in the update.
Towards this end, we have:
\begin{align}
A_1
\leq& \mathbb{E}_t \big[ \frac{1}{2} L^2 \eta^2 \| \e_t \|^2 + \frac{1}{2} \eta^2 \| \Delta_t \|^2 - K \eta \eta_L \| \nabla f(\x_t) \|^2 \nonumber\\
&+ \eta \big< \nabla f(\x_t), \Delta_t + K \eta_L \nabla f(\x_t) \big> \big]. \label{A_1}
\end{align}

Further bounding the last term of equation~\eqref{A_1}, we have:
\begin{align}
&A_1 \leq \mathbb{E}_t \big[ \frac{1}{2} L^2 \eta^2 \| \e_t \|^2 + \frac{1}{2} \eta^2 \| \Delta_t \|^2 \big] - K \eta \eta_L  \| \nabla f(\x_t) \|^2 \nonumber\\
&+ \eta \eta_L K (\frac{1}{2} + 15 K^2 \eta_L^2 L^2) \| \nabla f(x_t) \|^2 + \frac{5 \eta K^2 \eta_L^3 L^2}{2} \times \nonumber \\
& (\sigma_L^2 + 6 K \sigma_G^2) - \frac{\eta \eta_L}{2K m^2} \mathbb{E}_t \big\| \sum_{i=1}^{m} \sum_{k=0}^{K-1} \nabla F_i(\x_{t,k}^i) \big\|^2. \label{ineq_A1}
\end{align}

For $A_2$, by using $\mathbb{E}[\| \x \|^2] = \mathbb{E}[\| \x - \mathbb{E}[\x] \|^2] + \| \mathbb{E}[\x] \|^2$ to decompose $\Delta_t$ and assumption~\ref{a_variance}, we have:
\begin{equation}
\hspace{-0.05in}
A_2 = \mathbb{E}_t [\| \Delta_t \|^2] \leq \frac{K \eta_L^2}{m} \sigma_L^2 \!+\! \frac{\eta_L^2}{m^2} \big\| \sum_{i=1}^{m} \sum_{k=0}^{K-1} \nabla F_i(\x_{t,{}k}^i) \big\|^2. \label{ineq_A2}
\end{equation}

Combining \eqref{ineq_A1} and \eqref{ineq_A2}, $\mathbb{E}_t \big\| \sum_{i=1}^{m} \sum_{k=0}^{K-1} \nabla F_i(\x_{t,k}^i) \big\|^2$ is canceled with proper learning rates. 
Then, 
we simplify \eqref{ineq_compression_smooth} as:
\begin{align*}
&\mathbb{E} f(\hat{\x}_{t+1}) - \nabla f(\hat{\x}_t) 
\leq (\frac{1}{2} L^2 \eta^2 h(\gamma, \alpha_t) + \frac{1}{2} \eta^2 + \frac{L \eta^2}{2}) \frac{K \eta_L^2}{m} \sigma_L^2 \\
& \indent - c \eta \eta_L K \big\| \nabla f(x_t) \big\|^2+ \frac{5 \eta K^2 \eta_L^3 L^2}{2} (\sigma_L^2 + 6 K \sigma_G^2) .
\end{align*}

Telescoping and rearranging yields the stated result.
\end{proof}

Theorems~\ref{thm:FedAvg_decay} and ~\ref{thm:FedAvg_dynamic} can also be proved in a similar fashion.



\section{Experimental Results} \label{sec:exp}

\smallskip
{\bf 1) Experiment Settings:}
{\em 1-a) Datasets and Models:}
We use three datasets (non-i.i.d. version) in FL settings, including MNIST, Fashion-MNIST 
 and CIFAR-10 
 .
Each of these three datasets contains $10$ different classes of items.
To induce non-i.i.d. datasets, we partition the data based on the classes of items ($p$)  contained in these datasets.
By doing so, we can use the number of classes in worker's local dataset, denoted as $p$, to control the non-i.i.d. level of the datasets quantitatively.
We set four levels of non-i.i.d. datasets for comparison: $p = 1, 2, 5,$ and $10$.
We distribute the dataset among $m= 100$ workers randomly and evenly in a class-based manner, such that the local dataset at each worker contains only a subset of classes of items with the same number of training/test samples.
We experiment two learning models: i) convolution neural network (CNN) (architecture is detailed in our online technical report) on MNIST and Fashion-MNIST, and ii) ResNet-18 on CIFAR-10.
{\em 1-b) Compressors:}
We consider two compression methods: i) Top-$k$ sparsification and ii) random dropping (RD)~\cite{aji2017sparse,stich2018sparsified}.
For a given vector $\x$, Top-$k$ sparsification compresses $\x$ by retaining $k$ elements of this vector that have the largest absolute value and setting others to zero, while RD randomly drops each component with a fixed probability.
we use a constant compression parameter $comp \in (0, 1]$ to represent the compression rate.
{\em 1-c) Hyper-parameters:}
We set the default hyper-parameters as follows:
the number of workers $m \!\!=\!\! 100$, 
local learning rate $\eta_L \!\!=\!\! 0.1$, global learning rate $\eta \!\!=\!\! 1.0$, batch size $B \!\!=\!\! 64$, local steps $K \!\!=\!\! 10$ epochs, total rounds $T \!\!=\!\! 100$ for MNIST and Fashion-MNIST and $T\!\!=\!\!200$ for CIFAR-10.

\smallskip
{\bf 2) Numerical Results:}
We now present two types of experimental results.
The first is to show the effectiveness of our CFedAvg algorithm with significant communication reduction.
The second is to evaluate the importance of error feedback.
We only show a part of results due to space limitation.

\begin{figure}[t!]
	\hfill
	\begin{minipage}[t]{0.49\linewidth}
	\centering
	{\includegraphics[width=0.9\columnwidth]{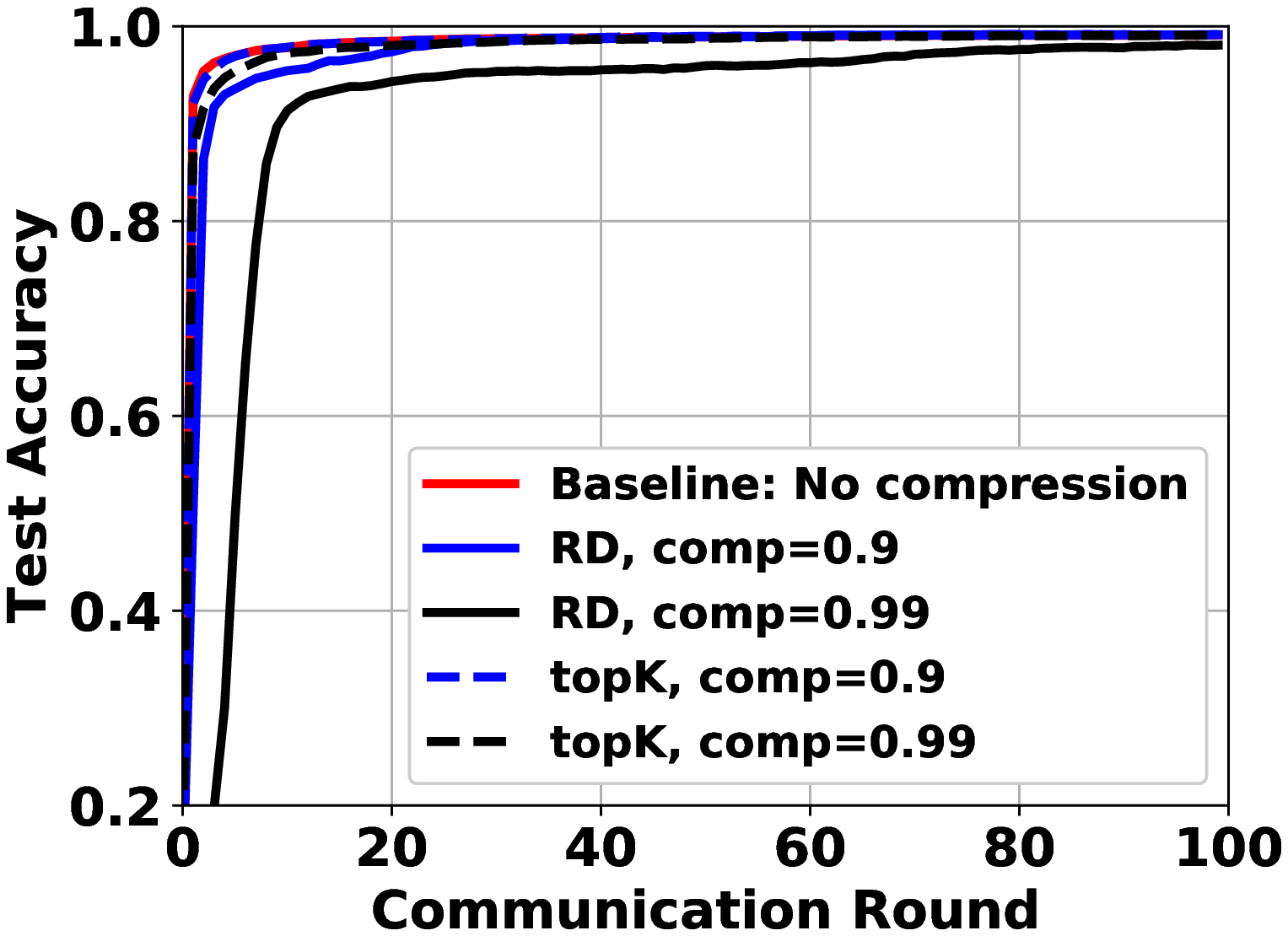}} 
	\vspace{-.2in}
	\end{minipage}
	\begin{minipage}[t]{0.49\linewidth}
	\centering
	{\includegraphics[width=0.94\columnwidth]{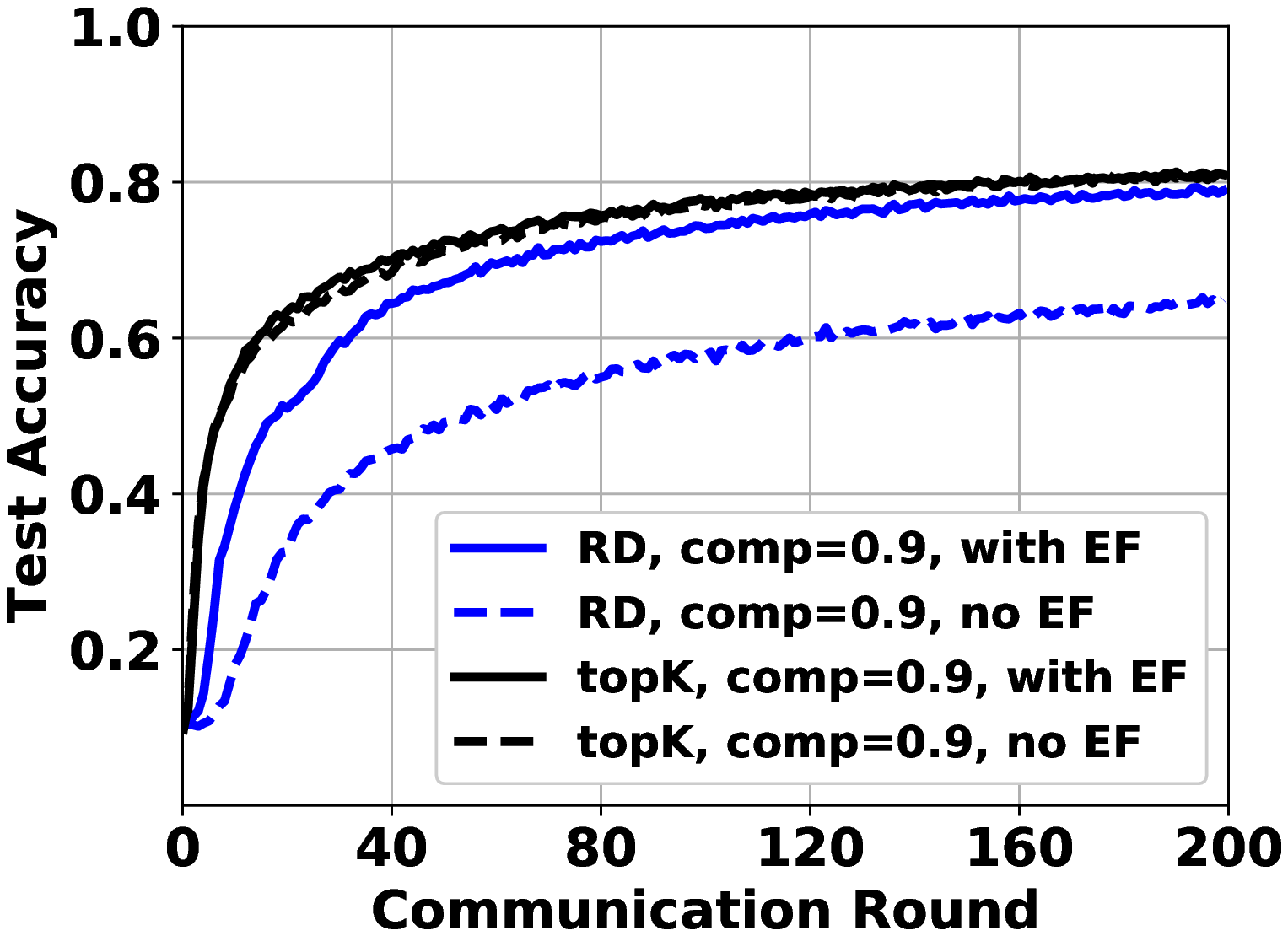}} 
	\vspace{-.2in}
	\end{minipage}

	\hfill
	\begin{minipage}[t]{0.49\linewidth}
	\centering
	{\includegraphics[width=0.9\columnwidth]{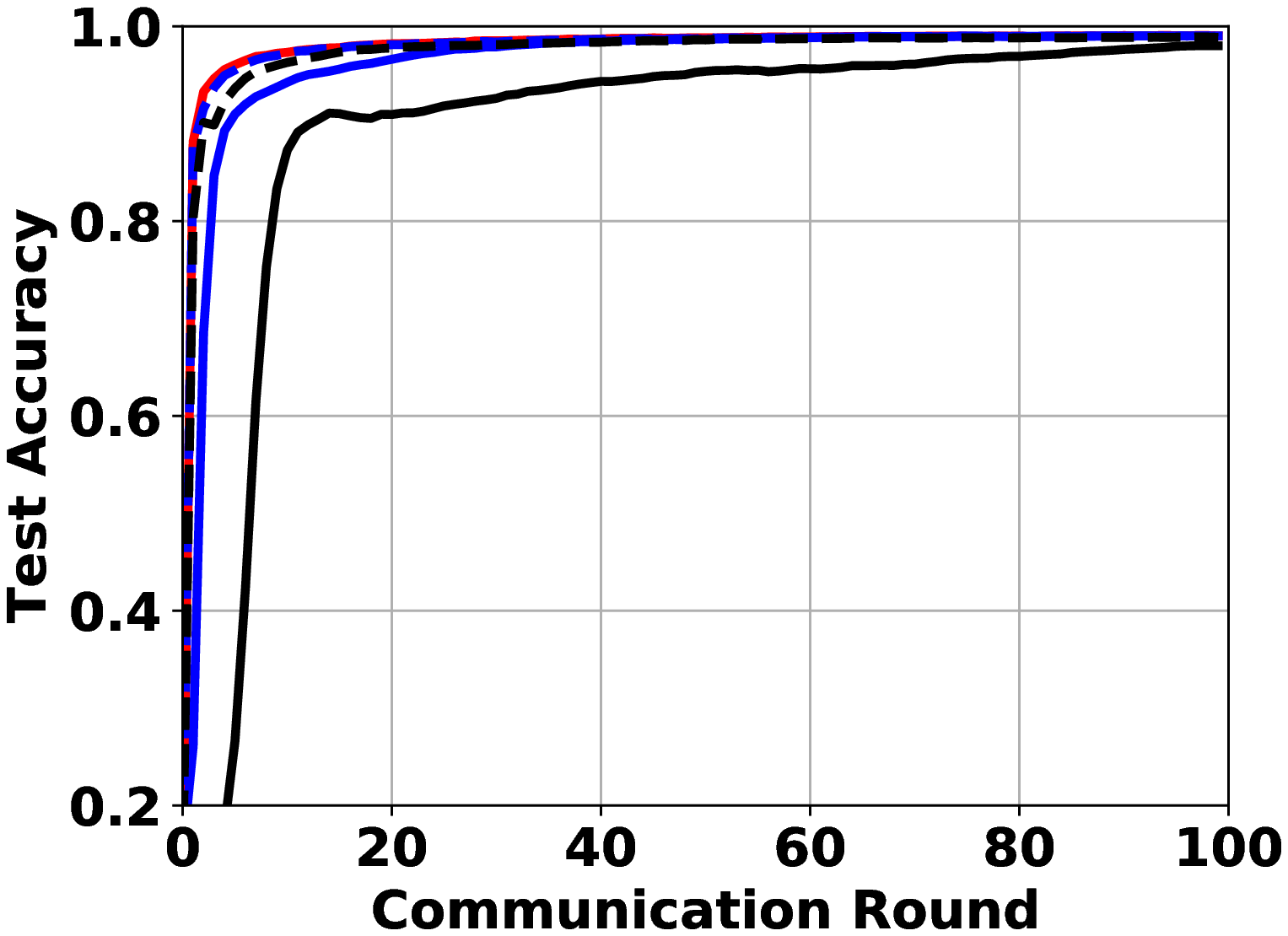}}
	\vspace{-.2in}
	\end{minipage}
	\begin{minipage}[t]{0.49\linewidth}
	\centering
	{\includegraphics[width=0.94\columnwidth]{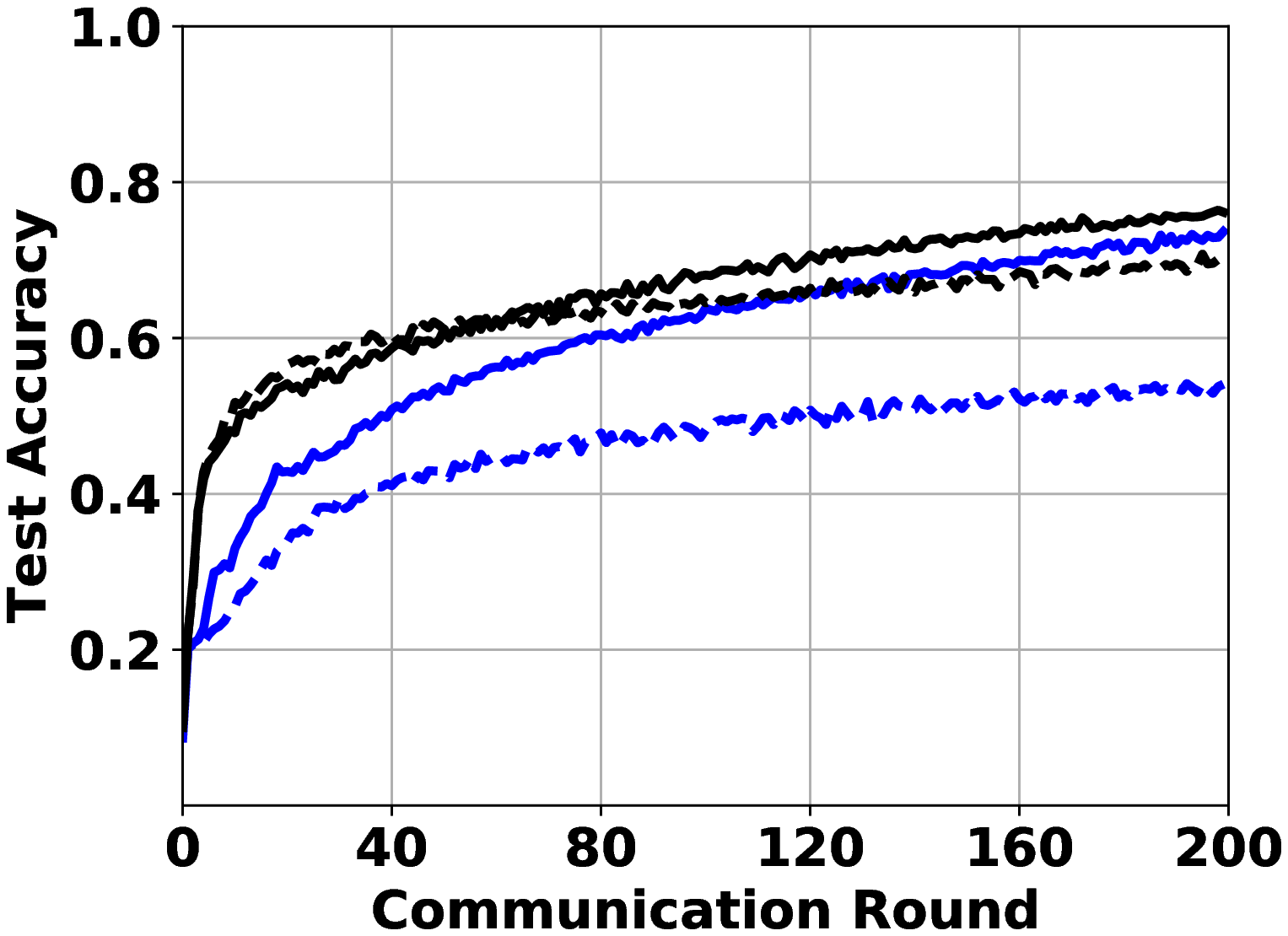}}
	\vspace{-.2in}
	\end{minipage}

	\hfill
	\begin{minipage}[t]{0.49\linewidth}
	\centering
	{\includegraphics[width=0.9\columnwidth]{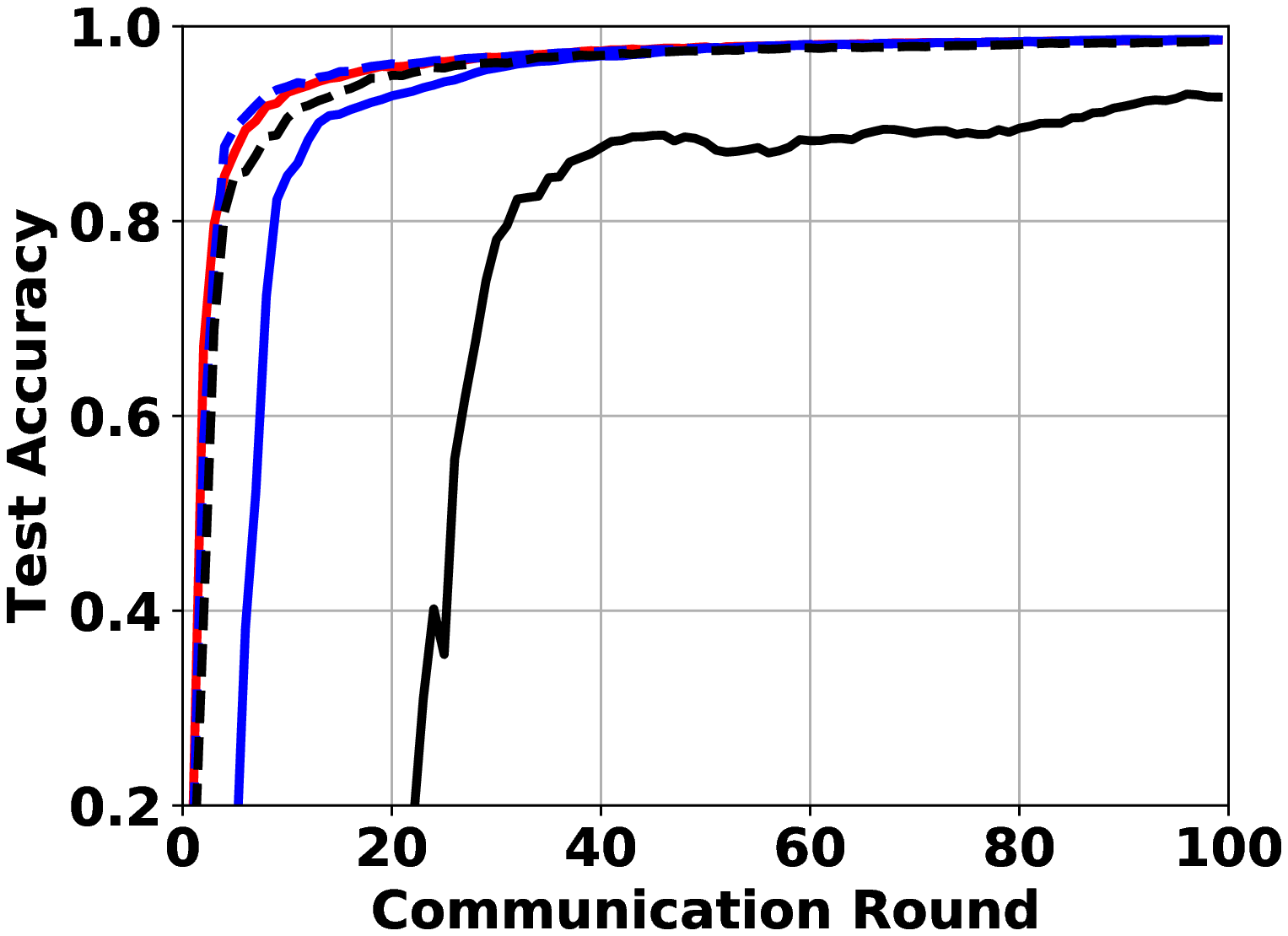}}
	\vspace{-.2in}
	\end{minipage}
	\begin{minipage}[t]{0.49\linewidth}
	\centering
	{\includegraphics[width=0.94\columnwidth]{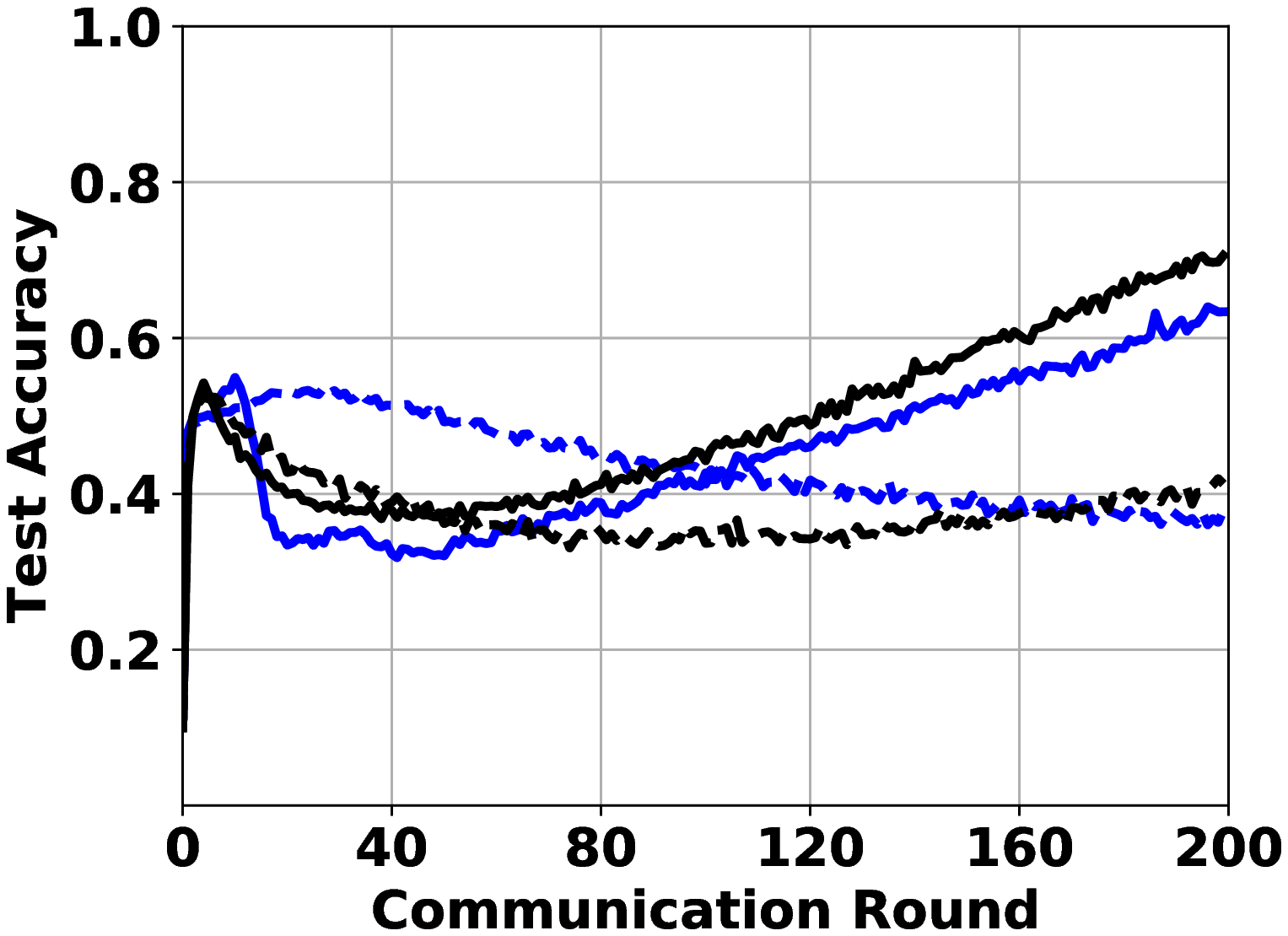}}
	\end{minipage}

	\begin{minipage}[t]{0.49\linewidth}
	\centering
	{\includegraphics[width=0.9\columnwidth]{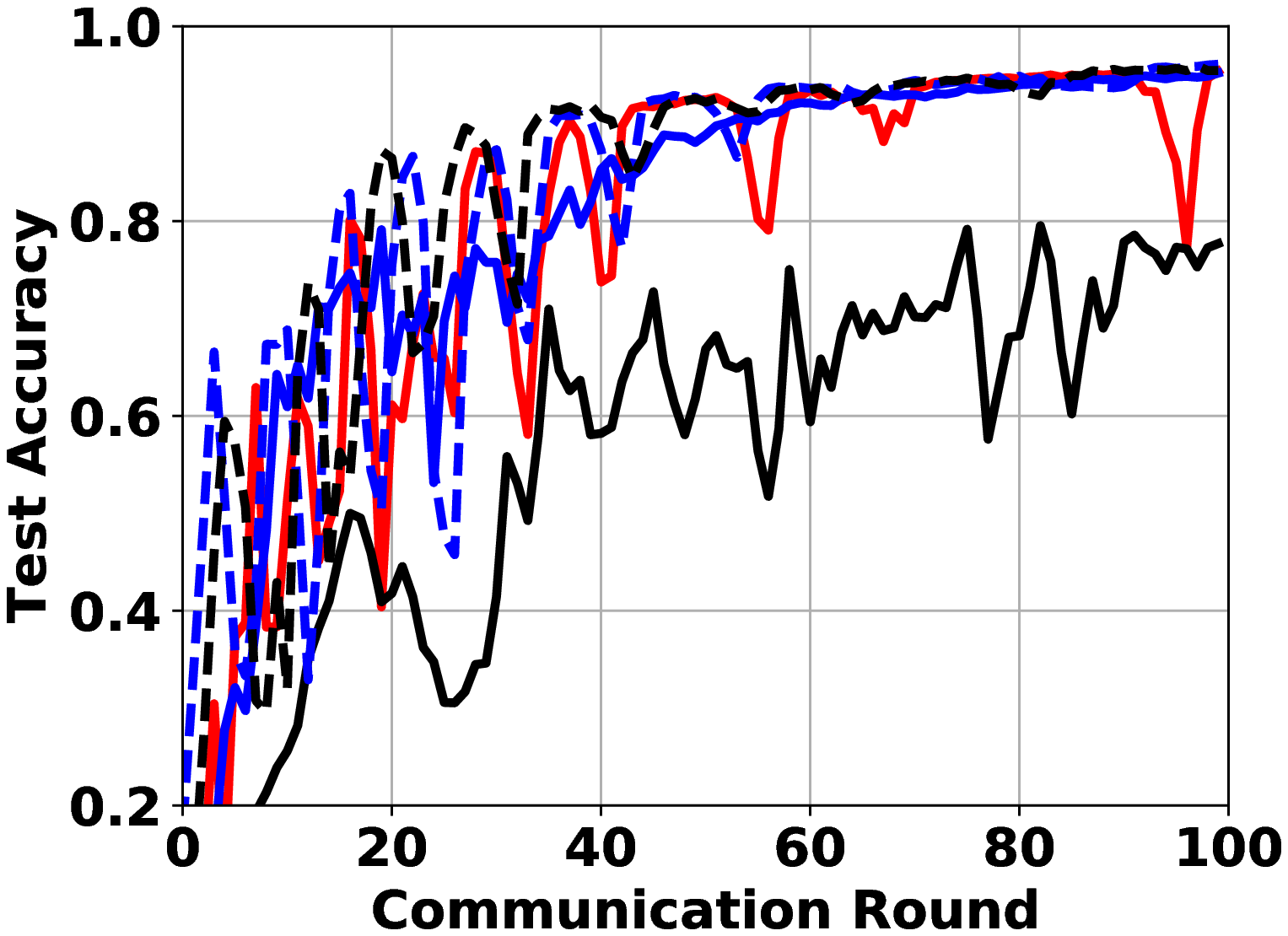}}
	\vspace{-.2in}
	\end{minipage}
	\begin{minipage}[t]{0.49\linewidth}
	\centering
	{\includegraphics[width=0.94\columnwidth]{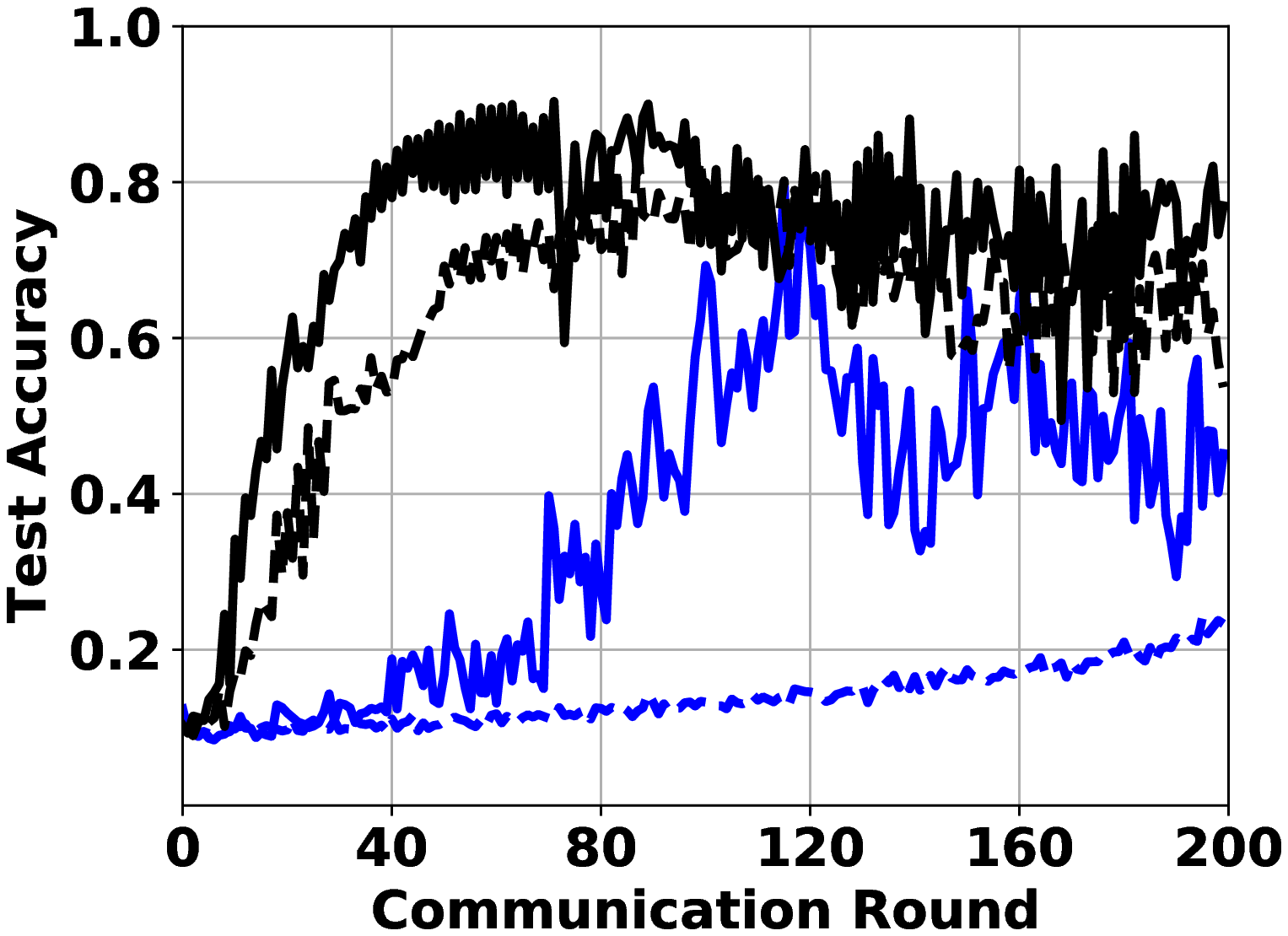}}
	\end{minipage}
\vspace{-.1in}
\caption{Test accuracy (left: MNIST, right: CIFAR-10). The non-i.i.d. levels are $p = 10, 5, 2, 1$ from top to bottom.}
\label{fig1}
\end{figure}%

\begin{figure}[!ht]
\vspace{-.1in}
\centering
	\begin{minipage}[t]{0.49\linewidth}
	\centering
	{\includegraphics[width=0.9\textwidth]{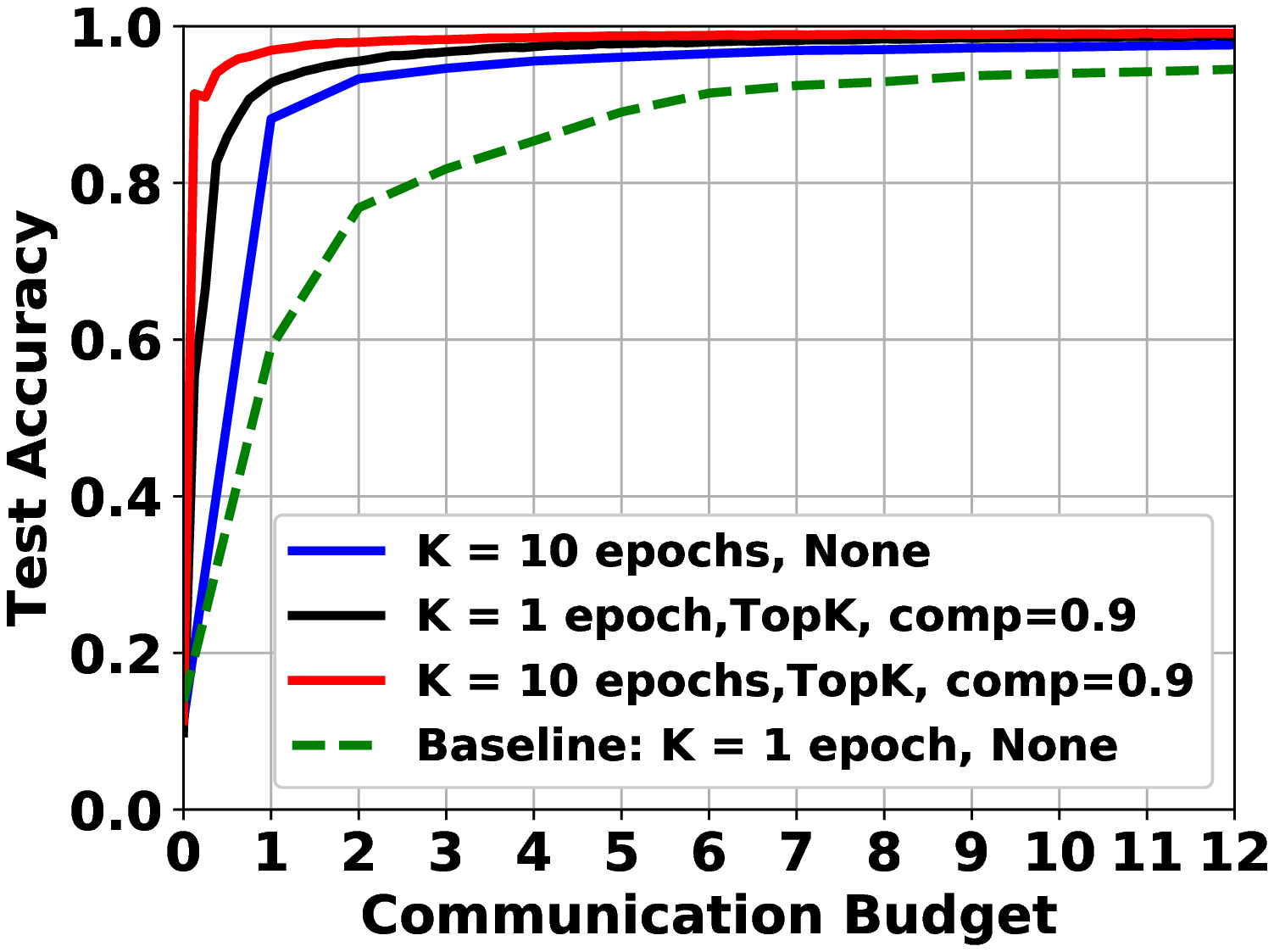}} 
	\text{ (a) MNIST.}
	\end{minipage}
	\begin{minipage}[t]{0.49\linewidth}
	\centering
	{\includegraphics[width=0.9\textwidth]{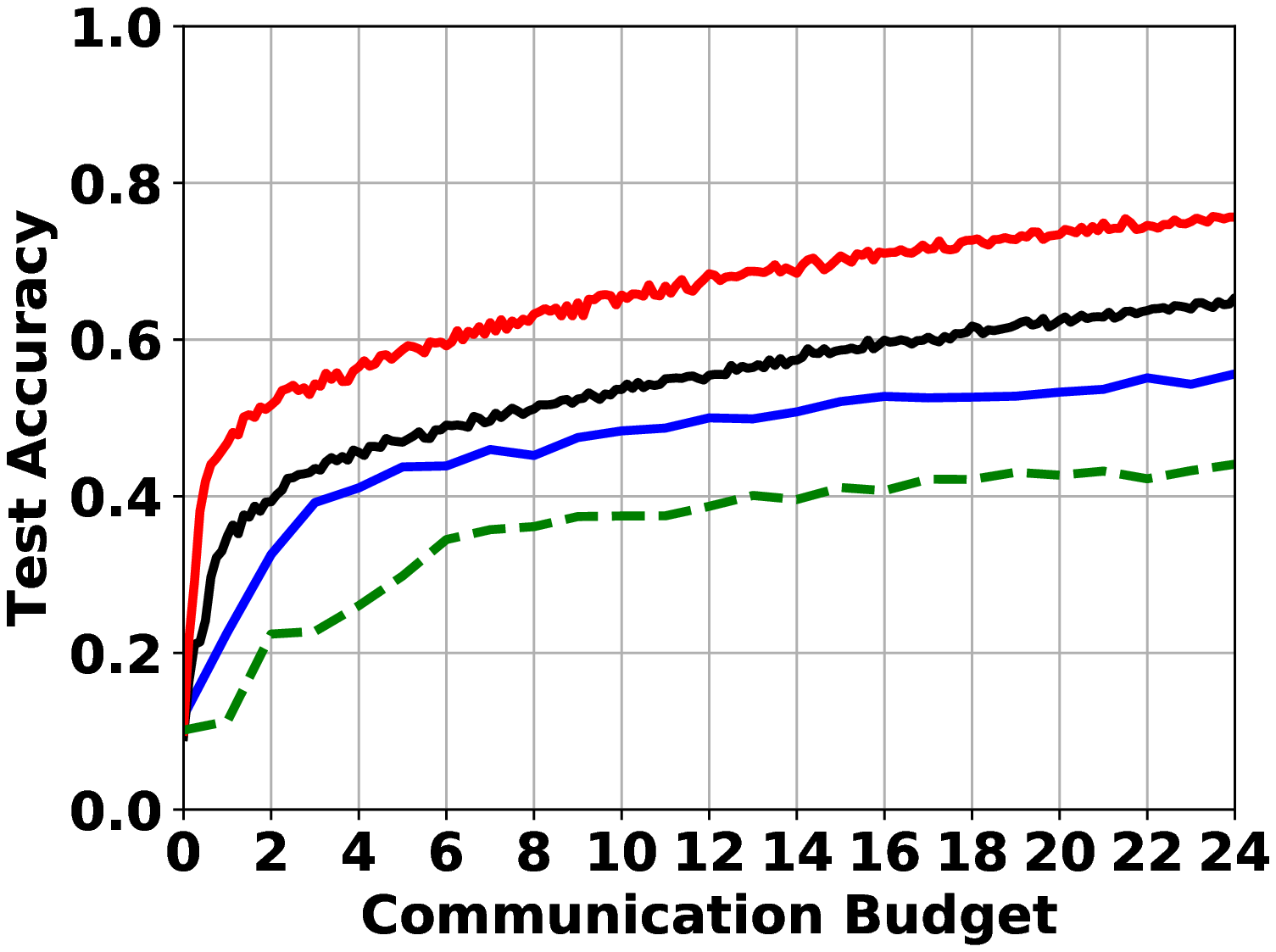}}
	\text{ (b) CIFAR-10.}
	\end{minipage}
\vspace{-.1in}
\caption{Comparison of test accuracy with same communication load for $p = 5$. The unit of the communication budget is the model size.}
\label{fig2}
\end{figure}

{\em 2-a) Effectiveness of Compressors:}
As shown in Fig.~\ref{fig1}, the figures are for test accuracy versus communication rounds of MNIST in the left column and CIFAR-10 in the right colomn. 
We can see that our CFedAvg algorithm with two compressors converges for all  heterogeneity levels of non-i.i.d. datasets from top ($p=10$) to bottom ($p=1$).
%
For the random dropping (RD) method, it becomes worse from $comp=0.9$ to $comp=0.99$.
While for Top-$k$ method, both cases ($comp=0.9$ and $comp=0.99$) can achieve almost the same convergence speed comparable to that without any compression.
This indicates that we can reduce the communication cost in each communication round by about $99\%$ with Top-$k$ for MNIST, i.e., we only need to transmit $1\%$ of coordinates in the gradient vector without significantly sacrificing the convergence rate.
This will greatly facilitates FL on such communication-constrained devices.
Another interesting observation is that the compression methods (with error feedback) help stabilize the training process for non-i.i.d. case.
Compared with i.i.d. datasets, the training curve is zigzagging for non-i.i.d. case.
As the heterogeneity level of non-i.i.d. datasets increases, this zigzagging phenomena of the curves is more pronounced as shown in the figures, which we believe is an inherent feature of non-i.i.d. dataset in FL.
Meanwhile, the learning curves are smoother with compression and error feedback, particularly with highly non-i.i.d. datasets, see Fig.~\ref{fig1} ($p=1$).
The intuition is that the compressor could filter some noises that lead to the instability of the learning curve due to model heterogeneity among workers originated from the non-i.i.d. datasets and local steps.
However, this appears to require proper compression rate $comp$ based on the compression method.
We do not rule out the possibility of mutual interactions that lead to poor performance between the compression and non-i.i.d. datasets.

In addition, we compare the test accuracy based on the same uplink communication budgets (the model size as the unit) among different compression methods under non-i.i.d. datasets $p=5$ in Fig.~\ref{fig2}.
The baseline is the FedAvg with a single  local step and no compression.
Consistent with previous work \cite{mcmahan2016communication,lin2017deep}, both compressors and local update steps are effective to reduce communication cost, confirming our theoretical analysis.


{\em 2-b) Importance of Error Feedback:}
Although it is a natural idea to apply those compression methods that have been proved to be useful in traditional distributed learning to FL,
there could be a significant information loss if one uses these compressors naively.
It has been shown that the learning performance is poor without error feedback in compression under i.i.d. datasets \cite{karimireddy2019error,stich2019error}.
This conclusion is confirmed in our experiments as shown in right column of Fig.~\ref{fig1}, where we observe the gap between cases with and without error feedback (EF) under i.i.d. case ($p=10$).
As the heterogeneity degree of non-i.i.d. datasets increases from $p=10$ to $p=1$, the gap becomes larger and is no longer negligible.
It is obvious that both compression methods, RD and Top-$k$, perform better with error feedback in Fig.~\ref{fig1}.
This indicates the significant impact of error feedback.
If naively applying compression, a huge amount of information could be lost, thus resulting in poor performance. 

\begin{figure}[t!]
	\begin{minipage}[t]{0.49\linewidth}
	\centering
	{\includegraphics[width=0.9\textwidth]{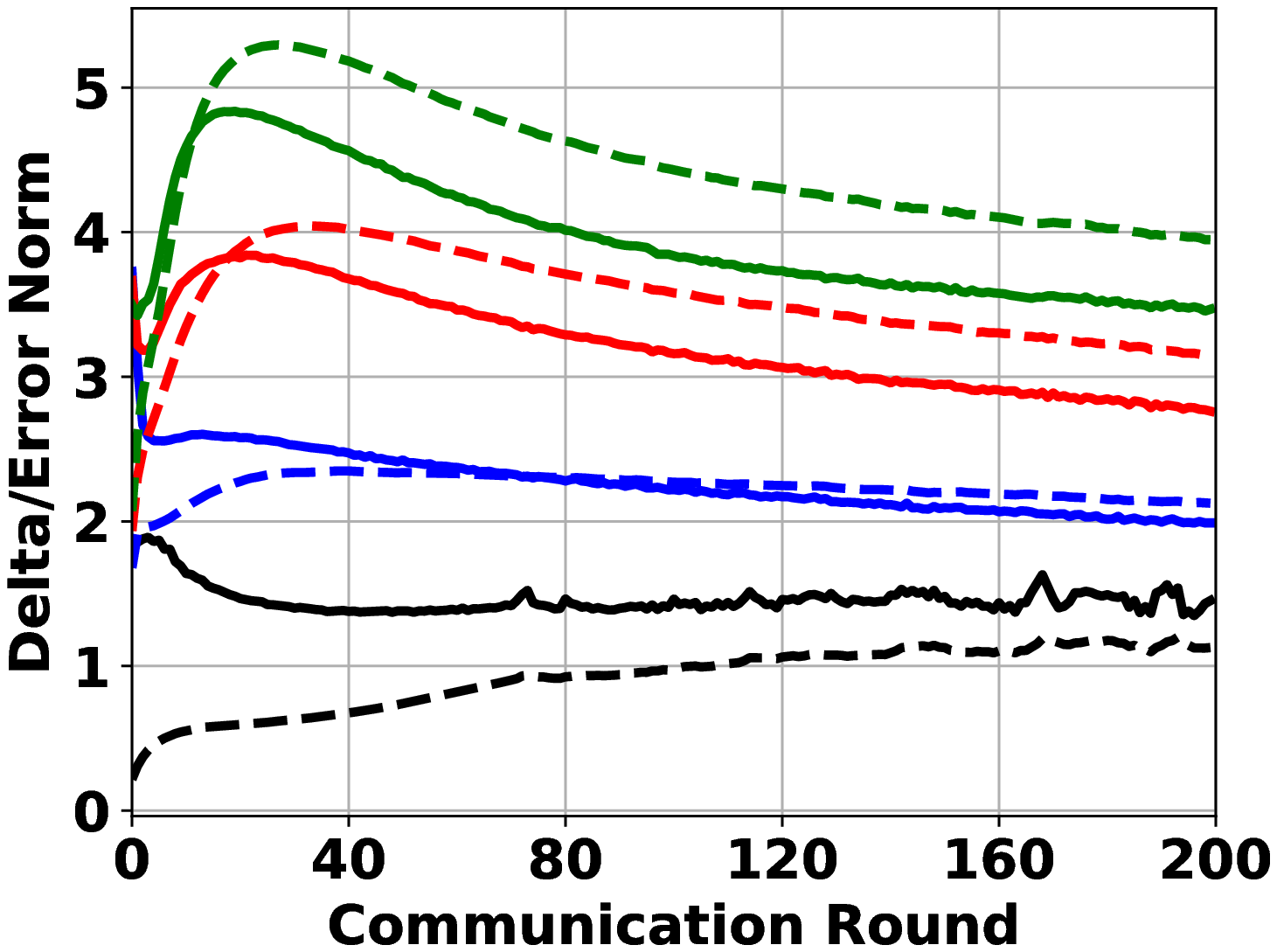}}
	\text{ (a) Top-K $comp=0.9$.}
	\end{minipage}
	\begin{minipage}[t]{0.49\linewidth}
	\centering
	{\includegraphics[width=0.93\textwidth]{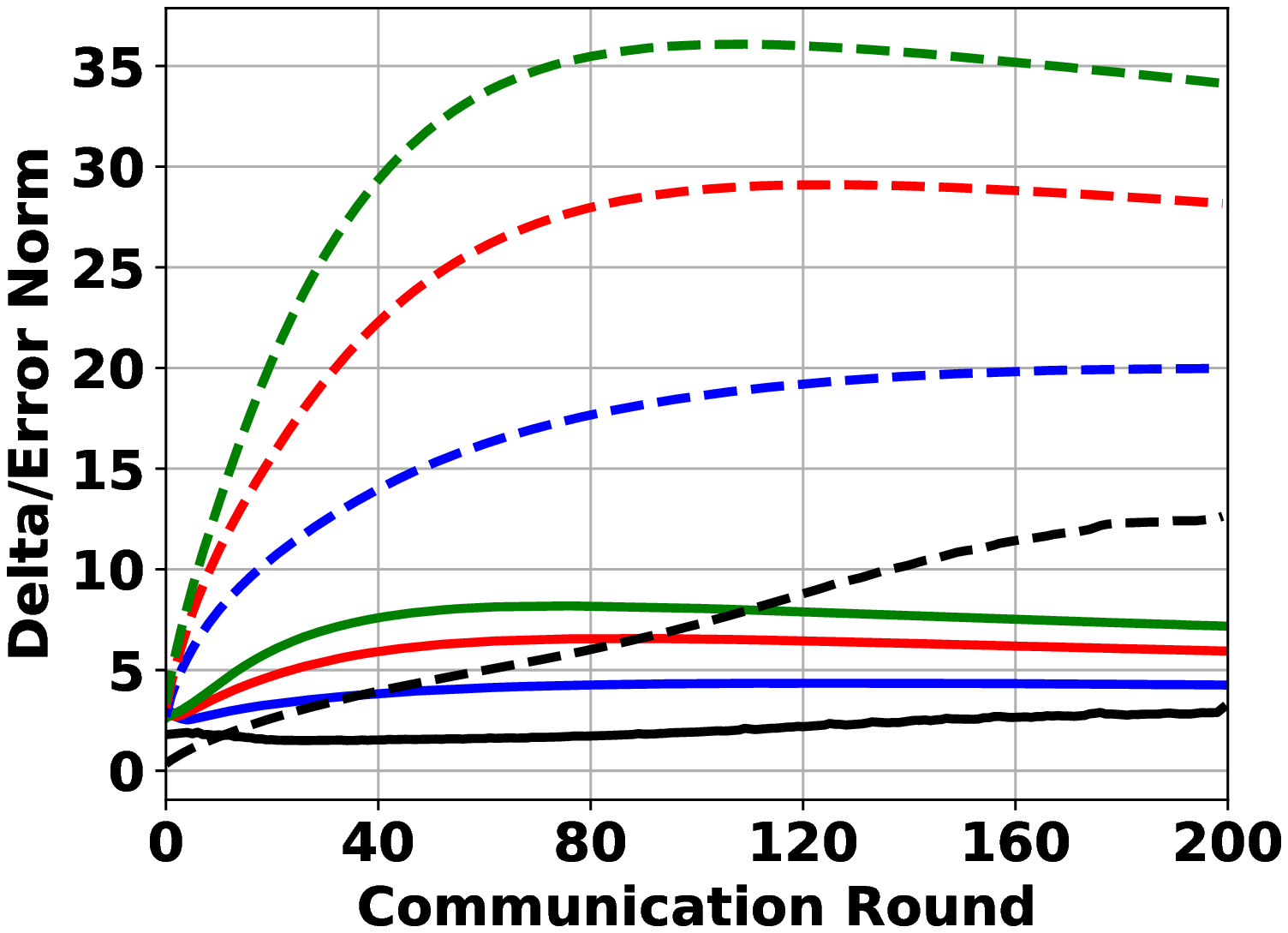}}
	\text{ (b) Top-K $comp=0.99$.}
	\end{minipage}
	\begin{minipage}[t]{0.49\linewidth}
	\centering
	{\includegraphics[width=0.9\textwidth]{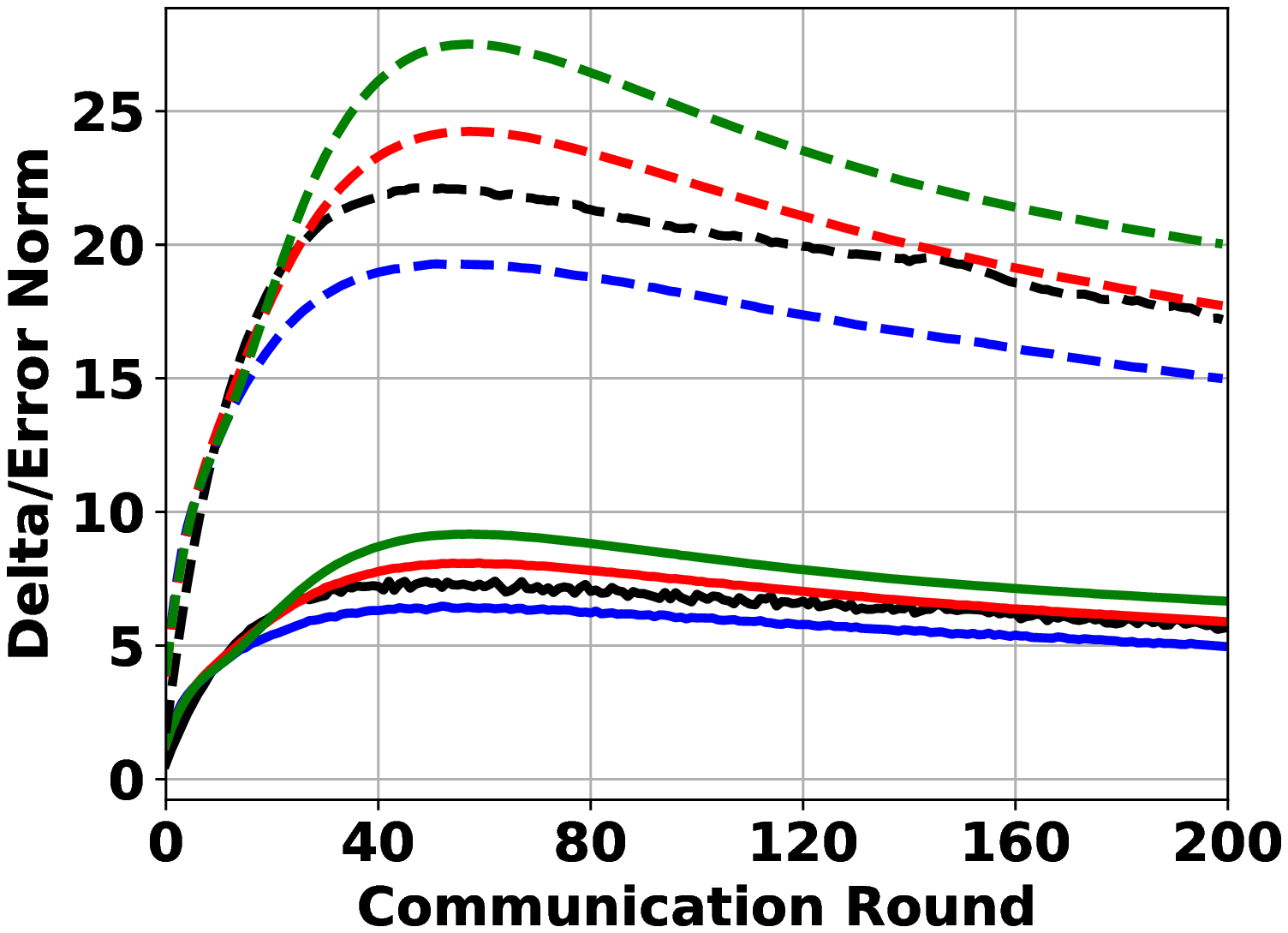}} 
	\text{ (c) RD $comp=0.9$.}
	\end{minipage}
	\begin{minipage}[t]{0.49\linewidth}
	\centering
	{\includegraphics[width=0.93\textwidth]{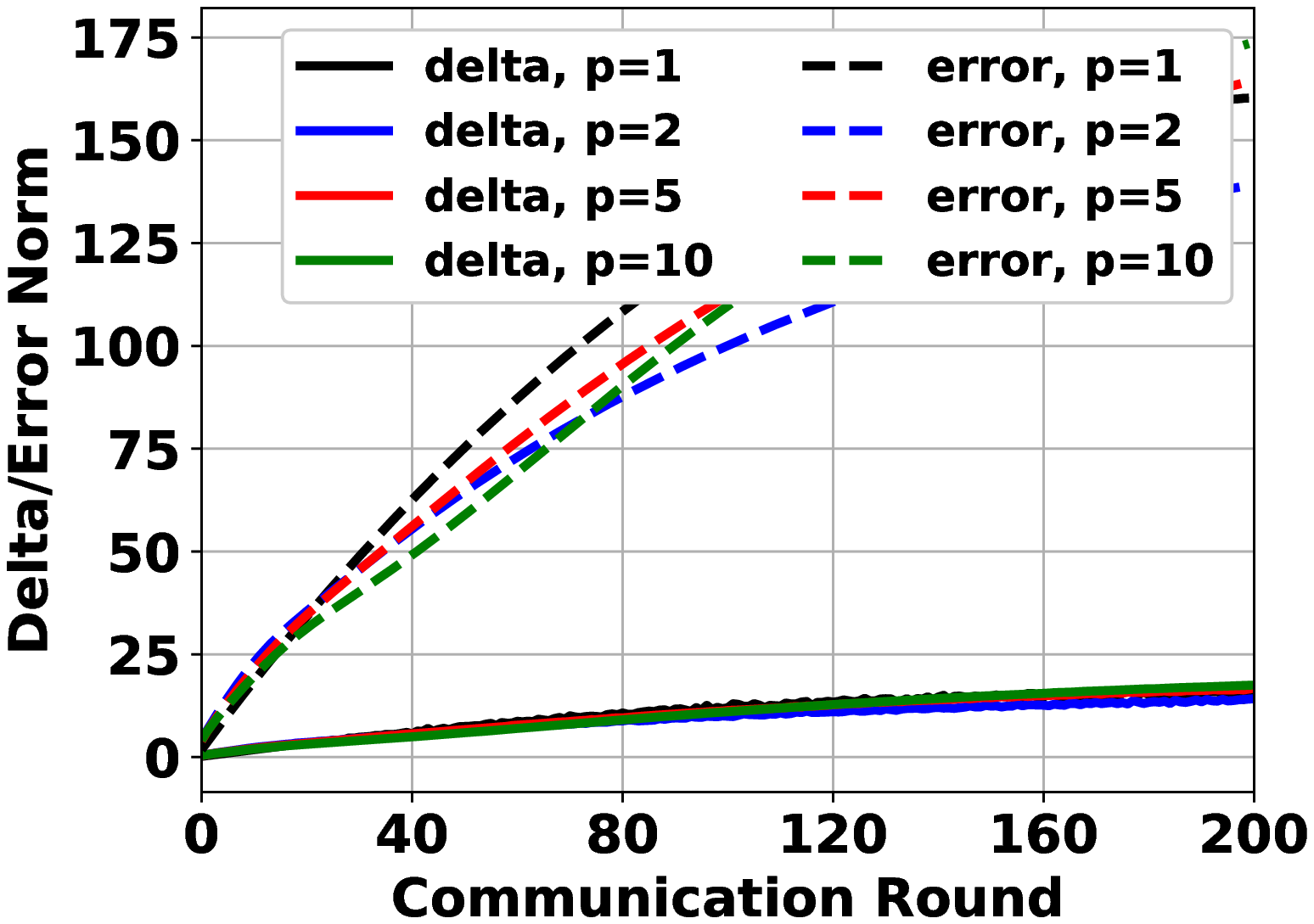}}
	\text{ (d) RD $comp=0.99$.}
	\end{minipage}

\caption{Mean of the norms of $\frac{1}{m} \sum_{i=1}^{100} \|\Delta_t^i \|^2$ and the error term $\frac{1}{m} \sum_{i=1}^{100} \|\e_t^i \|^2$ for the ResNet-18 on CIFAR-10.}
\label{fig4}
\end{figure}%

With error feedback, the error term accumulates the information that is not transmitted to the parameter server in the current communication round and then compensates the gradients in the next communication round.
This is verified in Fig.~\ref{fig4}, which shows the mean of gradient norm changes $\frac{1}{m} \sum_{i=1}^{m} \|\Delta_t^i \|^2$ and the error term $\frac{1}{m} \sum_{i=1}^{m} \|\e_t^i \|^2$ for the total worker number $m=100$.
One key observation is that the error term is bounded under appropriate compression methods and compression rates, which is usually several times of the gradient change term in general.
Under the same condition, the error term is much larger with aggressive compression rate.
With an overly aggressive compression (e.g., Fig.~\ref{fig4} with $comp=0.99$), the error term continues to grow.
Thus, in general, error feedback guarantees not too much information is dropped due to compression, verifying our theoretical analysis.
It has been shown that error feedback is effective in distributed/decentralized learning \cite{karimireddy2019error,stich2019error,koloskova2019decentralized}.
In this paper, we show that its effectiveness continues to hold in non-i.i.d. compressed FL.

\section{Conclusion} \label{sec:con}
In this paper, we proposed a communication-efficient algorithmic framework called CFedAvg for FL on non-i.i.d. datasets.
CFedAvg works with general (biased/unbiased) SNR-constrained compressors to reduce the communication cost and other techniques to accelerate the training.
Theoretically, we analyzed the convergence rates of CFedAvg for non-convex functions with constant learning and decaying learning rates. 
The convergence rates of CFedAvg match that of distributed/federated learning without compression, thus achieving high communication efficiency while not significantly sacrificing learning accuracy in FL.
Furthermore, we extended CFedAvg to heterogeneous local steps with convergence guarantees, which allows different workers perform different numbers of local steps to better adapt to their own circumstances.
Our key observation is that the noise/variance introduced by compressors does not affect the overall convergence rate order for non-i.i.d. FL. 
We verified the effectiveness of our CFedAvg algorithm on three datasets with two gradient compression schemes of different compression ratios.
Our results contribute to the first step toward developing advanced compression methods for communication-efficient FL.

\bibliographystyle{ACM-Reference-Format}
\bibliography{BIB/CommunicationEfficiency,BIB/FederatedLearning}

\newpage
\appendix
\allowdisplaybreaks[1]


\section{Appendix I: Proof} \label{proof}

\subsection{Proof of Lemma}
\lemmaError*

\begin{proof} [Proof of Lemma~\ref{lem: Bounded_error}]

Let $\| \bar{\e}_t \|^2 = \frac{1}{m} \sum_{i=1}^{m} \big\| \e_t^i \big\|^2$ and $ \| \bar{\Delta}_t \|^2 = \frac{1}{m} \sum_{i=1}^{m} \big\| \Delta_t^i \big\|^2$.
First we have: 
\begin{align*}
\| \e_t \|^2 
&= \big\| \frac{1}{m} \sum_{i=1}^{m} \e_t^i \big\|^2 \leq \frac{1}{m} \sum_{i=1}^{m} \big\| \e_t^i \big\|^2 = \big\| \bar{\e}_t \big\|^2 
\end{align*}
That is, we have the bound of $\| \e_t \|^2$ once we can bound $\| \bar{\e}_t \|^2$.
\begin{align*}
\big\| \bar{\e}_t \big\|^2 
&= \frac{1}{m} \sum_{i=1}^{m} \big\| \e_t^i \big\|^2 \\
&= \frac{1}{m} \sum_{i=1}^{m} \big\| \P_{t-1}^i - \tilde{\Delta}_{t-1}^i \big\|^2 \\
&\overset{(a1)}{\leq}  \frac{1}{m} \sum_{i=1}^{m} (1/\gamma) \big\| \P_{t-1}^i \big\|^2 \\
&\leq \frac{1}{m} \sum_{i=1}^{m} (1/\gamma) \big\| \g_{t-1}^i + \e_{t-1}^i \big\|^2 \\
&\overset{(a2)}{\leq} \frac{1}{m} (1/\gamma) \sum_{i=1}^{m} \big[ \frac{1}{1 - \epsilon} \big\| \g_{t-1}^i \big\|^2 + \frac{1}{\epsilon} \big\| \e_{t-1}^i \big\|^2 \big] \\
&= (1/\gamma) \big[ \frac{1}{\epsilon} \frac{1}{m} \sum_{i=1}^{m} \big\| \e_{t-1}^i \big\|^2 + \frac{1}{1 - \epsilon} \frac{1}{m} \sum_{i=1}^{m} \big\| \g_{t-1}^i \big\|^2 \big] \\
&= (1/\gamma) \frac{1}{\epsilon} \big\| \bar{\e}_{t-1} \big\|^2 + (1/\gamma)\frac{1}{1 - \epsilon} \big\| \bar{\Delta}_{t-1} \big\|^2, 
\end{align*}
where $(a1)$ is due to the definition of the compressor,
$(a2)$ follows from $\| \x + \y \|^2 \leq \frac{1}{\epsilon} \| \x \|^2 + \frac{1}{1 - \epsilon} \| \y \|^2$, where $\epsilon \in (0, 1)$.

Recursively using this relationship that $\big\| \bar{\e}_{t} \big\|^2 \leq (1/\gamma) \frac{1}{\epsilon} \big\| \bar{\e}_{t-1} \big\|^2 + (1/\gamma)\frac{1}{1 - \epsilon} \big\| \bar{\Delta}_{t-1} \big\|^2$ and note $\bar{\e}_0 = 0$, then we have:
\begin{align*}
\| \bar{\e}_t \|^2 &\leq \sum_{p=0}^{t-1} [(1/\gamma) \frac{1}{\epsilon}]^{t-1-p} (1/\gamma)\frac{1}{1 - \epsilon} \| \bar{\Delta}_{p} \|^2.
\end{align*}

Summing from $t=0$ to $t = T-1$: 
\begin{align}
\sum_{t=0}^{T-1} \| \bar{\e}_t \|^2 &= [(1/\gamma)\frac{1}{1 - \epsilon} \sum_{p=0}^{T-2} ((1/\gamma) \frac{1}{\epsilon})^p \| \bar{\Delta}_{0} \|^2] \nonumber\\
&\indent + [(1/\gamma)\frac{1}{1 - \epsilon} \sum_{p=0}^{T-3} ((1/\gamma) \frac{1}{\epsilon})^p \| \bar{\Delta}_{1} \|^2] + \cdots \nonumber \\
&\indent + [(1/\gamma)\frac{1}{1 - \epsilon} \sum_{p=0}^{1} ((1/\gamma) \frac{1}{\epsilon})^p \| \bar{\Delta}_{T-3} \|^2] \nonumber\\
&\indent + [(1/\gamma)\frac{1}{1 - \epsilon} \sum_{p=0}^{0} ((1/\gamma) \frac{1}{\epsilon})^p \| \bar{\Delta}_{T-2} \|^2] \nonumber \\
&\leq (1/\gamma)\frac{1}{1 - \epsilon} \sum_{t=0}^{T-1} \sum_{p=0}^{\infty} ((1/\gamma) \frac{1}{\epsilon})^p \| \bar{\Delta}_{t} \|^2 \nonumber \\
&\overset{(a3)}{\leq} (1/\gamma)\frac{1}{1 - \epsilon} \sum_{t=0}^{T-1} \sum_{p=0}^{\infty} ((1/\gamma) \frac{1}{\epsilon})^p \alpha_t \| \Delta_{t} \|^2 \nonumber \\
&\leq (1/\gamma) \sum_{t=0}^{T-1} ( 1 + \frac{1}{\gamma \epsilon - 1}) \frac{\alpha_t}{1 - \epsilon}  \| \Delta_{t} \|^2 \nonumber \\
&\overset{(a4)}{\leq} (1/ \gamma) (1 + 1/a) b \sum_{t=0}^{T-1} \alpha_t \| \Delta_{t} \|^2 \nonumber \\
&\overset{(a5)}{=}  \sum_{t=0}^{T-1} h(\gamma, \alpha_t) \| \Delta_{t} \|^2. \label{ineq_error}
\end{align}

$(a3)$ is due to $\alpha_t = \frac{\| \bar{\Delta}_{t} \|^2}{\| \Delta_{t} \|^2}$.
For any given $\gamma > 1$, we can choose $\epsilon \in (0, 1)$ such that $\gamma \epsilon - 1 \geq a$ and $\frac{1}{1 - \epsilon} \leq b$ for two constant $a$ and $b$, which yields $(a4)$.
$(a5)$ follows from the fact that $h(\gamma, \alpha_t) = (1/ \gamma) (1 + 1/a) b \alpha_t$


This completes the proof of Lemma~\ref{lem: Bounded_error}.
\end{proof}
\begin{lem} [Iterative Step]\label{lem: Iterative_step_appdx}
By letting $\hat{\x}_{t} = \x_t + \eta \e_t$, we show the iterative relationship of $\hat{\x}_{t}$ as follows.
$$
\hat{\x}_{t+1} = \hat{\x}_t + \eta \Delta_t.
$$
\end{lem}

\begin{proof} [Proof of Lemma~\ref{lem: Iterative_step_appdx}]
We set a set of virtual variables for convenience.
Denote
$\e_t = \frac{1}{m} \sum_{i=1}^{m} \e_t^i$,
$\Delta_t = \frac{1}{m} \sum_{i=1}^{m} \g_t^i$,
$\tilde{\Delta}_t = \frac{1}{m} \sum_{i=1}^{m} \tilde{\Delta}_t^i = \frac{1}{m} \sum_{i=1}^{m} \mathcal{C}{(\P_t^i)}$,
$\hat{\x}_{t} = \x_t + \eta \e_t$.
\begin{align*}
\hat{\x}_{t+1} 
&= \x_{t+1} + \eta \e_{t+1} \\
&= \x_t + \eta \tilde{\Delta}_t + \eta \e_{t+1} \\
&= \x_t + \frac{1}{m} \eta \sum_{i=1}^{m} \tilde{\Delta}_t^i + \frac{1}{m} \eta \sum_{i=1}^{m} \e_{t+1}^i \\
&= \x_t + \frac{1}{m} \eta \sum_{i=1}^{m} \tilde{\Delta}_t^i + \frac{1}{m} \eta \sum_{i=1}^{m} (\P_t^i - \tilde{\Delta}_t^i) \\
&= \x_t + \frac{1}{m} \eta \sum_{i=1}^{m} \P_t^i \\
&= \x_t + \frac{1}{m} \eta \sum_{i=1}^{m} (\g_t^i + \e_t^i) \\
&= \x_t + \frac{1}{m} \eta \sum_{i=1}^{m} \e_t^i + \frac{1}{m} \eta \sum_{i=1}^{m} \g_t^i \\
&= \hat{\x}_t + \eta \Delta_t.
\end{align*}

This completes the proof of Lemma~\ref{lem: Iterative_step_appdx}.
\end{proof}
\begin{lem} [One Communication Round Descent]\label{lem: one_step_descent_appdx}
Choose local and global learning rates $\eta_L$ and $\eta$ as $\eta_L \leq \frac{1}{8LK}$, $\eta \eta_L < \frac{1}{K L}$ and $ \eta \eta_L K ( L^2 h(\gamma, \alpha_t) + 1 + L) \leq 1$.
Under Assumptions~\ref{a_smooth}--\ref{a_variance}, one communication round descent of the virtual sequence of outputs $\{ \hat{\x}_t = \x_t + \eta \e_t \}$ generated by our Algorithm satisfies:
\begin{align*}
&\mathbb{E} f(\hat{\x}_{t+1}) - \nabla f(\hat{\x}_t) \\
&\leq (\frac{1}{2} L^2 \eta^2 h(\gamma, \alpha_t) + \frac{1}{2} \eta^2 + \frac{L \eta^2}{2}) \frac{K \eta_L^2}{m} \sigma_L^2 \\
& \indent - c \eta \eta_L K \big\| \nabla f(x_t) \big\|^2+ \frac{5 \eta K^2 \eta_L^3 L^2}{2} (\sigma_L^2 + 6 K \sigma_G^2)
\end{align*}
where $\hat{\x}_{t} = \x_t + \eta \e_t$, $c$ is a constant, $h(\gamma, \alpha_t)$ is defined the same as that in Lemma~\ref{lem: Bounded_error}.
\end{lem}

\begin{proof} [Proof of Lemma~\ref{lem: one_step_descent_appdx}]
Due to the Smoothness assumption \ref{a_smooth}, taking expectation of $f(\hat{\x}_{t+1})$ over the randomness at time step $t$, we have:
\begin{align}
	&\mathbb{E}_t f(\hat{\x}_{t+1}) \nonumber\\
	&\leq f(\hat{\x}_t) + \mathbb{E}_t < \nabla f(\hat{\x}_t),[\hat{\x}_{t+1} - \hat{\x}_t] > + \frac{L}{2} \mathbb{E}_t \| \hat{\x}_{t+1} - \hat{\x}_t \|^2 \nonumber \\
	&= f(\hat{\x}_t) + \underbrace{\mathbb{E}_t < \nabla f(\hat{\x}_t), \eta \Delta_t > }_{A_1} + \frac{L \eta^2}{2} \underbrace{ \mathbb{E}_t \| \Delta_t \|^2 }_{A_2}. \label{ineq_compression_smooth_1}
\end{align}

We bound $A_1$ and $A_2$ respectively in the following.

\textbf{Bounding $A_1$:}
\begin{align}
& A_1 = \mathbb{E}_t \big< \nabla f(\hat{\x}_t), \eta \Delta_t \big> \nonumber \\
	&= \mathbb{E}_t \big< \nabla f(\hat{\x}_t) - \nabla f(\x_t) + \nabla f(\x_t), \eta \Delta_t \big> \nonumber\\
	&= \mathbb{E}_t \big[ \big< \nabla f(\hat{\x}_t) - \nabla f(\x_t), \eta \Delta_t \big> + \eta \big< \nabla f(\x_t), \Delta_t \big> \big] \nonumber\\
	&\overset{(b1)}{\leq} \mathbb{E}_t \big[\frac{1}{2} (\| \nabla f(\hat{\x}_t) - \nabla f(\x_t) \|^2 + \eta^2 \| \Delta_t \|^2) \nonumber \\
	&\indent \indent + \eta \big< \nabla f(\x_t), \Delta_t + K \eta_L \nabla f(\x_t) - K \eta_L \nabla f(\x_t) \big>\big] \nonumber\\
	&\overset{(b2)}{\leq} \mathbb{E}_t \big[ \frac{1}{2} ( L^2 \| \hat{\x}_t - \x_t \|^2 + \eta^2 \| \Delta_t \|^2) \nonumber\\
	& \indent \indent + \eta \big< \nabla f(\x_t), \Delta_t + K \eta_L \nabla f(\x_t) \big> - K \eta \eta_L \| \nabla f(\x_t) \|^2 \big] \nonumber\\
	&= \mathbb{E}_t \big[ \frac{1}{2} L^2 \eta^2 \| \e_t \|^2 + \frac{1}{2} \eta^2 \| \Delta_t \|^2 - K \eta \eta_L \| \nabla f(\x_t) \|^2 \nonumber\\
	&\indent \indent + \eta \underbrace{ \big< \nabla f(\x_t), \Delta_t + K \eta_L \nabla f(\x_t) \big> }_{A_3} \big], \label{ineq_A1_1}
\end{align}
where $(b1)$ follows from the fact that $\big<\x, \y\big> = \frac{1}{2} [ \| \x \|^2 + \| \y \|^2 - \| \x - \y \|^2] \leq \frac{1}{2} [ \| \x \|^2 + \| \y \|^2 ]$, 
$(b2)$ is due to the Assumption~\ref{a_smooth}.

We can bound $A_3$ as follows.
\begin{align}
& A_3 = \mathbb{E}_t \big< \nabla f(\x_t), \Delta_t + \eta_L K \nabla f(\x_t) \big> \nonumber \\
&= \big< \nabla f(\x_t), \mathbb{E}_t \big[- \frac{1}{m}\sum_{i = 1}^m \sum_{k=0}^{K-1} \eta_L \g_{t,k}^i + \eta_L K \nabla f(x_t) \big] \big> \nonumber \\
& = \big< \nabla f(\x_t),  \mathbb{E}_t  \big[- \frac{1}{m} \sum_{i = 1}^m  \sum_{k=0}^{K-1} \eta_L \nabla F_i(\x_{t,k}^i) + \eta_L K \frac{1}{m}\sum_{i = 1}^m  \nabla F_i(\x_t)\big] \big> \nonumber \\
&=  \big< \sqrt{\eta_L K} \nabla f(\x_t), - \frac{\sqrt{\eta_L}}{m \sqrt{K}} \mathbb{E}_t \sum_{i = 1}^m \sum_{k=0}^{K-1} ( \nabla F_i(\x_{t,k}^i) - \nabla F_i(\x_t)) \big> \nonumber \\
&\overset{(c1)}{=} \frac{\eta_L K}{2} \big\| \nabla f(\x_t) \big\|^2 - \frac{\eta_L}{2K m^2} \big\| \sum_{i=1}^{m} \sum_{k=0}^{K-1} \nabla F_i(\x_{t,k}^i) \big\|^2 ] \nonumber \\
& \indent \indent + \frac{\eta_L }{2K m^2} \mathbb{E}_t \big[ \big\| \sum_{i = 1}^m \sum_{k=0}^{K-1} ( \nabla F_i(\x_{t,k}^i) - \nabla F_i(\x_t)) \big\|^2 \big] \nonumber \\
&\overset{(c2)}{\leq} \frac{\eta_L K}{2} \big\| \nabla f(\x_t) \big\|^2 - \frac{\eta_L}{2K m^2} \big\| \sum_{i=1}^{m} \sum_{k=0}^{K-1} \nabla F_i(\x_{t,k}^i) \big\|^2 \big] \nonumber \\
& \indent \indent + \frac{\eta_L}{2m} \mathbb{E}_t \big[ \sum_{i = 1}^m \sum_{k=0}^{K-1} \big\| \nabla F_i(\x_{t,k}^i) - \nabla F_i(\x_t) \big\|^2 \nonumber \\
&\overset{(c3)}{\leq} \frac{\eta_L K}{2} \big\| \nabla f(\x_t) \big\|^2 - \frac{\eta_L}{2K m^2} \mathbb{E}_t \big\| \sum_{i=1}^{m} \sum_{k=0}^{K-1} \nabla F_i(\x_{t,k}^i) \big\|^2 \nonumber \\
& \indent \indent + \frac{\eta_L L^2}{2m} \mathbb{E}_t \sum_{i = 1}^m \sum_{k=0}^{K-1}  \big\|  \x_{t,k}^i - \x_t \big\|^2 \nonumber \\
&\overset{(c4)}{\leq} \eta_L K (\frac{1}{2} + 15 K^2 \eta_L^2 L^2) \big\| \nabla f(x_t) \big\|^2 + \frac{5 K^2 \eta_L^3 L^2}{2} (\sigma_L^2 + 6 K \sigma_G^2) \nonumber \\
& \indent \indent - \frac{\eta_L}{2K m^2} \mathbb{E}_t \big\| \sum_{i=1}^{m} \sum_{k=0}^{K-1} \nabla F_i(\x_{t,k}^i) \big\|^2, \label{ineq_A1_2}
\end{align}
where $(c1)$ follows from the fact that $\big<\x, \y \big> = \frac{1}{2} [ \| \x \|^2 + \| \y \|^2 - \| \x - \y \|^2 ]$  for $\x = \sqrt{\eta_L K} \nabla f(\x_t)$ and $\y = - \frac{\sqrt{\eta_L}}{m \sqrt{K}} \sum_{i = 1}^m \sum_{k=0}^{K-1} ( \nabla F_i(\x_{t,k}^i) - \nabla F_i(\x_t))$,
$(c2)$ is due to that $\mathbb{E}[\|x_1 + \cdots + x_n \|^2] \leq n \mathbb{E}[\|x_1\|^2 + \cdots + \| x_n \|^2]$ ,
$(c3)$ is due to Assumption~\ref{a_smooth} 
and $(c4)$ follows from Lemma~\ref{lem: aux_bounded_x_t}.

Plugging the above results ~\ref{ineq_A1_2} into ~\ref{ineq_A1_1} yields:
\begin{align}
& A_1 \leq \mathbb{E}_t \big[ \frac{1}{2} L^2 \eta^2 \big\| \e_t \big\|^2 + \frac{1}{2} \eta^2 \big\| \Delta_t \big\|^2 \nonumber\\
& \indent \indent + \eta \big< \nabla f(\x_t), \Delta_t + K \eta_L \nabla f(\x_t) \big> - K \eta \eta_L \big\| \nabla f(\x_t) \big\|^2 \big] \nonumber \\
&\leq \mathbb{E}_t \big[ \frac{1}{2} L^2 \eta^2 \big\| \e_t \|^2 + \frac{1}{2} \eta^2 \big\| \Delta_t \|^2 - K \eta \eta_L  \big\| \nabla f(\x_t) \big\|^2 \big] \nonumber\\
& \indent \indent + \eta \eta_L K (\frac{1}{2} + 15 K^2 \eta_L^2 L^2) \big\| \nabla f(x_t) \big\|^2 \nonumber \\
& \indent \indent + \frac{5 \eta K^2 \eta_L^3 L^2}{2} (\sigma_L^2 + 6 K \sigma_G^2) \nonumber \\
& \indent \indent - \frac{\eta \eta_L}{2K m^2}  \mathbb{E}_t \big\| \sum_{i=1}^{m} \sum_{k=0}^{K-1} \nabla F_i(\x_{t,k}^i) \big\|^2 \nonumber \\
&\overset{(*)}{\leq} \mathbb{E}_t \big[ (\frac{1}{2} L^2 \eta^2 h(\gamma, \alpha_t)  + \frac{1}{2} \eta^2) \big\| \Delta_t \big\|^2 \big] - K \eta \eta_L  \big\| \nabla f(\x_t) \big\|^2 \nonumber\\
& \indent \indent + \eta \eta_L K (\frac{1}{2} + 15 K^2 \eta_L^2 L^2) \big\| \nabla f(x_t) \big\|^2 \nonumber \\
& \indent \indent + \frac{5 \eta K^2 \eta_L^3 L^2}{2} (\sigma_L^2 + 6 K \sigma_G^2) \nonumber \\
& \indent \indent - \frac{\eta \eta_L}{2K m^2} \mathbb{E}_t \big\| \sum_{i=1}^{m} \sum_{k=0}^{K-1} \nabla F_i(\x_{t,k}^i) \big\|^2. \label{ineq_A1_appdx}
\end{align}
Note that Lemma~\ref{lem: Bounded_error} does not necessarily leads to $\| \e_t \|^2 \leq h(\gamma, \alpha_t) \times \\ \big\| \Delta_t \|^2$ in general for $(*)$ to hold rigorously.
We abuse the notation and inequality here in order to show the function descent in one communication round more clearly.
In the processes of proving convergence in the following different cases, the focus is the summation $\sum_{t=0}^{T-1} \| \e_t \| ^2$ since a common approach is to sum the function descent from 0 to $T-1$.
This gives $\sum_{t=0}^{T-1} \| \e_t \|^2  \leq \sum_{t=0}^{T-1} h(\gamma, \alpha_t) \| \Delta_{t} \|^2$, thus the convergence rate results proven in the following rigorously hold.

\textbf{Bounding $A_2$:}
\begin{align}
&A_2 = \mathbb{E}_t \big[ \big\| \Delta_t \big\|^2 \big] \nonumber \\
&= \mathbb{E}_t \big[ \big\| \frac{1}{m} \sum_{i=1}^{m} \Delta_t^i \big\|^2 \big] \nonumber \\
&\leq \frac{1}{m^2} \mathbb{E}_t \big[ \big\| \sum_{i=1}^{m} \Delta_t^i \big\|^2 \big] \nonumber \\
&= \frac{\eta_L^2}{m^2} \mathbb{E}_t \big[ \big\| \sum_{i=1}^{m} \sum_{k=0}^{K-1} \g_{t,k}^i \big\|^2 \big] \nonumber \\
&\overset{(d1)}{=} \frac{\eta_L^2}{m^2} \mathbb{E}_t \big[ \big\| \sum_{i=1}^{m} \sum_{k=0}^{K-1} (\g_{t,k}^i - \nabla F_i(\x_{t,k}^i)) \big\|^2 \big] \nonumber \\
&\indent \indent + \frac{\eta_L^2}{m^2} \big\| \sum_{i=1}^{m} \sum_{k=0}^{K-1} \nabla F_i(\x_{t,k}^i) \big\|^2 \nonumber \\
&\overset{(d2)}{\leq} \frac{K \eta_L^2}{m} \sigma_L^2 + \frac{\eta_L^2}{m^2} \big\| \sum_{i=1}^{m} \sum_{k=0}^{K-1} \nabla F_i(\x_{t,{}k}^i) \big\|^2, \label{ineq_A2_appdx}
\end{align}
where $(d1)$ follows from the fact that $\mathbb{E}[\| \x \|^2] = \mathbb{E}[\| \x - \mathbb{E}[\x] \|^2] + \| \mathbb{E}[\x] \|^2$, $(d2)$ is due to the bounded variance assumption in Assumption~\ref{a_variance} and the fact that $\mathbb{E}[\|x_1 + \cdots + x_n \|^2] = \mathbb{E}[\|x_1\|^2 + \cdots + \| x_n \|^2]$ if $x_i$ are independent with mean $0$.

By combining the results in \eqref{ineq_A1_appdx}, \eqref{ineq_A2_appdx} and \eqref{ineq_compression_smooth_1}, we have:
\begin{align*}
&\mathbb{E} f(\hat{\x}_{t+1}) - \nabla f(\hat{\x}_t) \\
&\leq \underbrace{\mathbb{E} \big< \nabla f(\hat{\x}_t), \eta \Delta_t \big> }_{A_1} + \frac{L \eta^2}{2} \underbrace{ \mathbb{E} \big\| \Delta_t \big\|^2 }_{A_2}  \\
&\overset{(e1)}{\leq} \big[ (\frac{1}{2} L^2 \eta^2 h(\gamma, \alpha_t) + \frac{1}{2} \eta^2 + \frac{L \eta^2}{2}) \times \\
& \indent \indent (\frac{K \eta_L^2}{m} \sigma_L^2 + \frac{\eta_L^2}{m^2} \| \sum_{i=1}^{m} \sum_{k=0}^{K-1} \nabla F_i(\x_{t,{}k}^i) \|^2) \big] \\
&\indent \indent+ \eta \eta_L K (\frac{1}{2} + 15 K^2 \eta_L^2 L^2) \big\| \nabla f(x_t) \big\|^2 \\
&\indent \indent+ \frac{5 \eta K^2 \eta_L^3 L^2}{2} (\sigma_L^2 + 6 K \sigma_G^2) - K \eta \eta_L \big\| \nabla f(\x_t) \big\|^2  \\
&\indent \indent- \frac{\eta \eta_L}{2K m^2} \big\| \sum_{i=1}^{m} \sum_{k=0}^{K-1} \nabla F_i(\x_{t,k}^i) \big\|^2 \\
&\overset{(e2)}{\leq} (\frac{1}{2} L^2 \eta^2 h(\gamma, \alpha_t) + \frac{1}{2} \eta^2 + \frac{L \eta^2}{2}) \frac{K \eta_L^2}{m} \sigma_L^2 \\
&\indent \indent- \eta \eta_L K (\frac{1}{2} - 15 K^2 \eta_L^2 L^2) \big\| \nabla f(x_t) \big\|^2 \\
&\indent \indent+ \frac{5 \eta K^2 \eta_L^3 L^2}{2} (\sigma_L^2 + 6 K \sigma_G^2)  \\
&\overset{(e3)}{\leq} (\frac{1}{2} L^2 \eta^2 h(\gamma, \alpha_t) + \frac{1}{2} \eta^2 + \frac{L \eta^2}{2}) \frac{K \eta_L^2}{m} \sigma_L^2 \\
&\indent \indent- c \eta \eta_L K \big\| \nabla f(x_t) \big\|^2+ \frac{5 \eta K^2 \eta_L^3 L^2}{2} (\sigma_L^2 + 6 K \sigma_G^2) .
\end{align*}
$(e1)$ is due to the inequality~\eqref{ineq_A1_appdx} and ~\eqref{ineq_A2_appdx}.
$(e2)$ holds if $(\frac{1}{2} L^2 \eta^2 h(\gamma, \alpha_t) + \frac{1}{2} \eta^2 + \frac{L \eta^2}{2}) \frac{\eta_L^2}{m^2} - \frac{\eta \eta_L}{2K m^2} \leq 0$, that is, 
$ \eta \eta_L K ( L^2 h(\gamma, \alpha_t) + 1 + L) \leq 1$.
$(e3)$ holds for a constant $c$ such that $(\frac{1}{2} - 15 K^2 \eta_L^2 L^2) > c > 0$ if $\eta_L < \frac{1}{\sqrt{30} K L}$.

This completes the proof of Lemma~\ref{lem: one_step_descent_appdx}.
\end{proof}

\subsection{Proof of Theorem~\ref{thm:FedAvg_compression}}
\FedAvg*

\begin{proof} [Proof of Theorem~\ref{thm:FedAvg_compression}]
Summing from $t=0, \cdots, T-1$ of the one communication round descent in Lemma~\ref{lem: one_step_descent_appdx} yields: 
\begin{align*}
&\mathbb{E} f(\hat{\x}_{T}) - \nabla f(\hat{\x}_0) \\
&\leq \sum_{t=0}^{T-1} \big\{ (\frac{1}{2} L^2 \eta^2 h(\gamma, \alpha_t) + \frac{1}{2} \eta^2 + \frac{L \eta^2}{2}) \frac{K \eta_L^2}{m} \sigma_L^2 \\
&\indent - c \eta \eta_L K \big\| \nabla f(x_t) \big\|^2+ \frac{5 \eta K^2 \eta_L^3 L^2}{2} (\sigma_L^2 + 6 K \sigma_G^2) \big\} \\
&= \sum_{t=0}^{T-1} \big\{ (\frac{1}{2} L^2 \eta^2 h(\gamma, \alpha) + \frac{1}{2} \eta^2 + \frac{L \eta^2}{2}) \frac{K \eta_L^2}{m} \sigma_L^2 \\
&\indent - c \eta \eta_L K \big\| \nabla f(x_t) \big\|^2+ \frac{5 \eta K^2 \eta_L^3 L^2}{2} (\sigma_L^2 + 6 K \sigma_G^2) \big\},
\end{align*}
where $h(\gamma, \alpha) = \frac{1}{T} \sum_{t=0}^{T-1} h(\gamma, \alpha_t)$.

By rearranging, we have:
\begin{align*}
&\sum_{t=0}^{T-1} c \eta \eta_L K \| \nabla f(\x_t) \|^2 \\
&\leq \nabla f(\hat{\x}_0) - \mathbb{E} f(\hat{\x}_{T}) + T (\eta \eta_L K) \bigg[ \frac{5 K \eta_L^2 L^2}{2} (\sigma_L^2 + 6 K \sigma_G^2) \\
&\indent + (\frac{1}{2} L^2 \eta h(\gamma, \alpha) + \frac{1}{2} \eta + \frac{L \eta}{2}) \frac{\eta_L}{m} \sigma_L^2 \bigg].
\end{align*}

By letting $f_0 = f(\hat{\x}_{0})$ and $f_{*} = f(\x_{*}) \leq f(\hat{\x}_T)$, we have:
$$
	\min_{t \in [T]} \mathbb{E}\|\nabla f(\x_t)\|_2^2 \leq \frac{f_0 - f_{*}}{c \eta \eta_L K T} + \Phi,
$$
where $\Phi \triangleq \frac{1}{c} [(\frac{1}{2} L^2 h(\gamma, \alpha) + \frac{1}{2} + \frac{L}{2}) \frac{\eta \eta_L}{m} \sigma_L^2 + \frac{5 K \eta_L^2 L^2}{2} (\sigma_L^2 + 6 K \sigma_G^2)]$.

This completes the proof of Theorem~\ref{thm:FedAvg_compression}.
\end{proof}

\FedAvgC*

\begin{proof} [Proof of Corollary~\ref{cor:FedAvg_compression}]
Plugging the learning rate into the convergence rate bound in Theorem~\ref{thm:FedAvg_compression} completes the proof.
\end{proof}

\subsection{Proof of Theorem~\ref{thm:FedAvg_decay}}
\FedAvgDecaying*

\begin{proof} [Proof of Theorem~\ref{thm:FedAvg_decay}]
Note that Lemma~\ref{lem: Bounded_error} and Lemma~\ref{lem: Iterative_step_appdx} still hold with decaying local learning rate.
Thus the one communication round descent is as following:
\begin{align*}
&\mathbb{E} f(\hat{\x}_{t+1}) - \nabla f(\hat{\x}_t) \\
&\leq (\frac{1}{2} L^2 \eta^2 h(\gamma, \alpha_t) + \frac{1}{2} \eta^2 + \frac{L \eta^2}{2}) \frac{K \eta_{L, t}^2}{m} \sigma_L^2 \\
& \indent- c \eta \eta_{L, t} K \big\| \nabla f(x_t) \big\|^2+ \frac{5 \eta K^2 \eta_{L, t}^3 L^2}{2} (\sigma_L^2 + 6 K \sigma_G^2),
\end{align*}
where $\hat{\x}_{t} = \x_t + \eta \e_t$, $c$ is a constant, $h(\gamma, \alpha_t)$ is defined the same as that in Lemma~\ref{lem: Bounded_error}.

By telescoping the above result, we have: 
\begin{align*}
&\mathbb{E} f(\hat{\x}_{T}) - \nabla f(\hat{\x}_0) \\
&\leq \sum_{t=0}^{T-1} \big\{ (\frac{1}{2} L^2 \eta^2 h(\gamma, \alpha_t) + \frac{1}{2} \eta^2 + \frac{L \eta^2}{2}) \frac{K \eta_{L, t}^2}{m} \sigma_L^2 \\
&\indent - c \eta \eta_{L, t} K \big\| \nabla f(x_t) \big\|^2+ \frac{5 \eta K^2 \eta_{L, t}^3 L^2}{2} (\sigma_L^2 + 6 K \sigma_G^2) \big\}.
\end{align*}

Rearranging the terms:
\begin{align*}
&\sum_{t=0}^{T-1} \eta_{L, t} \big\| \nabla f(\x_t) \big\|^2 \\
&\leq \frac{\mathbb{E} f(\hat{\x}_{0}) - \nabla f(\hat{\x}_T)}{c \eta K} + \sum_{t=0}^{T-1} \big\{ (\frac{1}{2} L^2 h(\gamma, \alpha_t) + \frac{1}{2} + \frac{L}{2}) \frac{\eta\eta_{L, t}^2}{c m} \sigma_L^2 \\
&\indent + \frac{5 K \eta_{L, t}^3 L^2}{2 c} (\sigma_L^2 + 6 K \sigma_G^2) \big\}.
\end{align*}

Let $H_T = \sum_{t=0}^{T-1} \eta_{L, t}$ and $\z$ is sampled from $\{ \x_t \}, \forall t \in [T]$ with probability $\mathbb{P}[\z = \x_t] = \frac{\eta_{L, t}}{H_T}$ which results in $\mathbb{E} \big\| \nabla f(\z) \big\|^2 = \frac{1}{H_T} \sum_{t=0}^{T-1} \eta_{L, t} \big\| \nabla f(\x_t) \big\|^2$.
That is:
\begin{align*}
&\mathbb{E} \big\| \nabla f(\z) \big\|^2 \\
&\leq \frac{\mathbb{E} f(\hat{\x}_{0}) - \nabla f(\hat{\x}_T)}{c \eta K H_T} + (\frac{1}{2} L^2 h(\gamma, \alpha) + \frac{1}{2} + \frac{L}{2}) \frac{\eta}{c m H_T} \sigma_L^2 \sum_{t=0}^{T-1} \eta_{L, t}^2 \\
&\indent + \frac{5 K L^2}{2 c H_T} (\sigma_L^2 + 6 K \sigma_G^2) \sum_{t=0}^{T-1} \eta_{L, t}^3,
\end{align*}
where $h(\gamma, \alpha) = \frac{1}{T} \sum_{t=0}^{T-1} h(\gamma, \alpha_t)$.

By letting $f_0 = f(\hat{\x}_{0})$ and $f_{*} = f(\x_{*}) \leq f(\hat{\x}_T)$, we complete the proof of Theorem~\ref{thm:FedAvg_decay}.
\end{proof}

\FedAvgDecayingC*

\begin{proof} [Proof of Corollary~\ref{cor:FedAvg_compression_decaying}]
\begin{align*}
&H_T = \sum_{t=0}^{T-1} \eta_{L, t} = \sum_{t=0}^{T-1} \frac{1}{\sqrt{t+a}KL} = \frac{1}{KL} \Theta(T^{1/2}). \\
&\sum_{t=0}^{T-1} \eta_{L, t}^2 = \sum_{t=0}^{T-1} (\frac{1}{\sqrt{t+a}KL})^2 = \frac{1}{K^2L^2} \mathcal{O}(\ln(T)) \\
&\sum_{t=0}^{T-1} \eta_{L, t}^3 = \sum_{t=0}^{T-1} (\frac{1}{\sqrt{t+a}KL})^3 = \frac{1}{K^3L^3} \mathcal{O}(1) \\
\end{align*}
So we have:
\begin{align*}
\mathbb{E} \big\| \nabla f(\z) \big\|^2 
&= \mathcal{O}(\frac{1}{\sqrt{mKT}} \ln(T)) + \mathcal{O}(\frac{1}{\sqrt{T}}) \\
&= \mathcal{\tilde{O}}( \frac{1}{\sqrt{mKT}}) + \mathcal{O}(\frac{1}{\sqrt{T}}).
\end{align*}
\end{proof}

\subsection{Proof of Theorem~\ref{thm:FedAvg_dynamic}}
\FedAvgDistinct*

\begin{proof} [Proof of Theorem~\ref{thm:FedAvg_dynamic}]
First, note that Lemma~\ref{lem: Bounded_error} still holds for distinct local steps among different workers.
With gradients scaled by the local steps $K_i$, $\Delta_t$ is:
$$
\Delta_t = \frac{1}{m} \sum_{i=1}^{m} \Delta_t^i 
= \sum_{i=1}^{m} \frac{1}{K_i} \sum_{k=0}^{K_i-1} \g_{t,k}^i
$$

Due to the Smoothness assumption \ref{a_smooth}, taking expectation of $f(\hat{\x}_{t+1})$ over the randomness at time step $t$, we have:
\begin{align}
	&\mathbb{E}_t f(\hat{\x}_{t+1}) \nonumber\\
	&\leq f(\hat{\x}_t) + \mathbb{E}_t < \nabla f(\hat{\x}_t),[\hat{\x}_{t+1} - \hat{\x}_t] > + \frac{L}{2} \mathbb{E}_t \| \hat{\x}_{t+1} - \hat{\x}_t \|^2 \nonumber \\
	&= f(\hat{\x}_t) + \underbrace{\mathbb{E}_t < \nabla f(\hat{\x}_t), \eta \Delta_t > }_{A_1} + \frac{L \eta^2}{2} \underbrace{ \mathbb{E}_t \| \Delta_t \|^2 }_{A_2}. \label{ineq_compression_smooth_dynamic}
\end{align}

\textbf{Bounding $A_2$:}
\begin{align}
&A_2 = \mathbb{E}_t \big[ \big\| \Delta_t \big\|^2 \big] \nonumber \\
&= \mathbb{E}_t \big[ \big\| \frac{1}{m} \sum_{i=1}^{m} \Delta_t^i \big\|^2 \big] \nonumber \\
&\leq \frac{1}{m^2} \mathbb{E}_t \big[ \big\| \sum_{i=1}^{m} \Delta_t^i \big\|^2 \big] \nonumber \\
&= \frac{\eta_L^2}{m^2} \mathbb{E}_t \big[ \big\| \sum_{i=1}^{m} \frac{1}{K_i} \sum_{k=0}^{K_i-1} \g_{t,k}^i \big\|^2 \big] \nonumber \\
&\overset{(d1)}{=} \frac{\eta_L^2}{m^2} \mathbb{E}_t \big[ \big\| \sum_{i=1}^{m} \frac{1}{K_i} \sum_{k=0}^{K_i-1} (\g_{t,k}^i - \nabla F_i(\x_{t,k}^i)) \big\|^2 \big] \nonumber \\
&\indent \indent + \frac{\eta_L^2}{m^2} \big\| \sum_{i=1}^{m} \frac{1}{K_i} \sum_{k=0}^{K_i-1} \nabla F_i(\x_{t,k}^i) \big\|^2 \nonumber \\
&\overset{(d2)}{\leq} \frac{\eta_L^2 }{m^2} \sum_{i = 1}^{m} \frac{1}{K_i} \sigma_L^2 + \frac{\eta_L^2}{m^2} \big\| \sum_{i=1}^{m} \frac{1}{K_i} \sum_{k=0}^{K_i-1} \nabla F_i(\x_{t,{}k}^i) \big\|^2, \label{ineq_A2_dynamic}
\end{align}
where $(d1)$ follows from the fact that $\mathbb{E}[\| \x \|^2] = \mathbb{E}[\| \x - \mathbb{E}[\x] \|^2] + \| \mathbb{E}[\x] \|^2]$, $(d2)$ is due to the bounded variance assumption in Assumption~\ref{a_variance} and the fact that $\mathbb{E}[\|x_1 + \cdots + x_n \|^2] = \mathbb{E}[\|x_1\|^2 + \cdots + \| x_n \|^2]$ if $x_i$ are independent with mean $0$.

\textbf{Bounding $A_1$:}
We can have the same result as in ~\ref{ineq_A1_1}.
\begin{align}
&A_1 = \mathbb{E}_t \big< \nabla f(\hat{\x}_t), \eta \Delta_t \big> \nonumber \\
	&= \mathbb{E}_t \big[ \frac{1}{2} L^2 \eta^2 \| \e_t \|^2 + \frac{1}{2} \eta^2 \| \Delta_t \|^2 - \eta \eta_L \| \nabla f(\x_t) \|^2 \nonumber\\
	&\indent \indent + \eta \underbrace{ < \nabla f(\x_t), \Delta_t + \eta_L \nabla f(\x_t)> }_{A_3} \big], \label{ineq_A1_1_dynamic}
\end{align}

We can bound $A_3$ as follows.
\begin{align}
& A_3 = \mathbb{E}_t \big< \nabla f(\x_t), \Delta_t + \eta_L \nabla f(\x_t) \big> \nonumber \\
&= \big< \nabla f(\x_t), \mathbb{E}_t \big[- \frac{1}{m}\sum_{i = 1}^m \frac{1}{K_i} \sum_{k=0}^{K_i-1} \eta_L \g_{t,k}^i + \eta_L \nabla f(x_t) \big] \big> \nonumber \\
& = \big< \nabla f(\x_t),  \mathbb{E}_t  \big[- \frac{1}{m} \sum_{i = 1}^m \frac{1}{K_i} \sum_{k=0}^{K_i-1} \eta_L \nabla F_i(\x_{t,k}^i) + \eta_L \frac{1}{m}\sum_{i = 1}^m  \nabla F_i(\x_t)\big] \big> \nonumber \\
&= \big< \sqrt{\eta_L} \nabla f(\x_t), - \frac{\sqrt{\eta_L}}{m} \mathbb{E}_t \sum_{i = 1}^m \frac{1}{K_i} \sum_{k=0}^{K_i-1} ( \nabla F_i(\x_{t,k}^i) - \nabla F_i(\x_t)) \big> \nonumber \\
&\overset{(c1)}{=} \frac{\eta_L}{2} \big\| \nabla f(\x_t) \big\|^2 - \frac{\eta_L}{2 m^2} \mathbb{E}_t \big\| \sum_{i=1}^{m} \frac{1}{K_i} \sum_{k=0}^{K_i-1} \nabla F_i(\x_{t,k}^i) \big\|^2 ] \nonumber \\
& \indent \indent + \frac{\eta_L }{2 m^2} \mathbb{E}_t \big[ \big\| \sum_{i = 1}^m \frac{1}{K_i} \sum_{k=0}^{K_i-1} ( \nabla F_i(\x_{t,k}^i) - \nabla F_i(\x_t)) \big\|^2 \big] \nonumber \\
&\overset{(c2)}{\leq} \frac{\eta_L}{2m} \mathbb{E}_t \big[ \sum_{i = 1}^m \frac{1}{K_i} \sum_{k=0}^{K-1} \big\| \nabla F_i(\x_{t,k}^i) - \nabla F_i(\x_t) \big\|^2 \big] \nonumber \\
& \indent \indent +\frac{\eta_L}{2} \big\| \nabla f(\x_t) \big\|^2 - \frac{\eta_L}{2 m^2} \mathbb{E}_t \big\| \sum_{i=1}^{m} \frac{1}{K_i} \sum_{k=0}^{K_i-1} \nabla F_i(\x_{t,k}^i) \big\|^2  \nonumber \\
&\overset{(c3)}{\leq} \frac{\eta_L}{2} \big\| \nabla f(\x_t) \big\|^2 + \frac{\eta_L L^2}{2m} \mathbb{E}_t \sum_{i = 1}^m \frac{1}{K_i} \sum_{k=0}^{K_i-1}  \big\|  \x_{t,k}^i - \x_t \big\|^2 \nonumber \\
&- \frac{\eta_L}{2 m^2} \mathbb{E}_t \big\| \sum_{i=1}^{m} \frac{1}{K_i} \sum_{k=0}^{K_i-1} \nabla F_i(\x_{t,k}^i) \big\|^2 \nonumber \\
&\overset{(c4)}{\leq} \eta_L (\frac{1}{2} + 15 \eta_L^2 L^2 \frac{1}{m} \sum_{i=1}^{m} K_i^2) \big\| \nabla f(x_t) \big\|^2 \nonumber\\
&\indent \indent + \frac{5 \eta_L^3 L^2}{2} \frac{1}{m} \sum_{i=1}^{m} K_i (\sigma_L^2 + 6 K_i \sigma_G^2) \nonumber \\
& \indent \indent - \frac{\eta_L}{2 m^2} \mathbb{E}_t \big\| \sum_{i=1}^{m} \frac{1}{K_i} \sum_{k=0}^{K_i-1} \nabla F_i(\x_{t,k}^i) \big\|^2, \label{ineq_A1_2_dynamic}
\end{align}
where $(c1)$ follows from the fact that $\big<\x, \y \big> = \frac{1}{2} [ \| \x \|^2 + \| \y \|^2 - \| \x - \y \|^2 ]$,
$(c2)$ is due to that $\mathbb{E}[\|x_1 + \cdots + x_n \|^2] \leq n \mathbb{E}[\|x_1\|^2 + \cdots + \| x_n \|^2]$ ,
$(c3)$ is due to Assumption~\ref{a_smooth} 
and $(c4)$ follows from Lemma~\ref{lem: aux_bounded_x_t} (Here for each worker $i$, $\mathbb{E} \| \x_{t, k}^i - \x_t \|^2 \leq 5 K_i \eta_L^2 (\sigma_L^2 + 6 K_i \sigma_G^2) + 30 K_i^2 \eta_L^2 \| \nabla f(\x_t) \|^2$).

Following the same path for proof in Theorem~\ref{thm:FedAvg_compression}, the following holds with $\eta \eta_L (L^2 h(\gamma, \alpha_t) + 1 + L) \leq 1$:
\begin{align*}
&\sum_{t=0}^{T-1} c \eta \eta_L \| \nabla f(\x_t) \|^2 \\
&\leq \nabla f(\hat{\x}_0) - \mathbb{E} f(\hat{\x}_{T}) \\
&\indent + T (\eta \eta_L) \bigg[ \frac{5 \eta_L^2 L^2}{2} \frac{1}{m} \sum_{i=1}^{m} K_i (\sigma_L^2 + 6 K_i \sigma_G^2) \\
& \indent + (\frac{1}{2} L^2 h(\gamma, \alpha) + \frac{1}{2} + \frac{L}{2}) \frac{\eta \eta_L}{m^2} \sum_{i = 1}^{m} \frac{1}{K_i} \sigma_L^2 \bigg],
\end{align*}
where $c$ is a constant such that $(\frac{1}{2} - 15 \eta_L^2 L^2 \frac{1}{m} \sum_{i=1}^{m} K_i^2) > c > 0$ if $\eta_L < \frac{1}{\sqrt{30} K_i L}$.

By letting $f_0 = f(\hat{\x}_{0})$ and $f_{*} = f(\x_{*}) \leq f(\hat{\x}_T)$, we have:
$$
	\min_{t \in [T]} \mathbb{E}\|\nabla f(\x_t)\|_2^2 \leq \frac{f_0 - f_{*}}{c \eta \eta_L T} + \Phi,
$$
where $\Phi \triangleq \frac{1}{c} [(\frac{1}{2} L^2 h(\gamma, \alpha) + \frac{1}{2} + \frac{L}{2}) \frac{\eta \eta_L}{m^2} \sum_{i = 1}^{m} \frac{1}{K_i} \sigma_L^2 + \frac{5 \eta_L^2 L^2}{2} \frac{1}{m} \sum_{i=1}^{m} K_i (\sigma_L^2 + 6 K_i \sigma_G^2)]$.

This completes the proof of Theorem~\ref{thm:FedAvg_dynamic}.
\end{proof}

\FedAvgDistinctC*

\begin{proof} [Proof of Corollary~\ref{cor:FedAvg_compression_dynamic}]
\begin{align*}
\frac{\eta \eta_L}{m^2} \sum_{i = 1}^{m} \frac{1}{K_i} &= \frac{\sqrt{K_{min} m}}{m^2 \sqrt{T}L} \sum_{i = 1}^{m} \frac{1}{K_i} \\
&= \frac{\sqrt{K_{min}}}{\sqrt{mT}L} \frac{1}{m} \sum_{i = 1}^{m} \frac{1}{K_i} \\
&\leq \frac{\sqrt{K_{min}}}{\sqrt{mT}L} \frac{1}{K_{min}} \\
&= \frac{1}{\sqrt{mK_{min}T}L} \\
\eta_L^2 \frac{1}{m} \sum_{i = 1}^{m} K_i^2 &= \frac{1}{TL^2} \frac{1}{m} \sum_{i = 1}^{m} K_i^2 \\
&\leq \frac{K_{max}^2}{TL^2}.
\end{align*}

\end{proof}

\subsection{Auxiliary Lemma}

\begin{lem} \label{lem: aux_bounded_x_t}
[Lemma 4 in \cite{reddi2020adaptive}]
For any step-size satisfying $\eta_L \leq \frac{1}{8LK}$, we can have the following results:
$$\mathbb{E} \| \x_{t, k}^i - \x_t \|^2 \leq 5 K \eta_L^2 (\sigma_L^2 + 6 K \sigma_G^2) + 30 K^2 \eta_L^2 \| \nabla f(\x_t) \|^2$$
\end{lem}

{\em Proof of Lemma.} \\
For the completeness of the proof, we rewrite the proof of this lemma in \cite{reddi2020adaptive}.

For any worker $i \in [m]$ and $k \in [K]$, we have:
\begin{align*}
& \mathbb{E} \|\x_{t, k}^i - \x_t \|^2 = \mathbb{E} \|\x_{t, k-1}^i - \x_t -\eta_L g_{t, k-1}^t \|^2 \\
&\leq \mathbb{E} \|\x_{t, k-1}^i - \x_t -\eta_L (g_{t, k-1}^t - \nabla F_i(\x_{t, k-1}^i) \\
&\indent + \nabla F_i(\x_{t, k-1}^i) - \nabla F_i(\x_{t}) + \nabla F_i(\x_{t}) - \nabla f(\x_t) + \nabla f(\x_t)) \|^2 \\
&\leq (1 + \frac{1}{2K-1}) \mathbb{E} \| \x_{t, k-1}^i - \x_t \|^2 + \mathbb{E} \| \eta_L (g_{t, k-1}^t - \nabla F_i(\x_{t, k-1}^i)) \|^2 \\
& \indent + 6K \mathbb{E} \| \eta_L (\nabla F_i(\x_{t, k-1}^i) - \nabla F_i(\x_{t})) \|^2 \\
& \indent + 6K \mathbb{E} \| \eta_L (\nabla F_i(\x_{t}) - \nabla f(\x_t))) \|^2 + 6K \mathbb{E} \| \eta_L \nabla f(\x_t)) \|^2 \\
&\leq (1 + \frac{1}{2K-1}) \mathbb{E} \| \x_{t, k-1}^i - \x_t \|^2 + \eta_L^2 \sigma_L^2 \\
& \indent + 6K \eta_L^2 L^2 \mathbb{E} \| \x_{t, k-1}^i -\x_{t} \|^2 + 6K \eta_L^2 \sigma_G^2 + 6K \mathbb{E} \| \eta_L \nabla f(\x_t)) \|^2 \\
&= (1 + \frac{1}{2K-1} + 6K \eta_L^2 L^2) \mathbb{E} \| \x_{t, k-1}^i - \x_t \|^2 \\
& \indent + \eta_L^2 \sigma_L^2 + 6K \eta_L^2 \sigma_G^2 + 6K \mathbb{E} \| \eta_L \nabla f(\x_t)) \|^2 \\
&\leq (1 + \frac{1}{K-1}) \mathbb{E} \| \x_{t, k-1}^i - \x_t \|^2 + \eta_L^2 \sigma_L^2 \\
& \indent + 6K \eta_L^2 \sigma_G^2 + 6K \mathbb{E} \| \eta_L \nabla f(\x_t)) \|^2
\end{align*}
Unrolling the recursion, we get:
\begin{align*}
& \mathbb{E} \|\x_{t, k}^i - \x_t \|^2 \\
& \leq \sum_{p=0}^{k-1} (1+\frac{1}{K-1} )^p [\eta_L^2 \sigma_L^2+6K \sigma_G^2 + 6K\eta_L^2 \mathbb{E} \| \eta_L \nabla f(\x_t)) \|^2] \\
&\leq \!\! (K \!\!- \!\!1) \big[(1 \!\!+\!\! \frac{1}{K-1})^K \!\!-\!\! 1 \big] \! \big[\eta_L^2 \sigma_L^2 \!\!+\!\! 6K\sigma_G^2 \!\!+\!\! 6K\eta_L^2 \mathbb{E} \| \eta_L \nabla f(\x_t)) \|^2 \big] \\
&\leq  5 K \eta_L^2 (\sigma_L^2 + 6 K \sigma_G^2) + 30 K^2 \eta_L^2 \| \nabla f(\x_t) \|^2
\end{align*}







\section{Appendix II: Experiments} \label{exp}

\begin{figure}[t!]
	\begin{minipage}[t]{0.49\linewidth}
	\centering
	{\includegraphics[width=0.9\columnwidth]{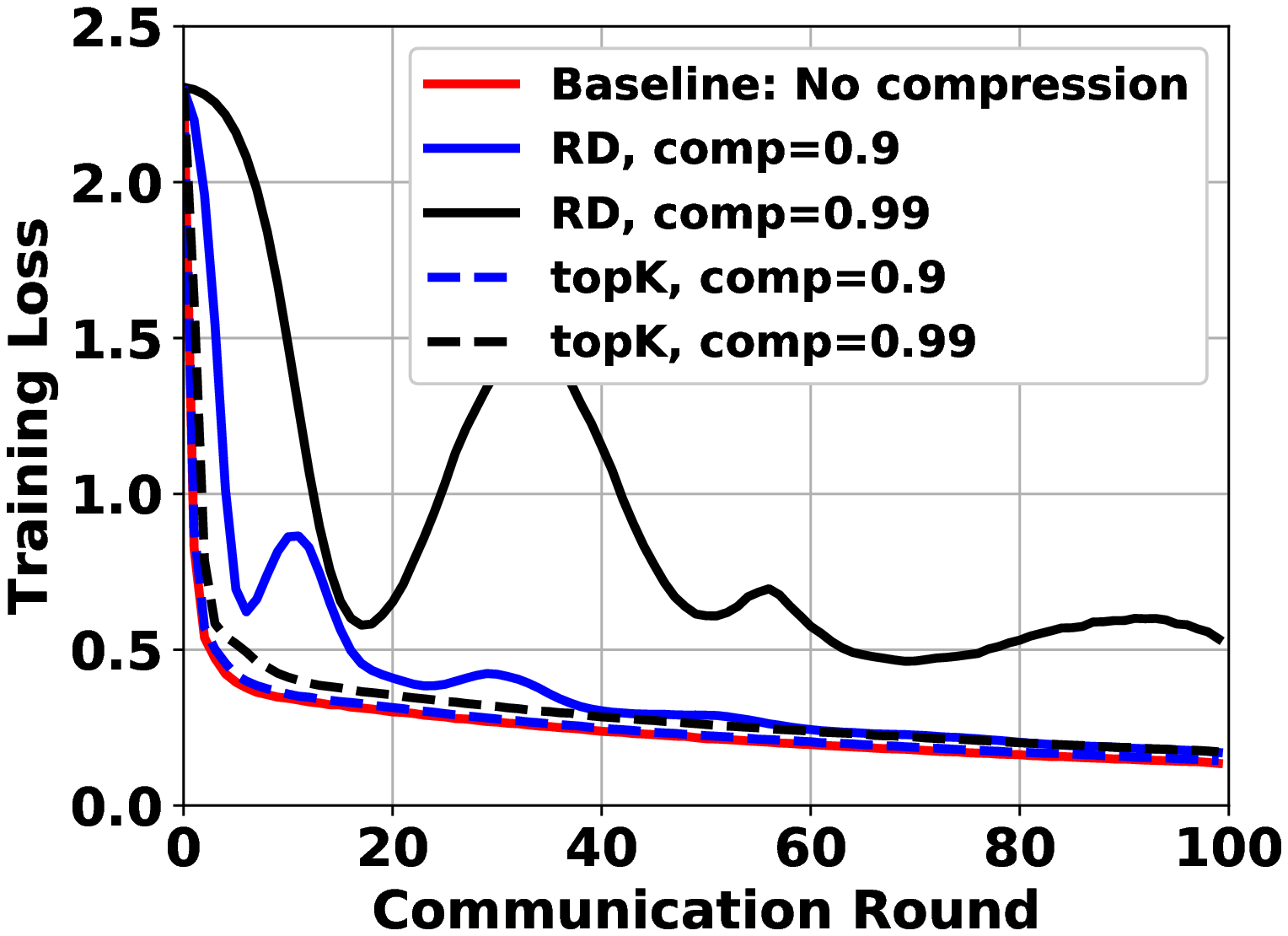}}
	\vspace{-.2in}
	\end{minipage}
	\hfill
	\begin{minipage}[t]{0.49\linewidth}
	\centering
	{\includegraphics[width=0.9\columnwidth]{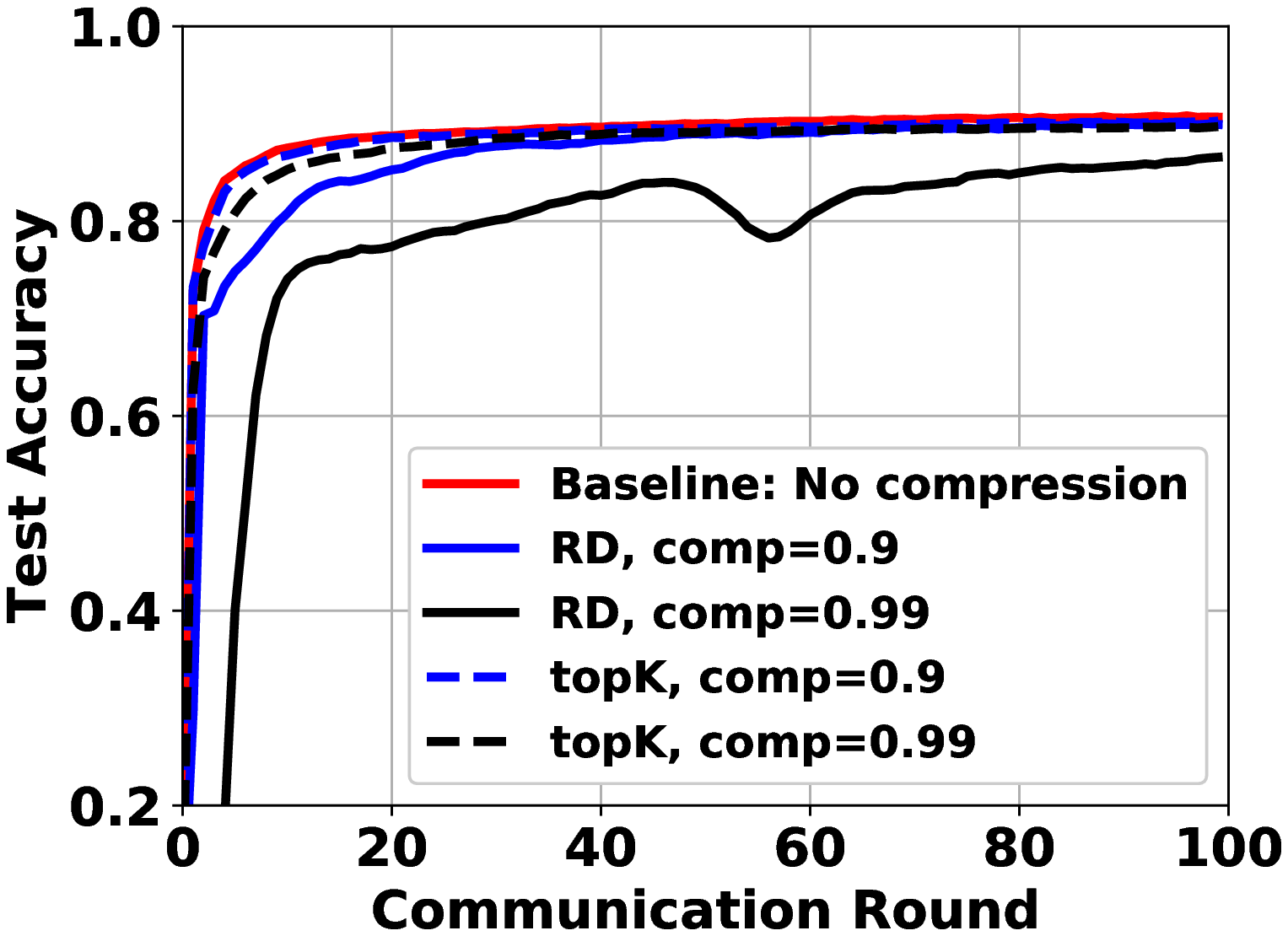}} 
	\vspace{-.2in}
	\end{minipage}

	\begin{minipage}[t]{0.49\linewidth}
	\centering
	{\includegraphics[width=0.9\columnwidth]{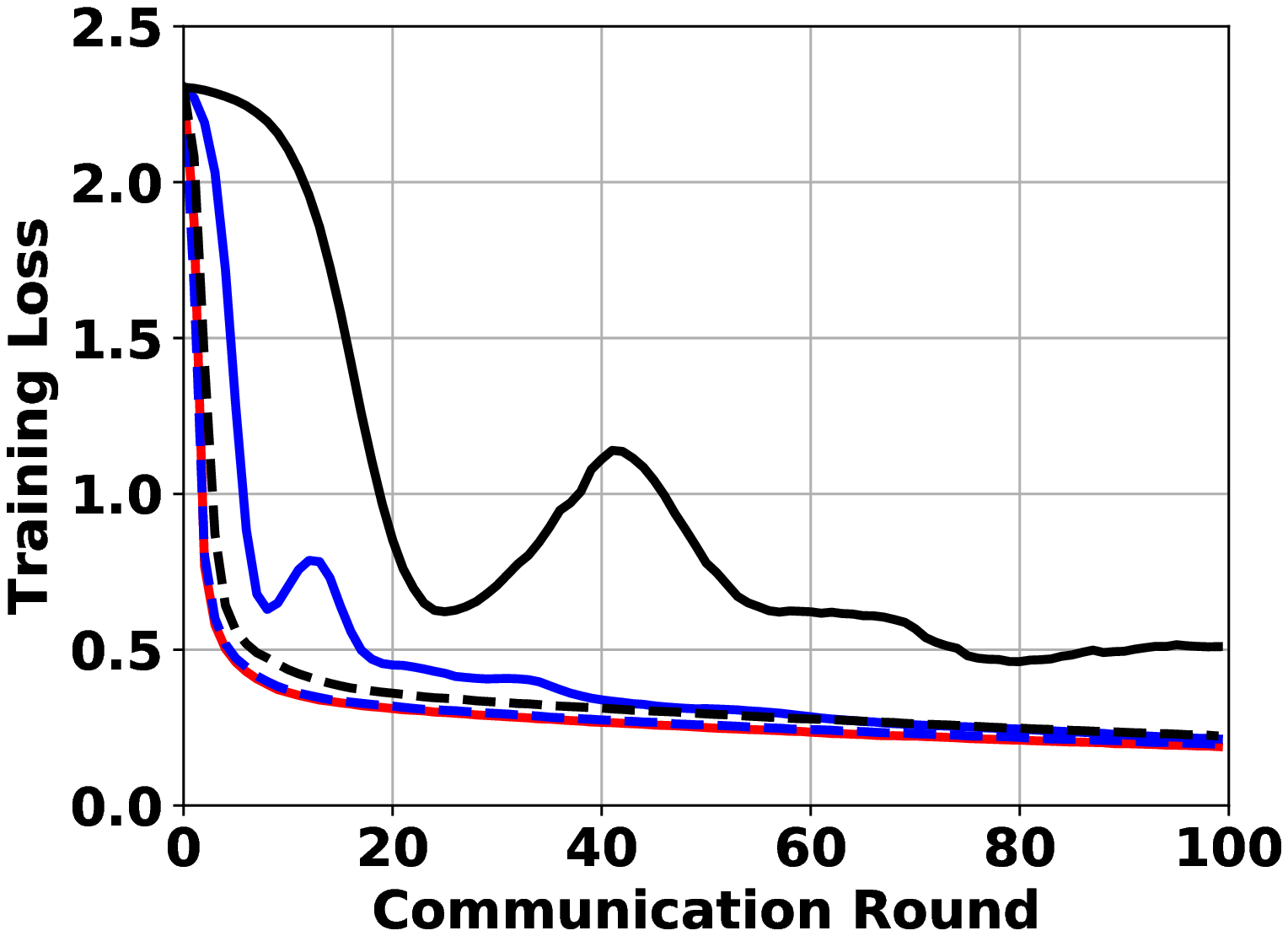}}
	\vspace{-.2in}
	\end{minipage}
	\hfill
	\begin{minipage}[t]{0.49\linewidth}
	\centering
	{\includegraphics[width=0.9\columnwidth]{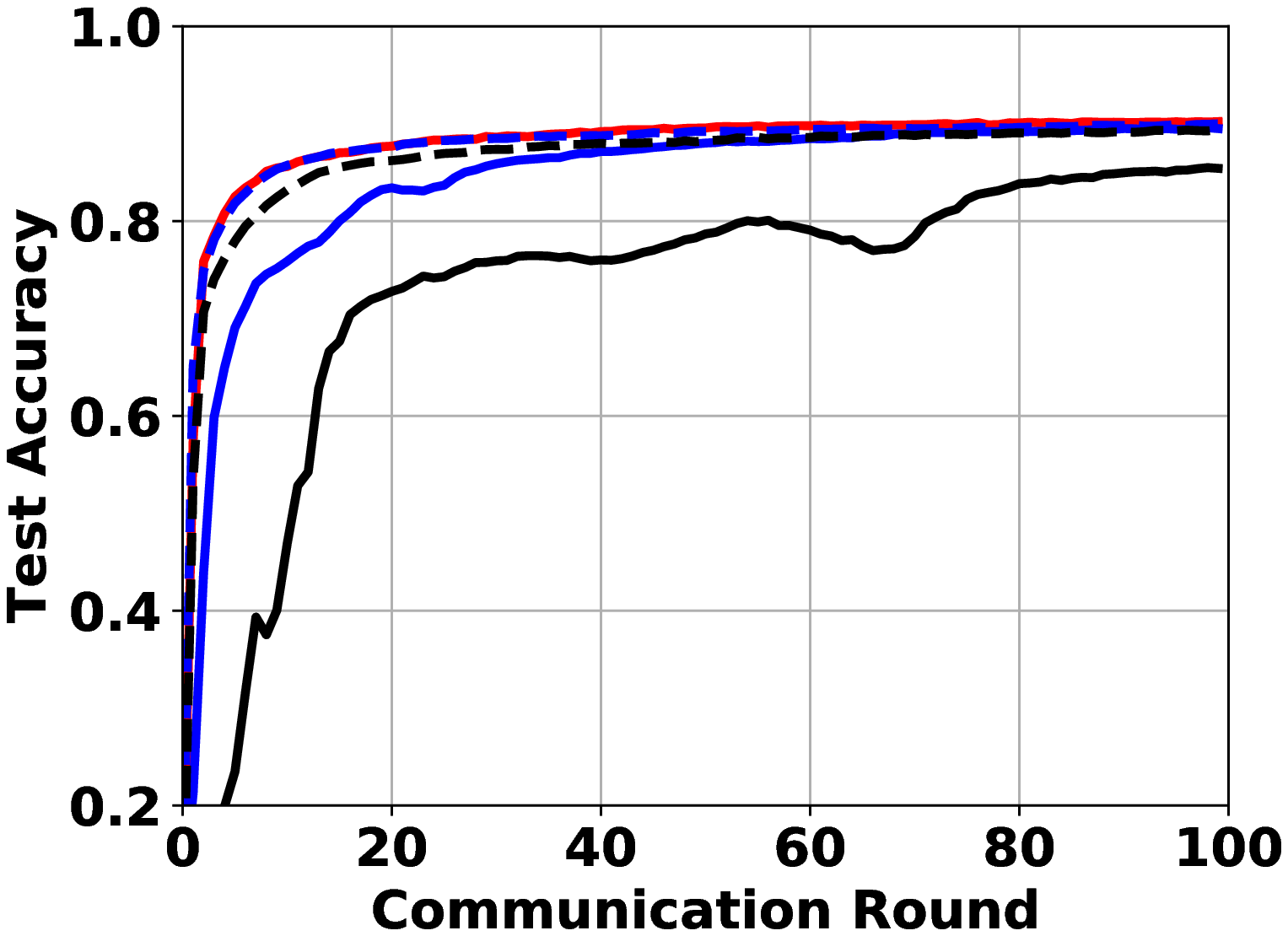}}
	\vspace{-.2in}
	\end{minipage}

	\begin{minipage}[t]{0.49\linewidth}
	\centering
	{\includegraphics[width=0.9\columnwidth]{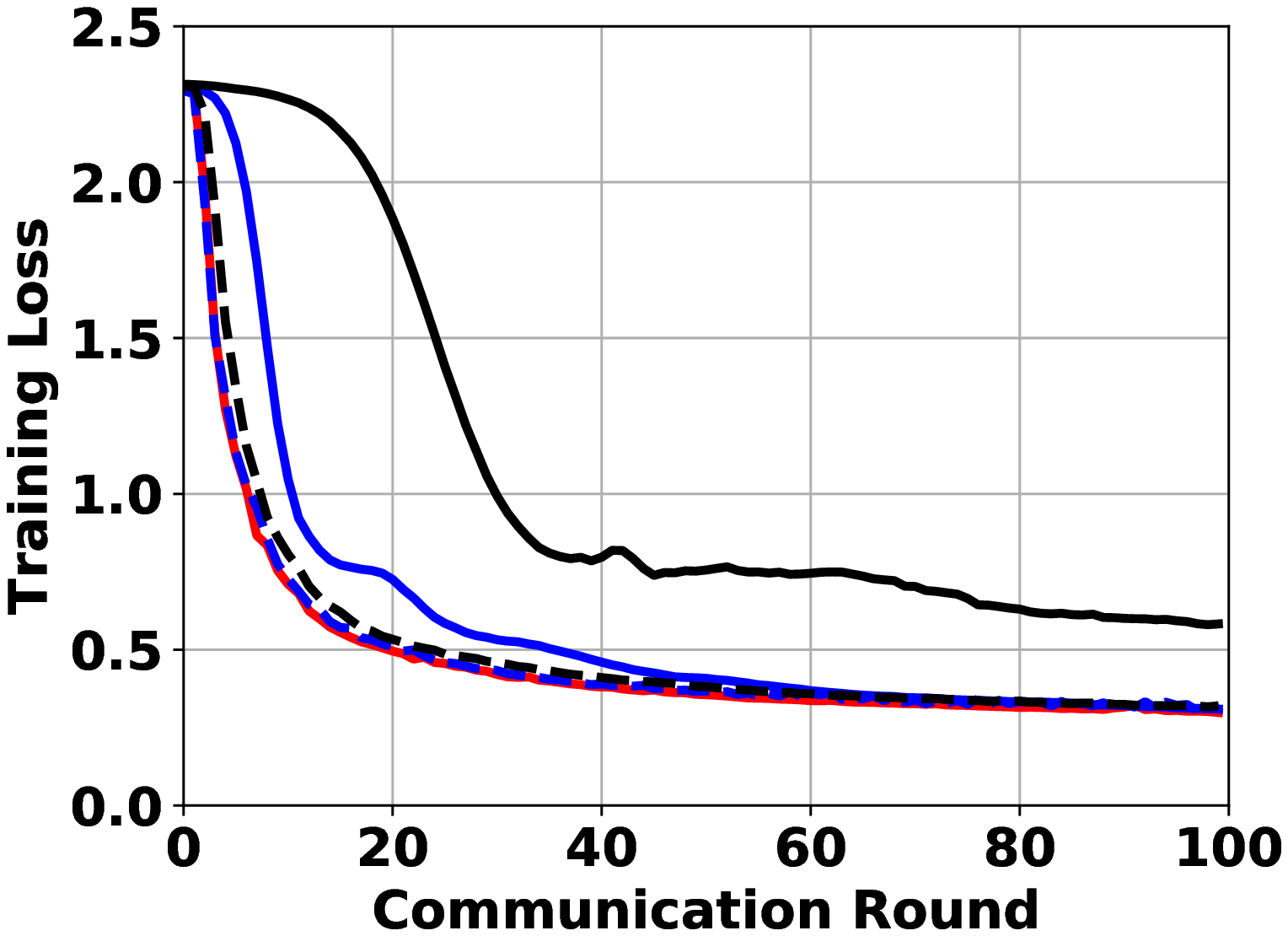}}
	\vspace{-.2in}
	\end{minipage}
	\hfill
	\begin{minipage}[t]{0.49\linewidth}
	\centering
	{\includegraphics[width=0.9\columnwidth]{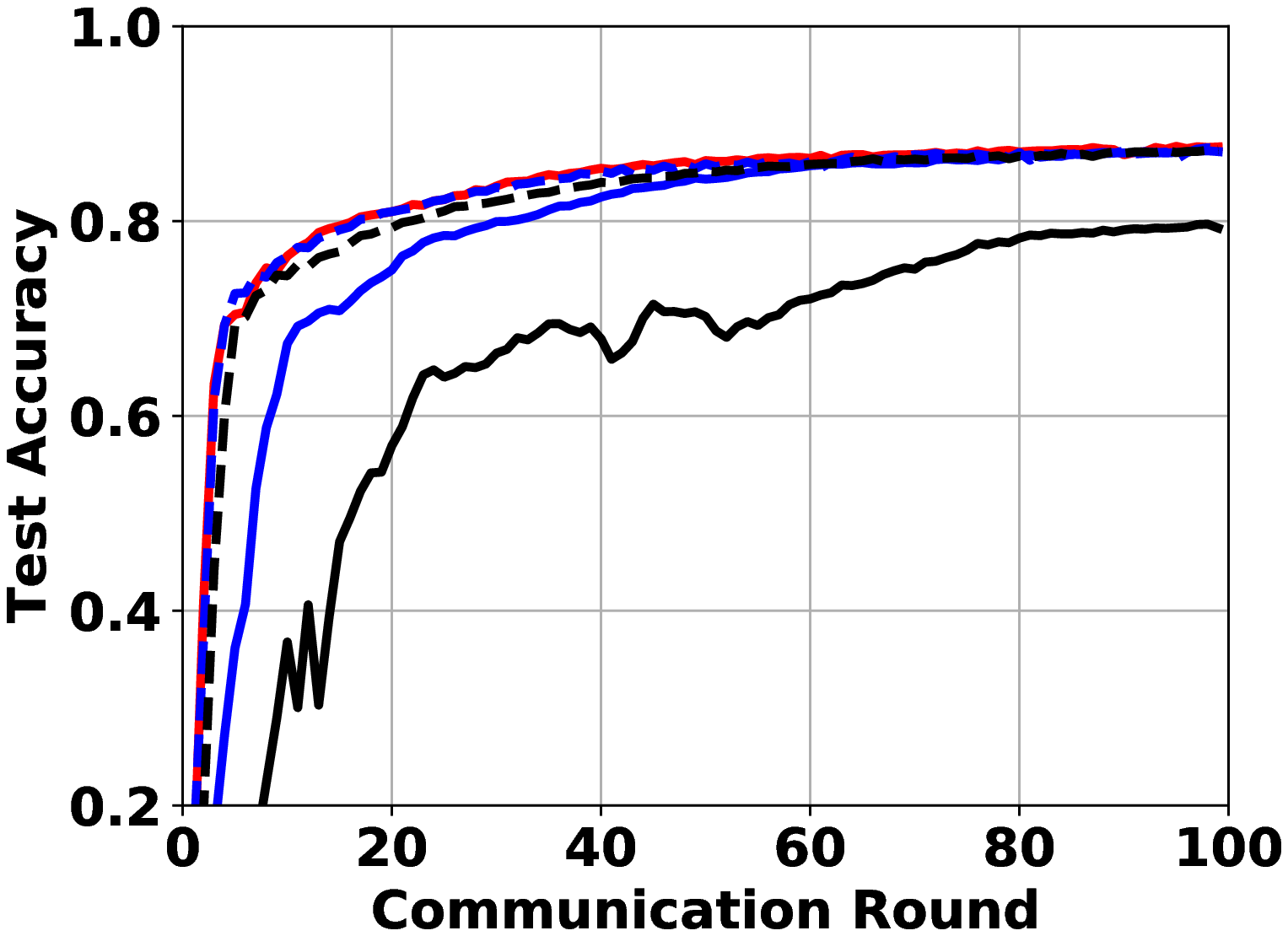}}
	\vspace{-.2in}
	\end{minipage}

	\begin{minipage}[t]{0.49\linewidth}
	\centering
	{\includegraphics[width=0.9\columnwidth]{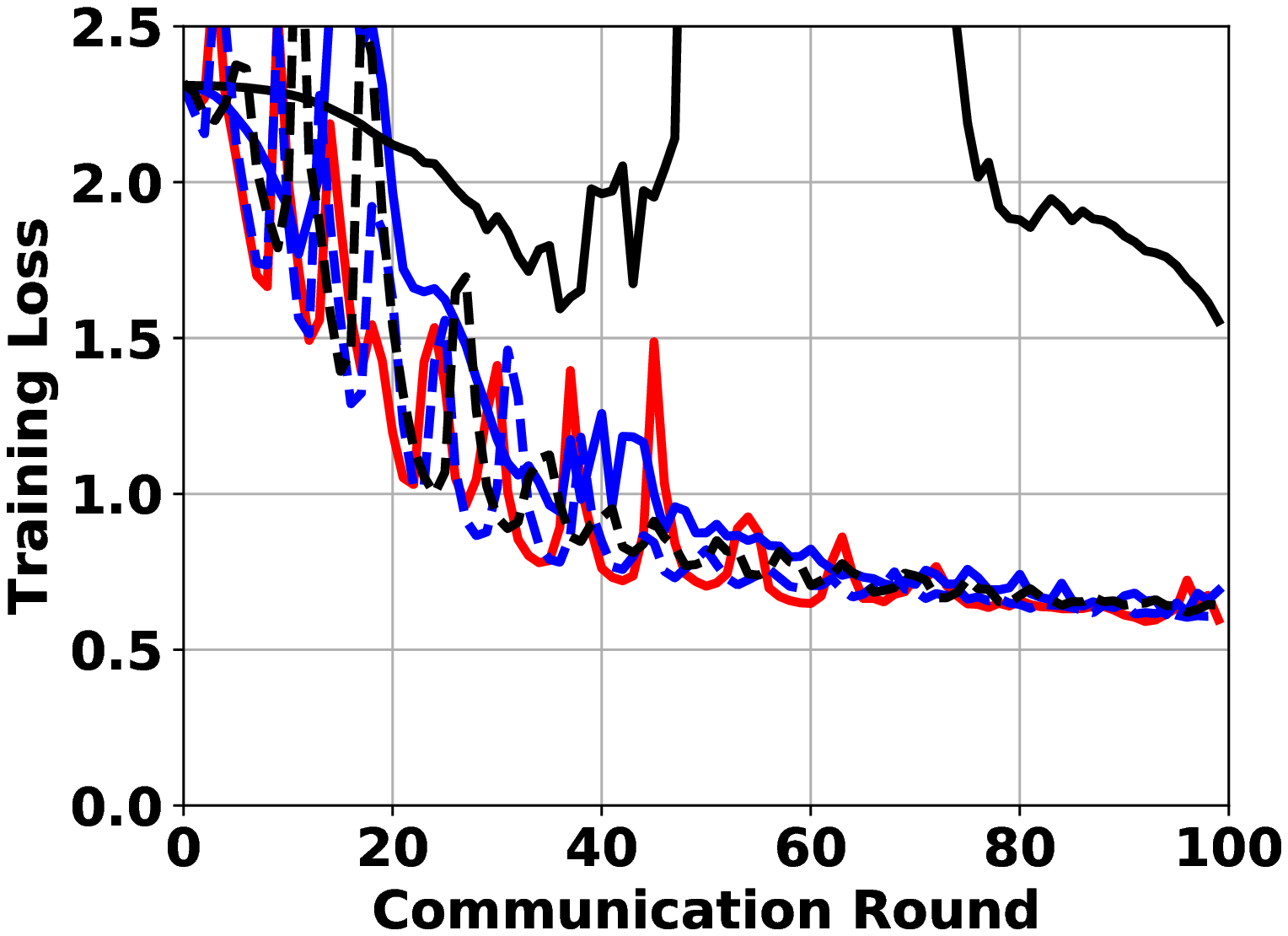}}
	\vspace{-.2in}
	\end{minipage}
	\begin{minipage}[t]{0.49\linewidth}
	\centering
	{\includegraphics[width=0.9\columnwidth]{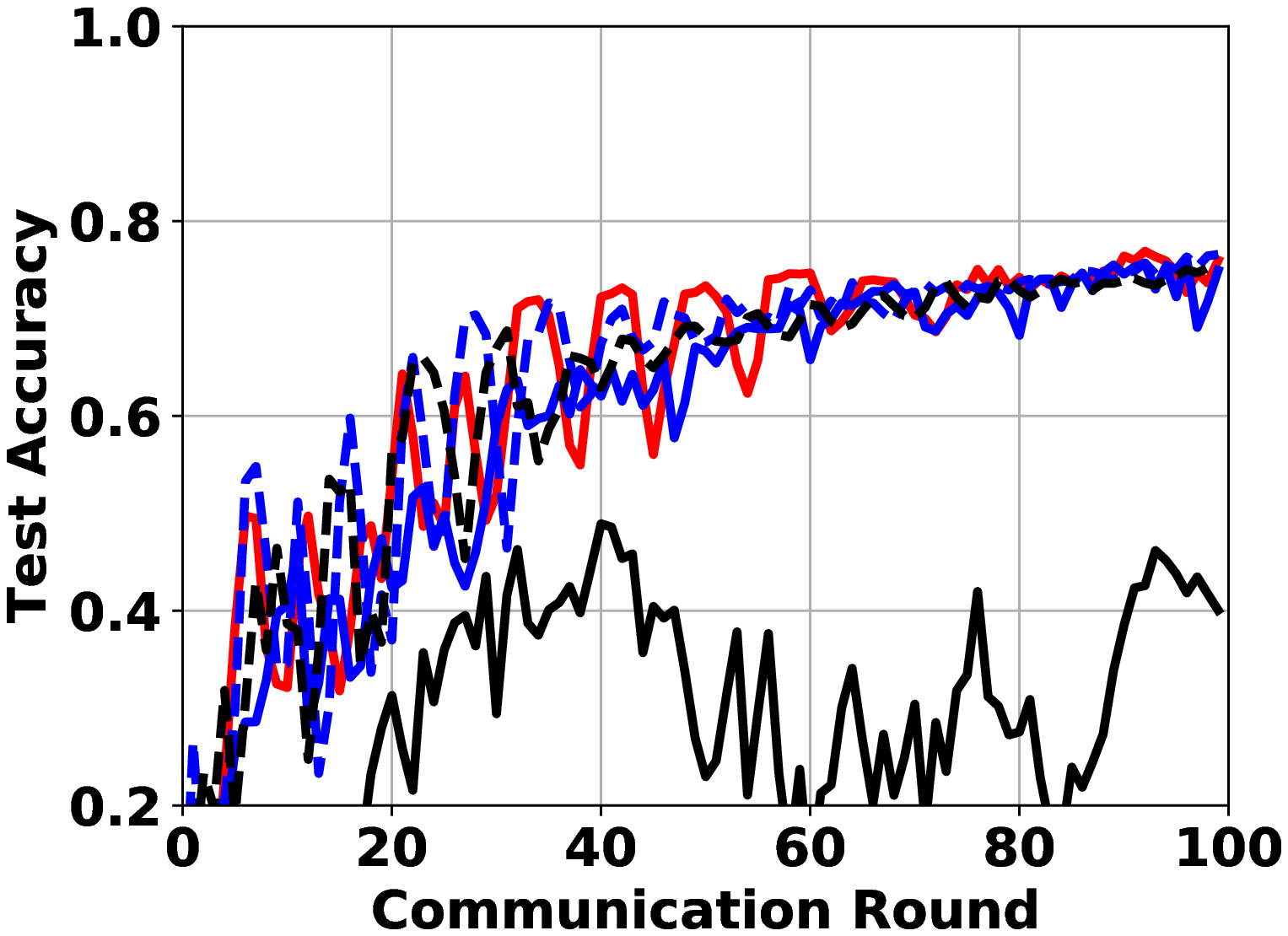}}
	\vspace{-.2in}
	\end{minipage}
\caption{Training loss (left) and test accuracy (right) for the CNN model for Fashion-MNIST. The non-i.i.d. levels are $p = 10, 5, 2, 1$ from top to bottom.}
\label{appdx_fig1}
\end{figure}%

\begin{figure}[t!]
	\begin{minipage}[t]{0.49\linewidth}
	\centering
	{\includegraphics[width=0.9\columnwidth]{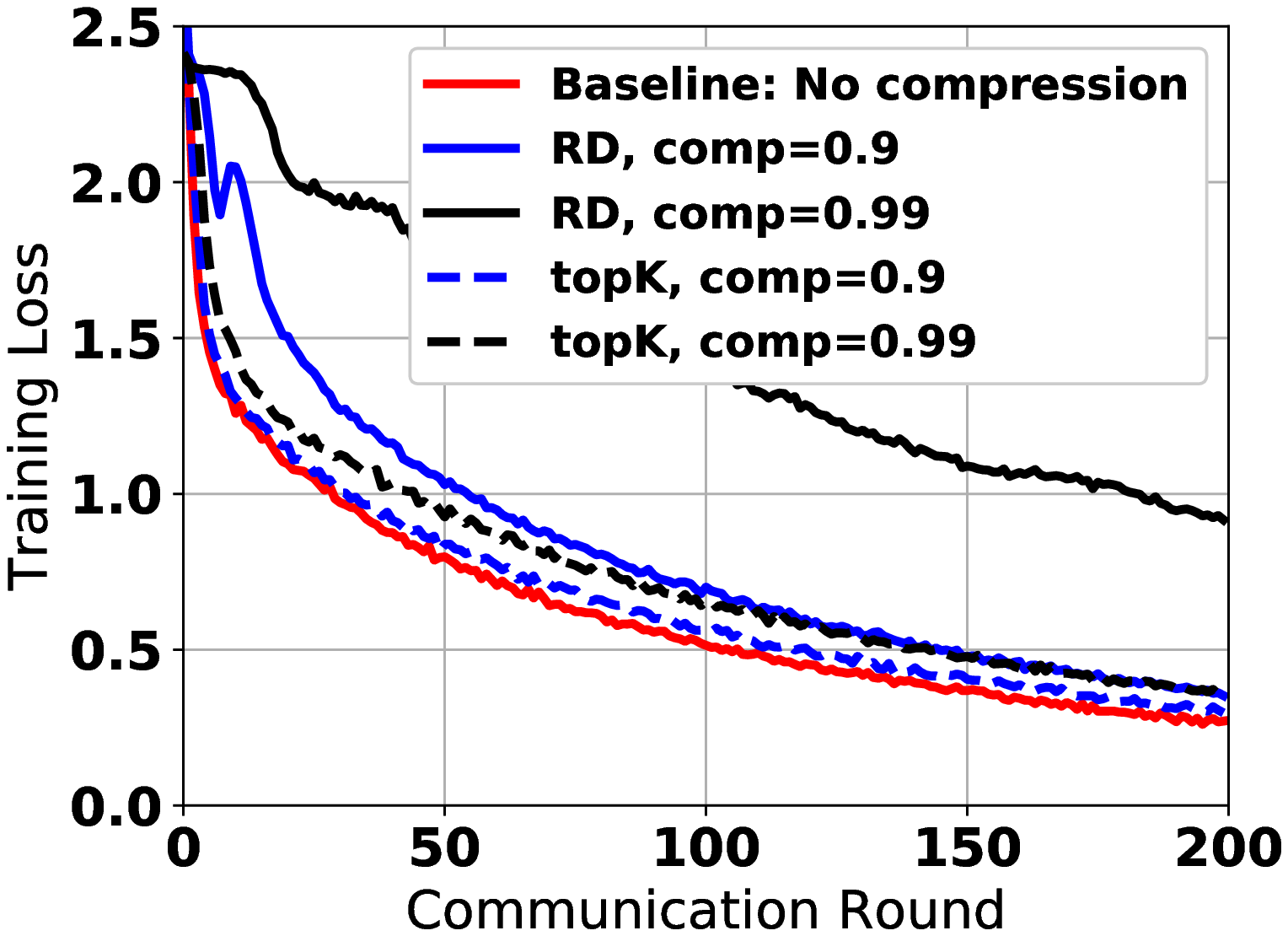}}
	\vspace{-.2in}
	\end{minipage}
	\hfill
	\begin{minipage}[t]{0.49\linewidth}
	\centering
	{\includegraphics[width=0.9\columnwidth]{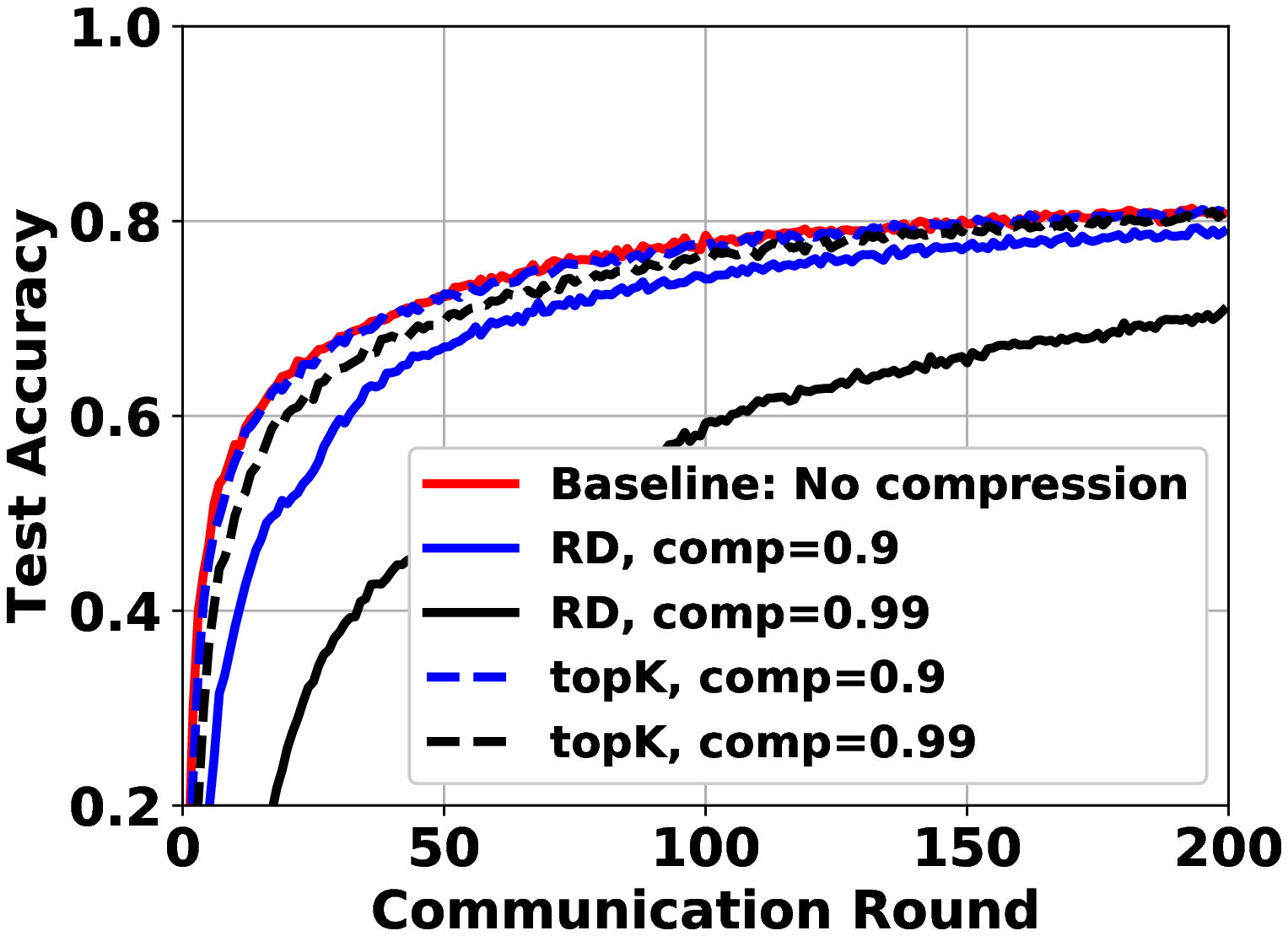}} 
	\vspace{-.2in}
	\end{minipage}

	\begin{minipage}[t]{0.49\linewidth}
	\centering
	{\includegraphics[width=0.9\columnwidth]{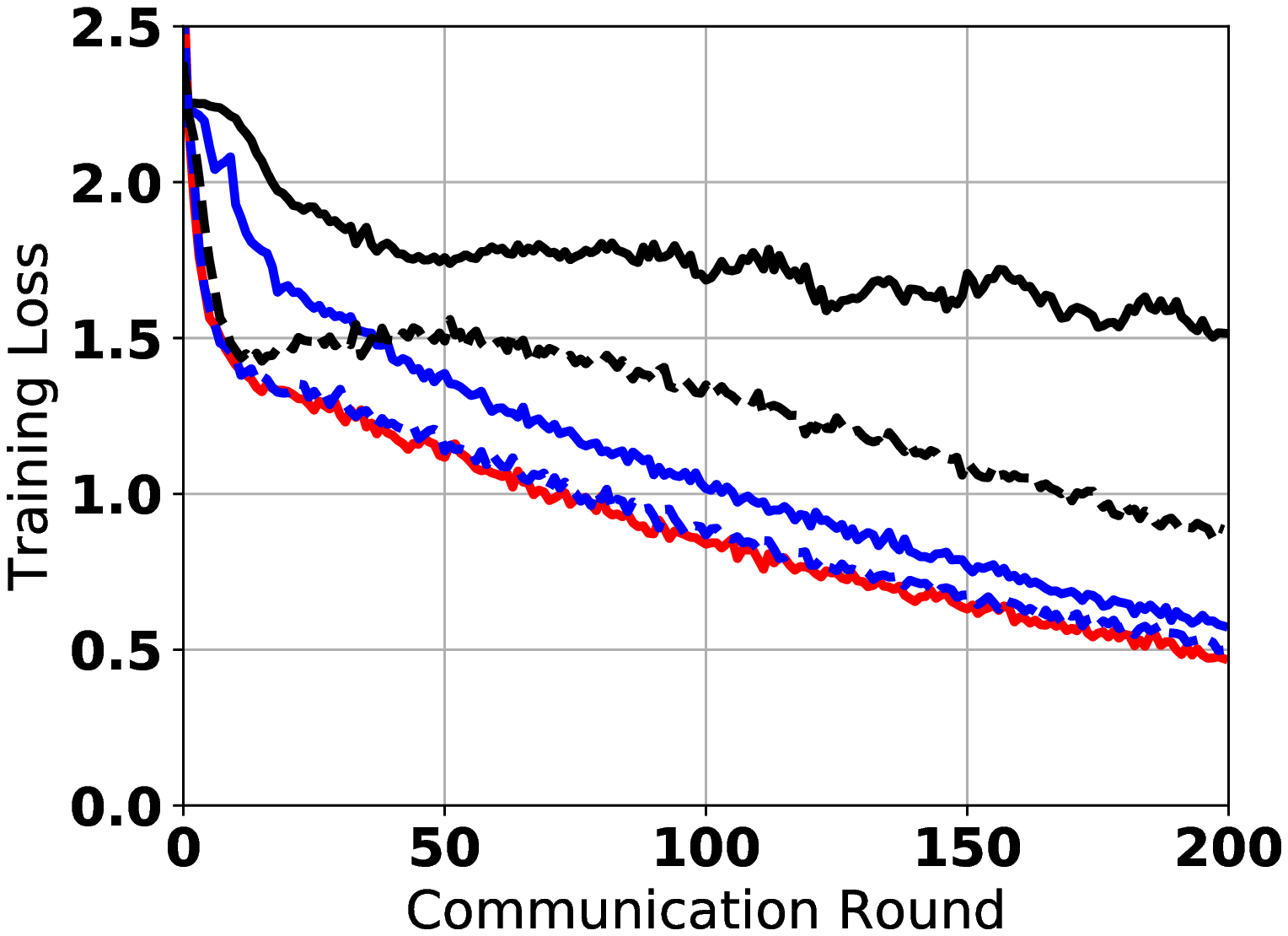}}
	\vspace{-.2in}
	\end{minipage}
	\hfill
	\begin{minipage}[t]{0.49\linewidth}
	\centering
	{\includegraphics[width=0.9\columnwidth]{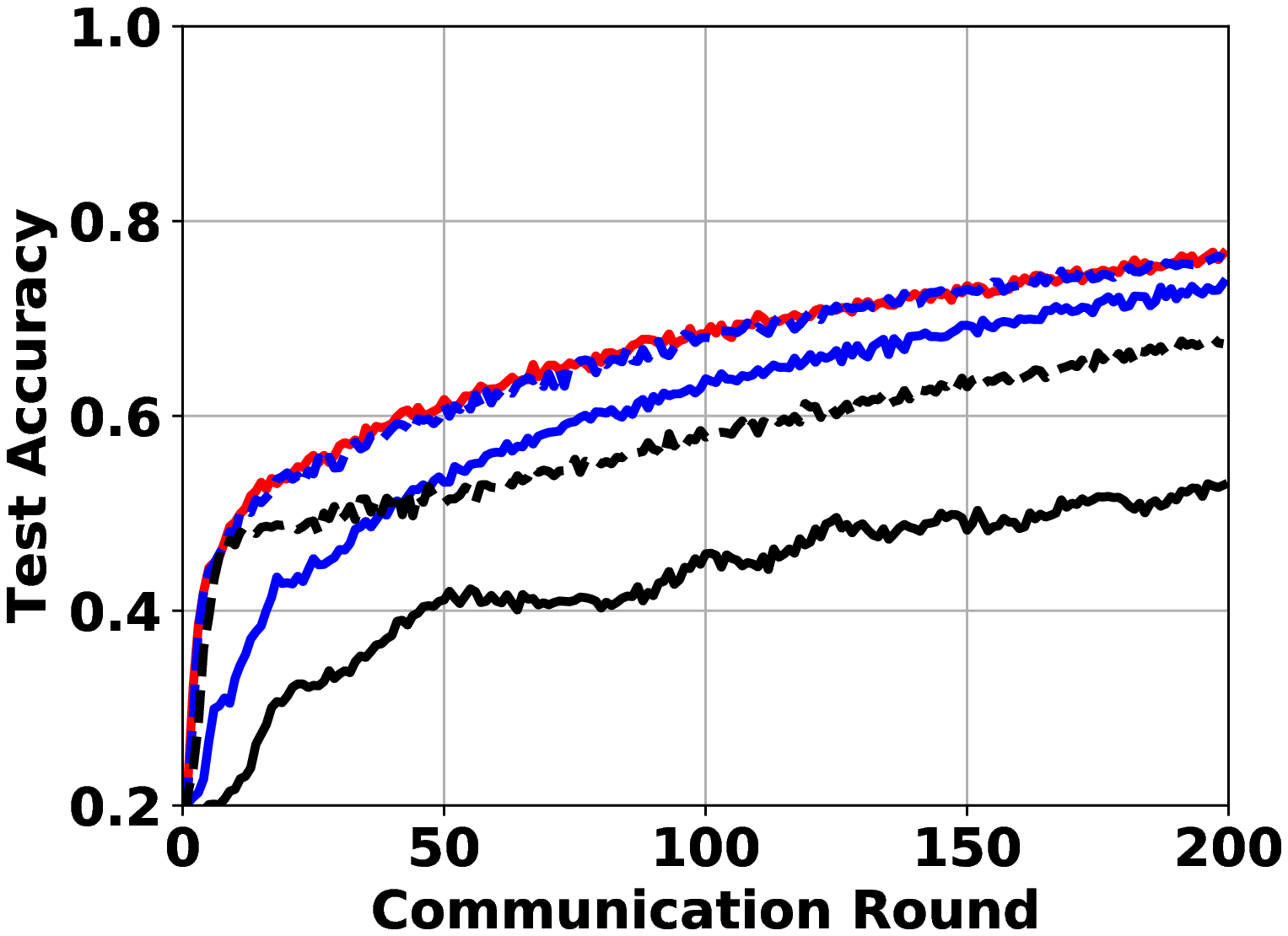}}
	\vspace{-.2in}
	\end{minipage}

	\begin{minipage}[t]{0.49\linewidth}
	\centering
	{\includegraphics[width=0.9\columnwidth]{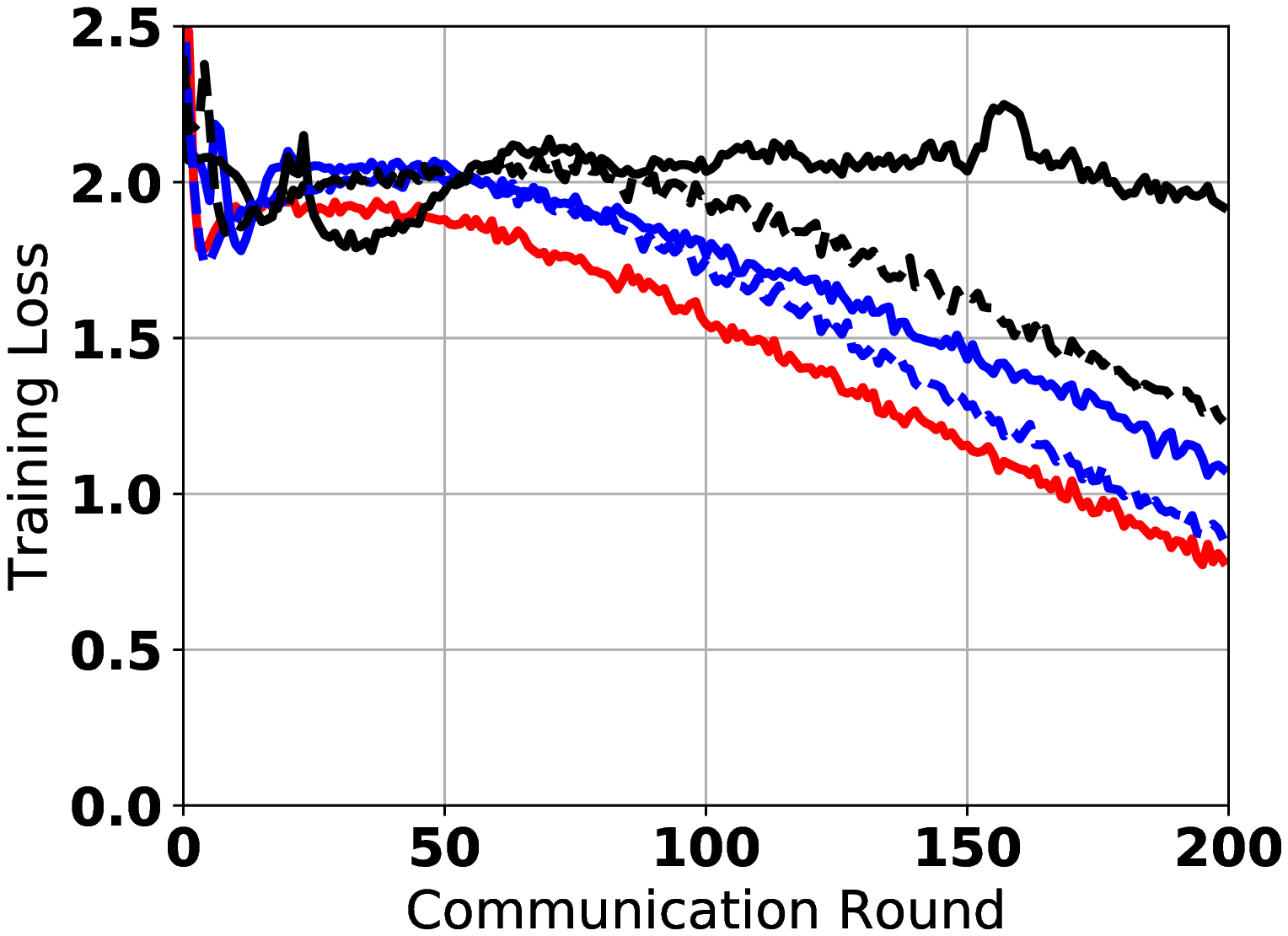}}
	\vspace{-.2in}
	\end{minipage}
	\hfill
	\begin{minipage}[t]{0.49\linewidth}
	\centering
	{\includegraphics[width=0.9\columnwidth]{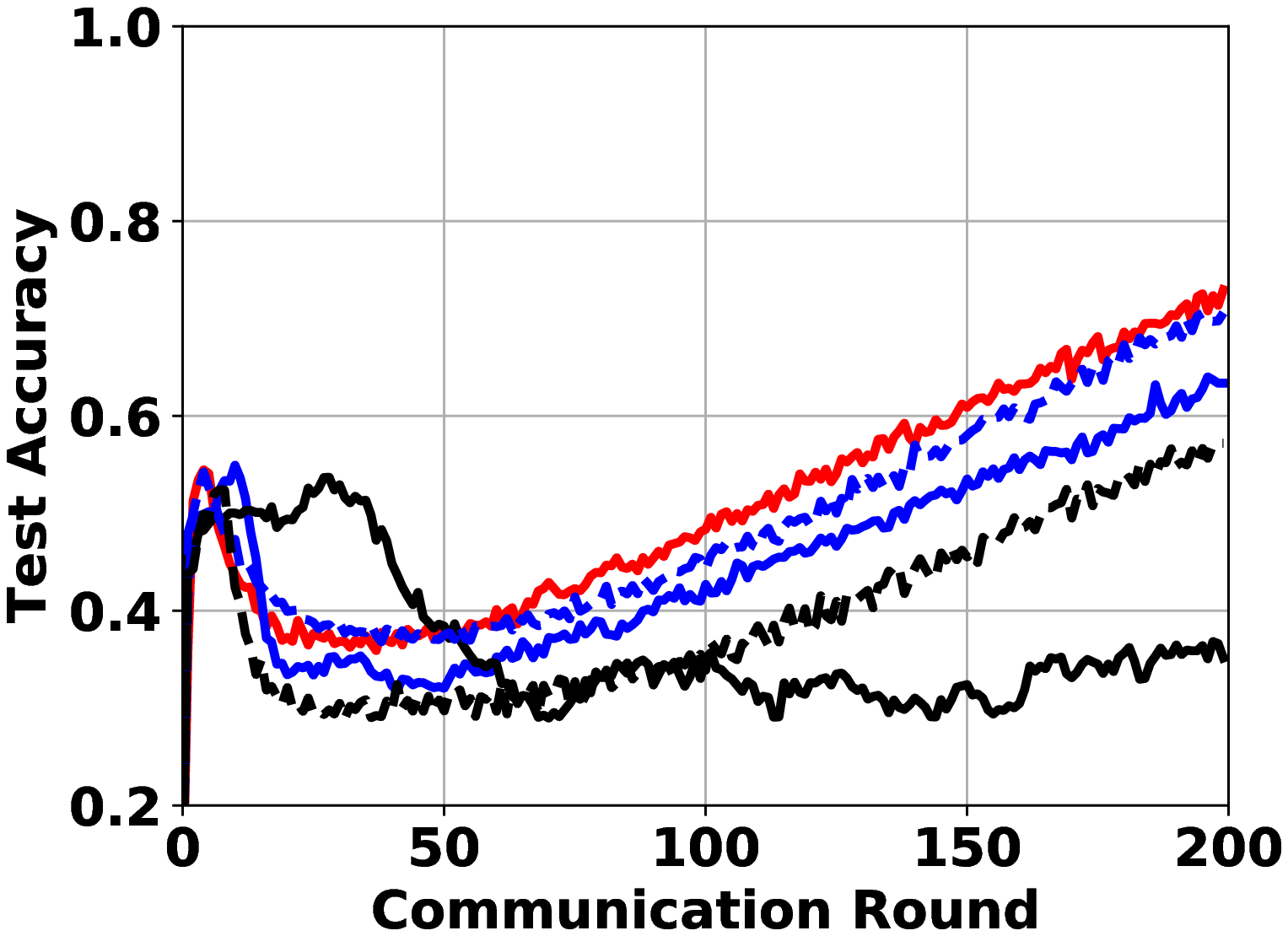}}
	\vspace{-.2in}
	\end{minipage}

	\begin{minipage}[t]{0.49\linewidth}
	\centering
	{\includegraphics[width=0.9\columnwidth]{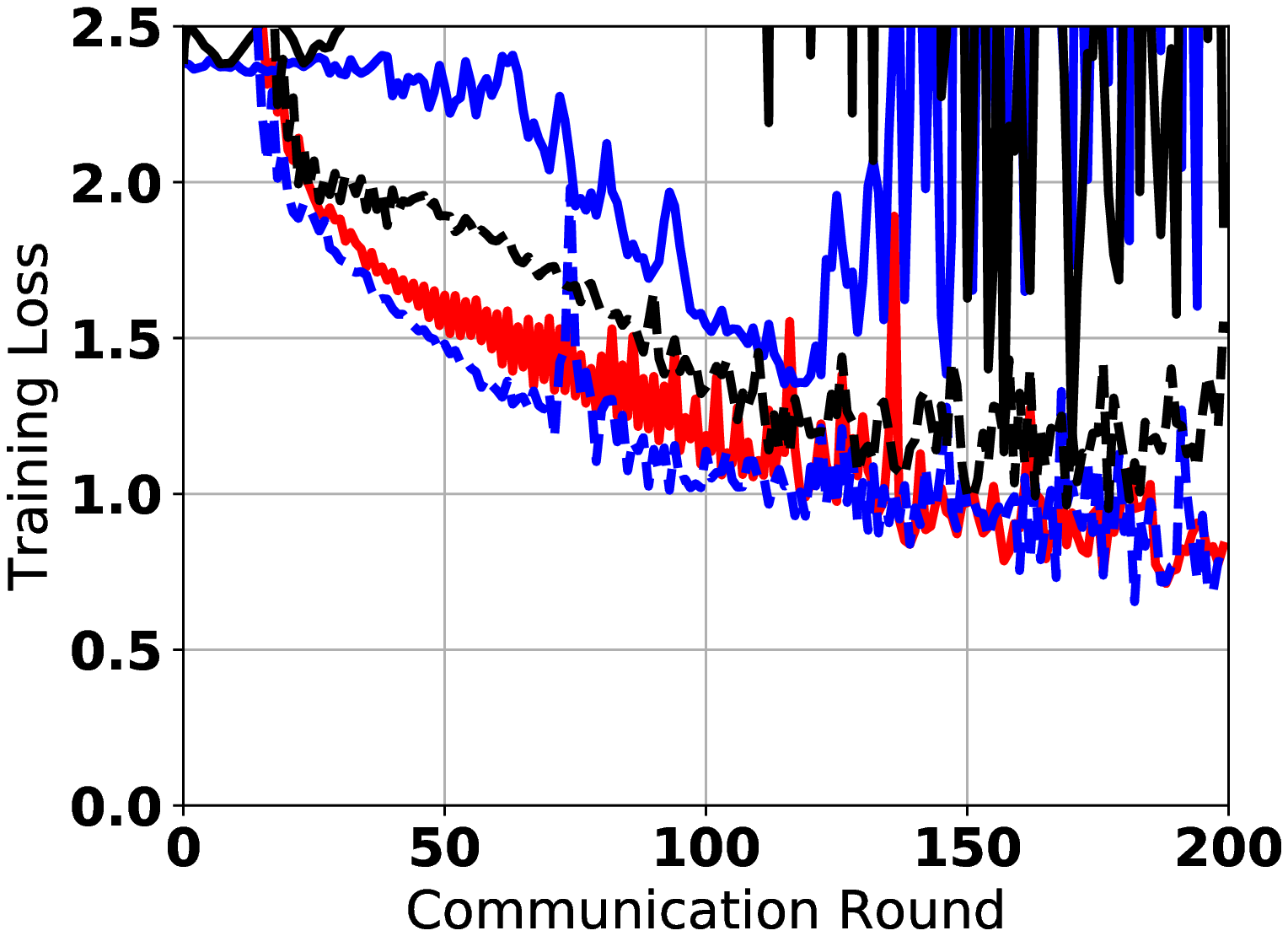}}
	\vspace{-.2in}
	\end{minipage}
	\begin{minipage}[t]{0.49\linewidth}
	\centering
	{\includegraphics[width=0.9\columnwidth]{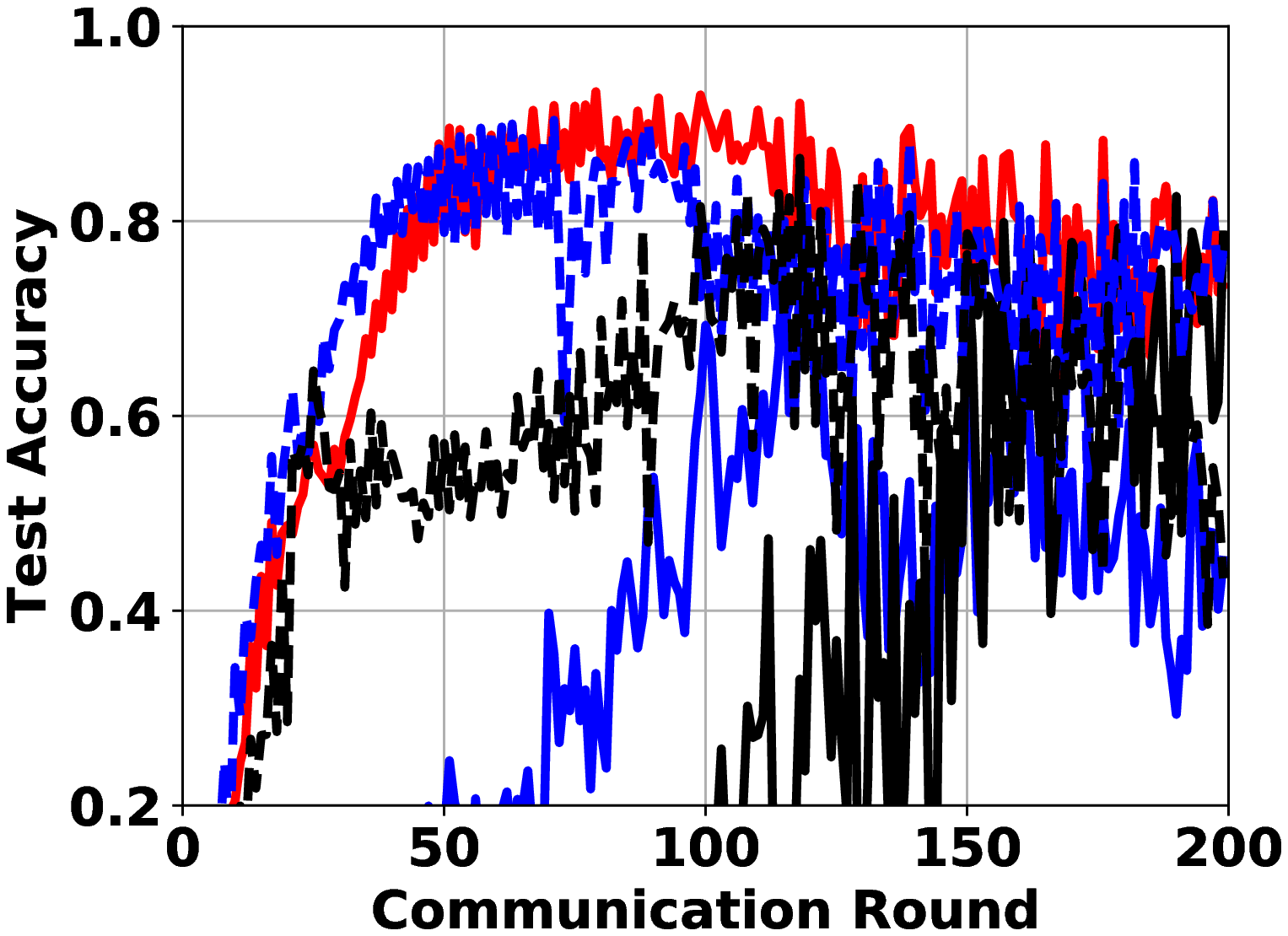}}
	\vspace{-.2in}
	\end{minipage}

\caption{Training loss (top) and test accuracy (bottom) for the Resnet-18 model for CIFAR-10. The non-i.i.d. levels are $p = 10, 5, 2, 1$ from top to bottom.}
\label{appdx_cifar10_fig1}
\end{figure}%

We provide the full detail of the experiments.
We use three datasets in FL (non-i.i.d. version) settings, including MNIST, Fashion-MNIST and CIFAR-10.
In the appendix, we show the results of Fashion-MNIST and CIFAR-10 while these of MNIST are shown in the paper.

We split the data based on the classes of items ($p$) they contain in their datasets.
The system is containing $1$ central server and $m=100$ workers, whose local dataset is distributed randomly and evenly in a class-based manner. The local dataset for every worker contains only certain classes of items with the same number of training/test samples.
Each of these three datasets contains $10$ different classes of items.
So parameter $p$ can be used to represent the non-i.i.d. degree of the datasets qualitatively.
For example, for $p = 1$, each worker only has training/testing samples labeled with one class, which causes heterogeneity among different workers.
For $p = 10$, each worker has samples with 10 classes, which is essentially i.i.d. case since the total classes for these three datasets are $10$.
We set four levels of non-i.i.d. version for comparison, i.e., $p = 1, 2, 5, 10$.

We run two models: convolutional neural network (CNN) on MNIST and Fashion-MNIST and Resnet-18 on CIFAR-10.

\begin{table} [!htb]
\begin{center}
\caption{CNN Architecture for MNIST and Fashion-MNIST.}
\begin{tabular}{c c} 
	\hline
	Layer Type & Size \\
	\hline
	Convolution + ReLu & $5 \times 5 \times 32$ \\
	Max Pooling & $2 \times 2$\\
	Convolution + ReLu & $5 \times 5 \times 64$ \\
	Max Pooling & $2 \times 2$\\
	Fully Connected + ReLU & $1024 \times 512$\\
	Fully Connected & $512 \times 10$ \\
	\hline
\end{tabular}
\label{tab: 1}
\end{center}
\end{table}

\begin{figure}[t!]
	\begin{minipage}[t]{0.49\linewidth}
	\centering
	{\includegraphics[width=0.9\columnwidth]{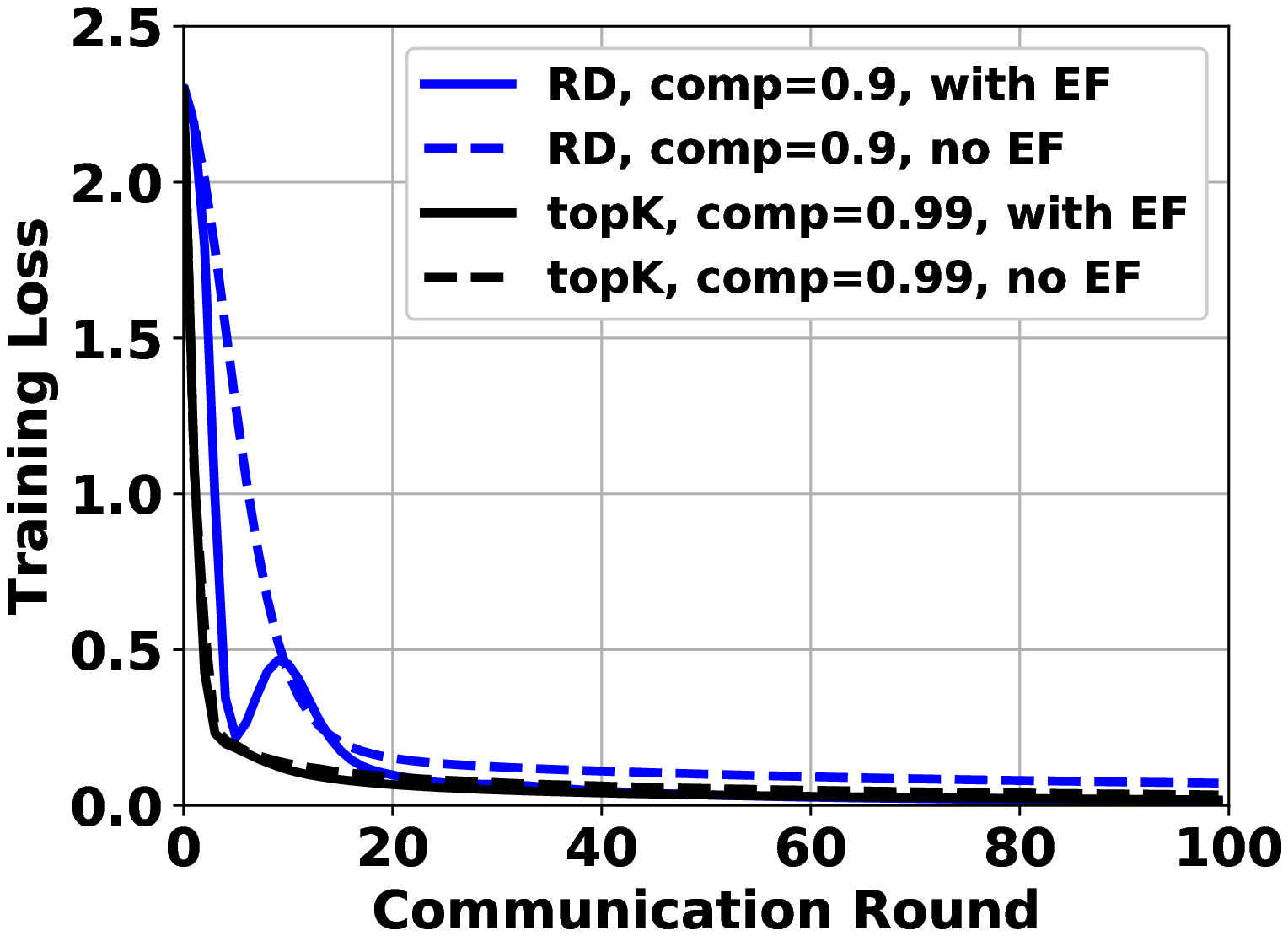}}
	\vspace{-.2in}
	\end{minipage}
	\hfill
	\begin{minipage}[t]{0.49\linewidth}
	\centering
	{\includegraphics[width=0.9\columnwidth]{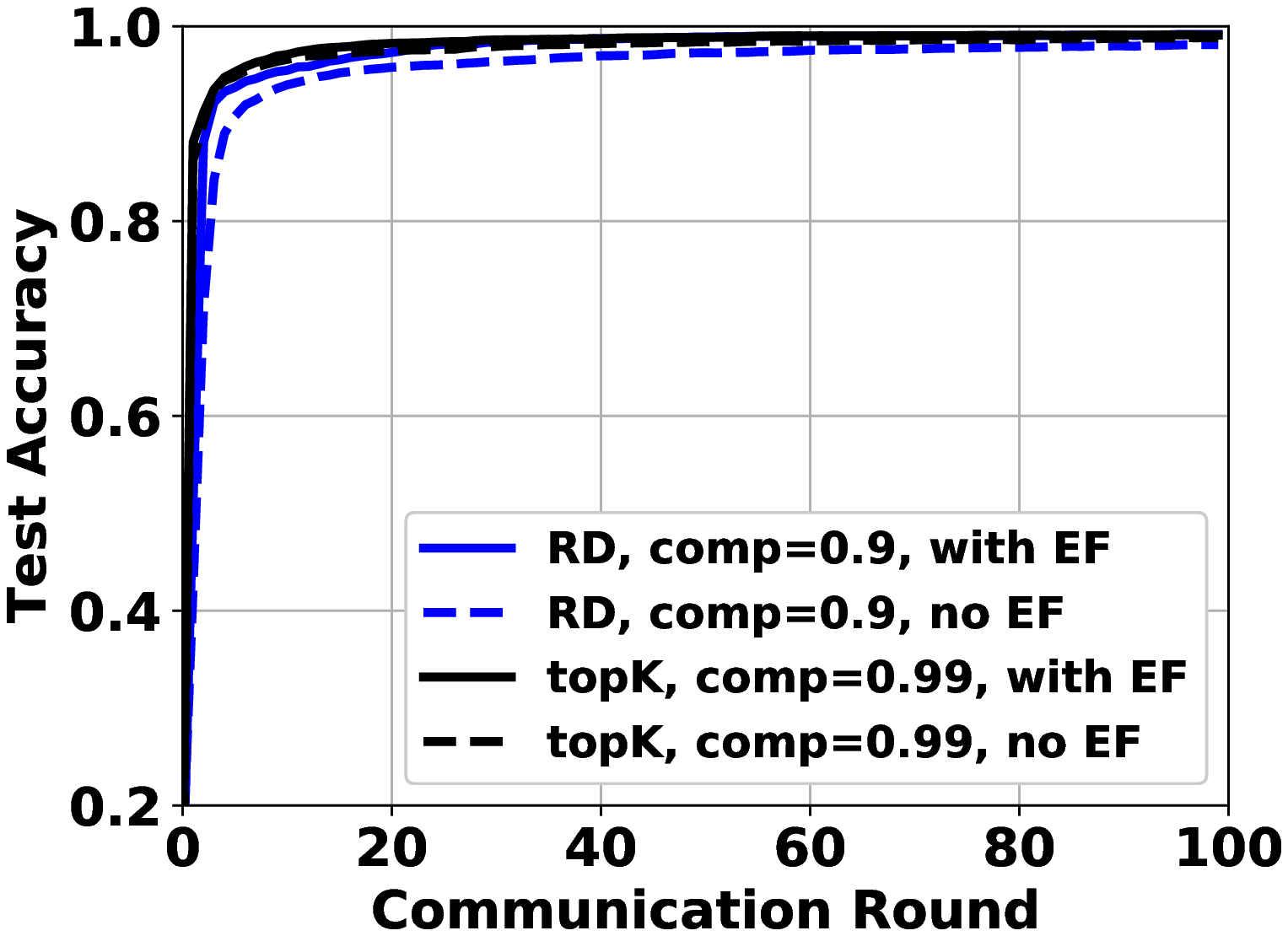}} 
	\vspace{-.2in}
	\end{minipage}

	\begin{minipage}[t]{0.49\linewidth}
	\centering
	{\includegraphics[width=0.9\columnwidth]{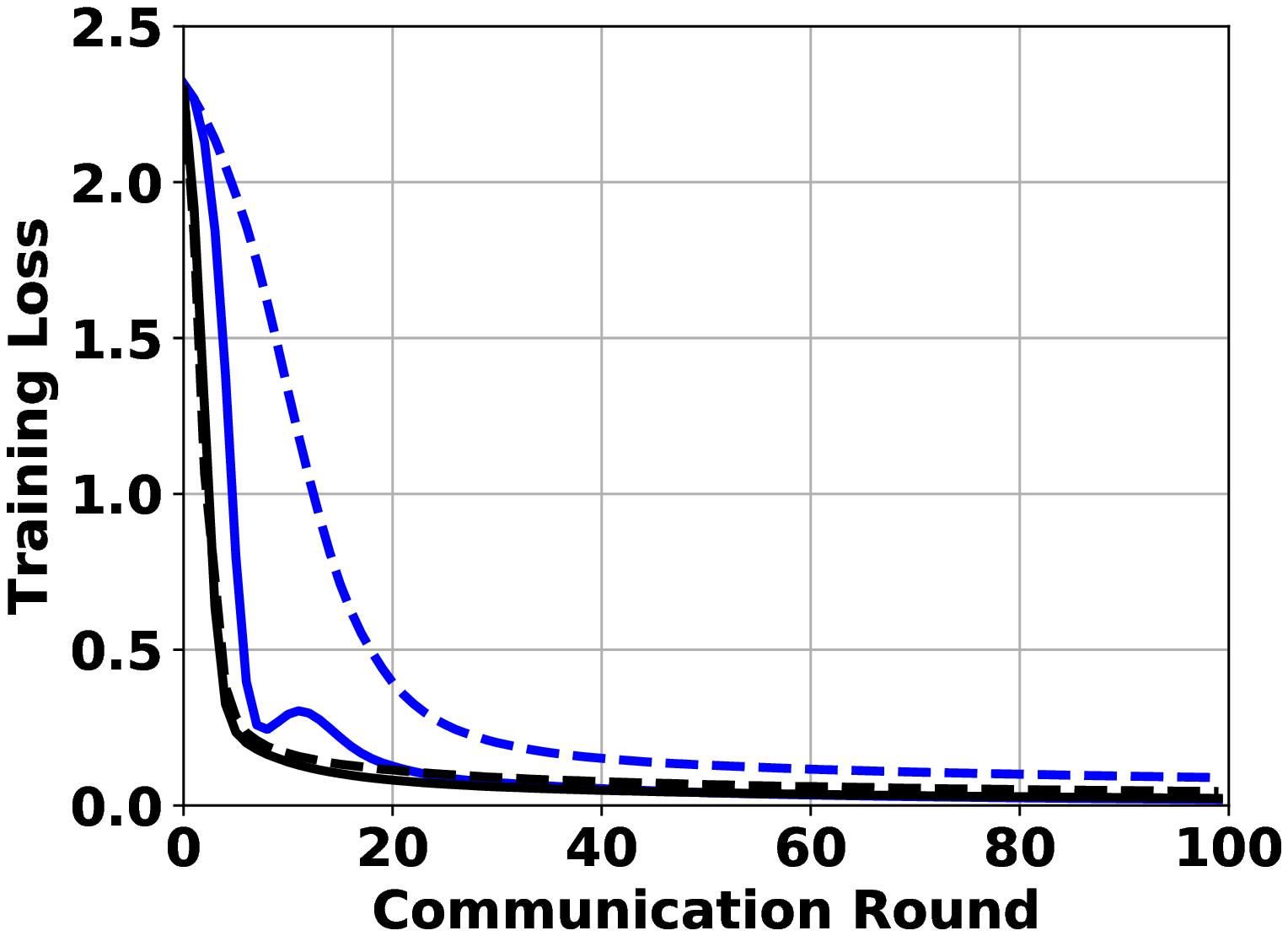}}
	\vspace{-.2in}
	\end{minipage}
	\hfill
	\begin{minipage}[t]{0.49\linewidth}
	\centering
	{\includegraphics[width=0.9\columnwidth]{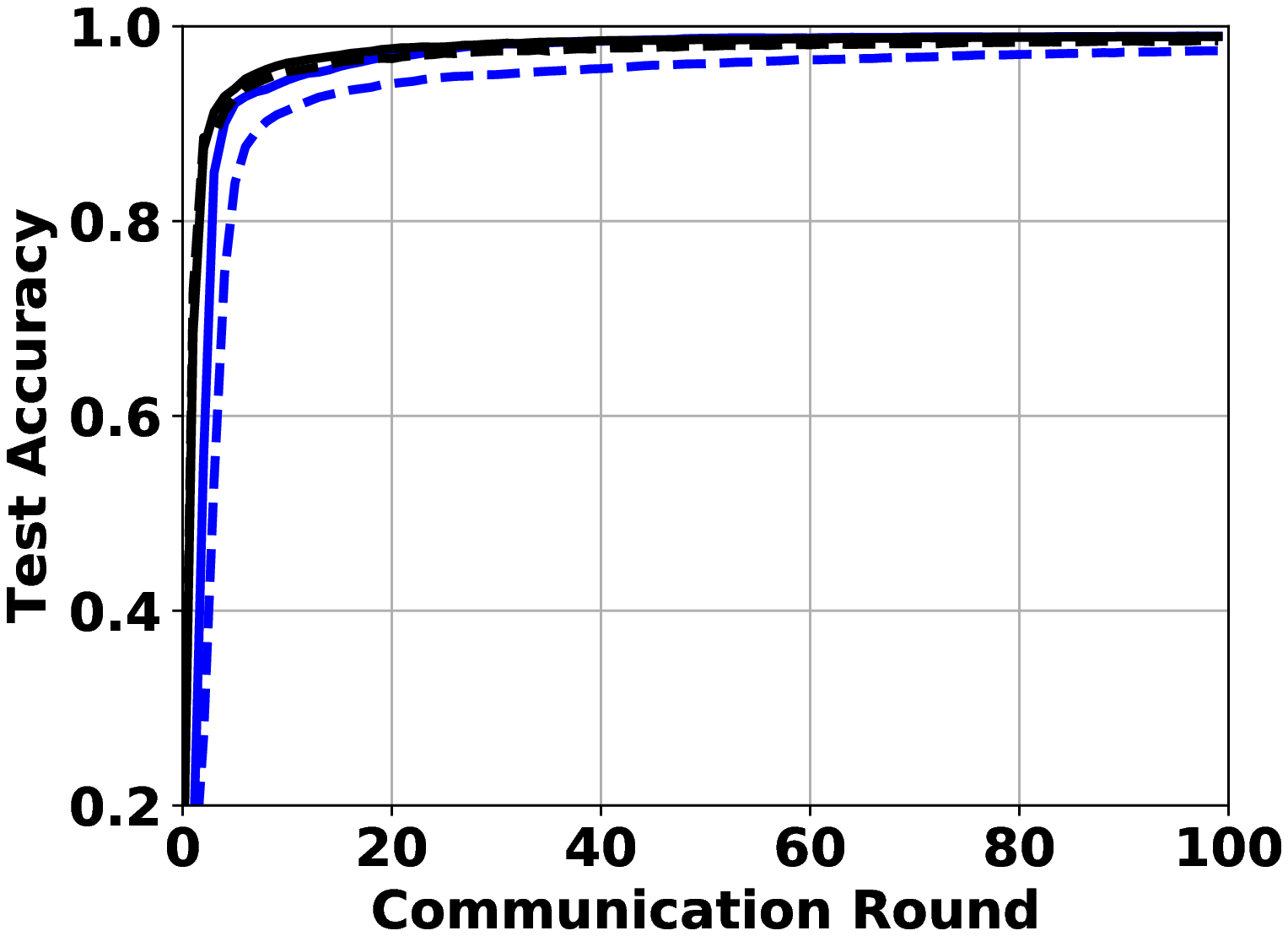}}
	\vspace{-.2in}
	\end{minipage}

	\begin{minipage}[t]{0.49\linewidth}
	\centering
	{\includegraphics[width=0.9\columnwidth]{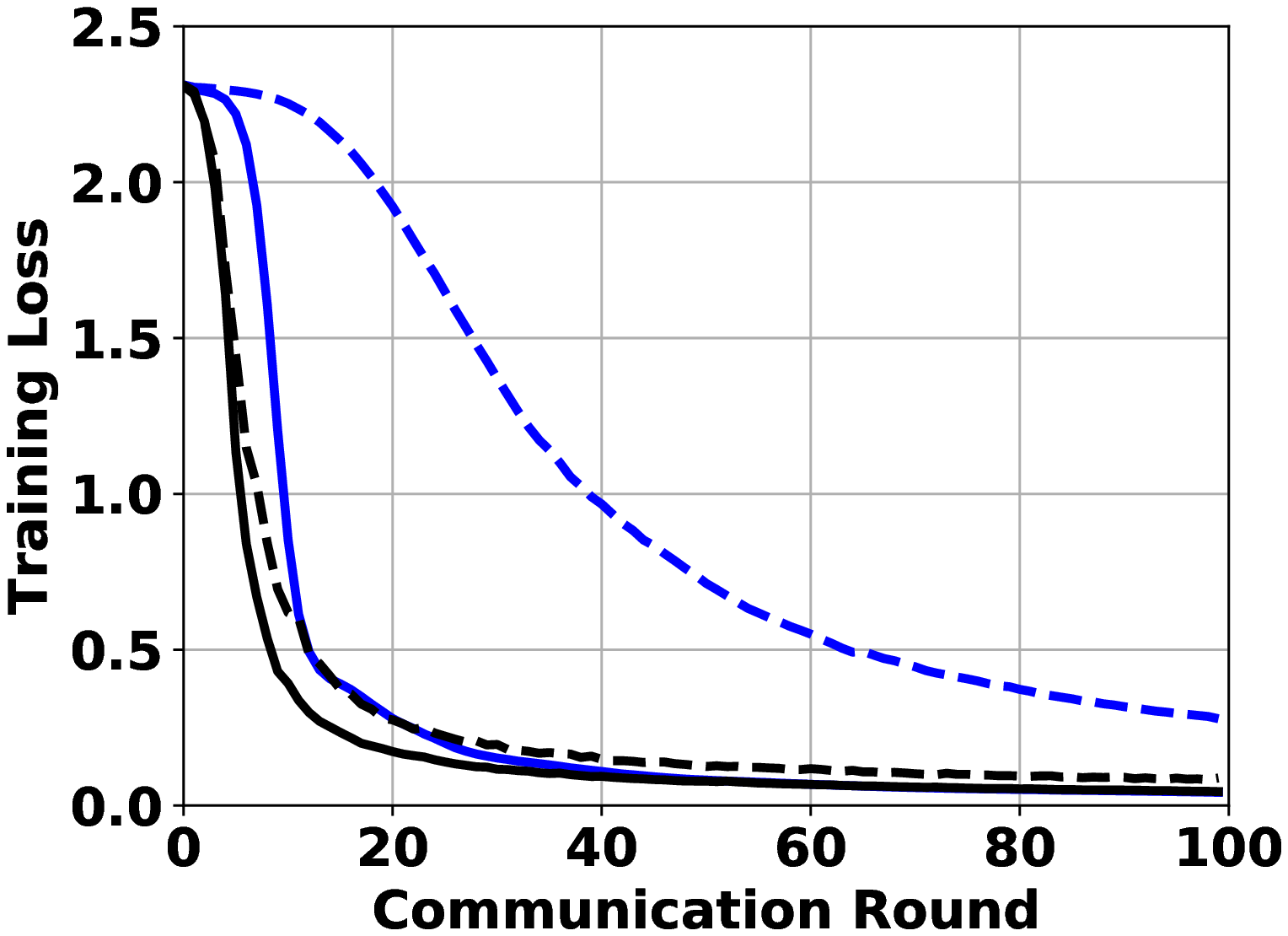}}
	\vspace{-.2in}
	\end{minipage}
	\hfill
	\begin{minipage}[t]{0.49\linewidth}
	\centering
	{\includegraphics[width=0.9\columnwidth]{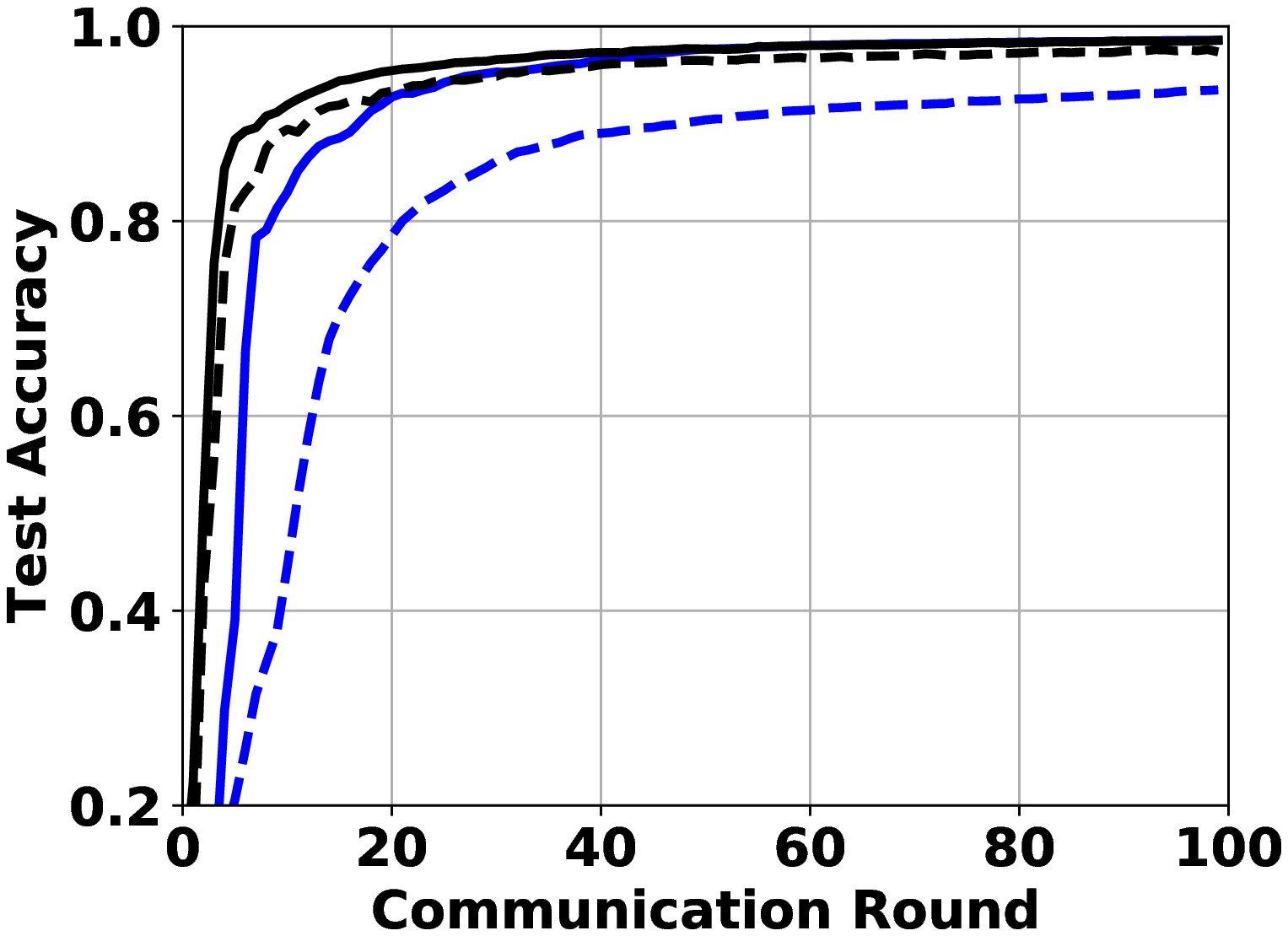}}
	\vspace{-.2in}
	\end{minipage}

	\begin{minipage}[t]{0.49\linewidth}
	\centering
	{\includegraphics[width=0.9\columnwidth]{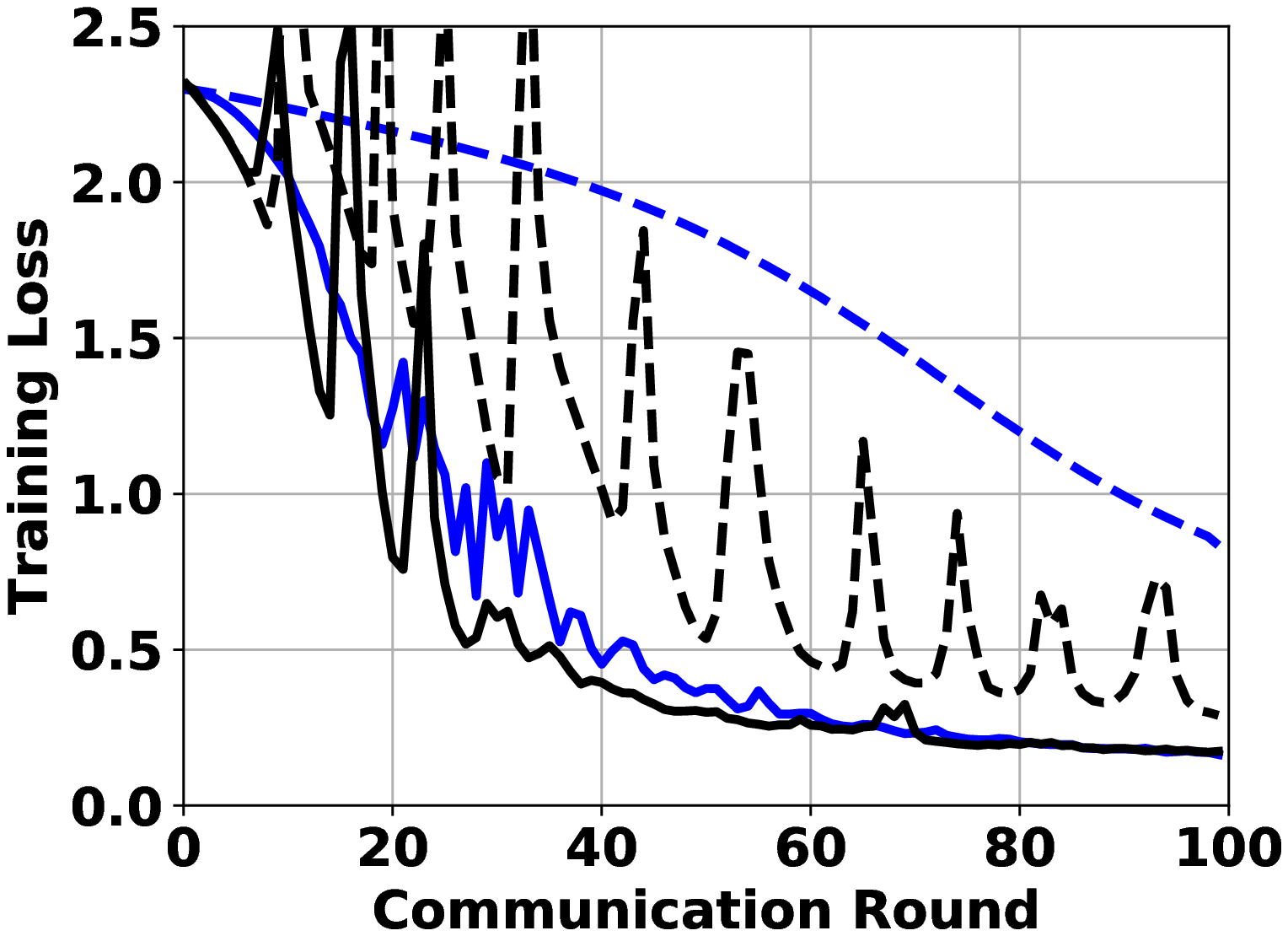}}
	\vspace{-.2in}
	\end{minipage}
	\begin{minipage}[t]{0.49\linewidth}
	\centering
	{\includegraphics[width=0.9\columnwidth]{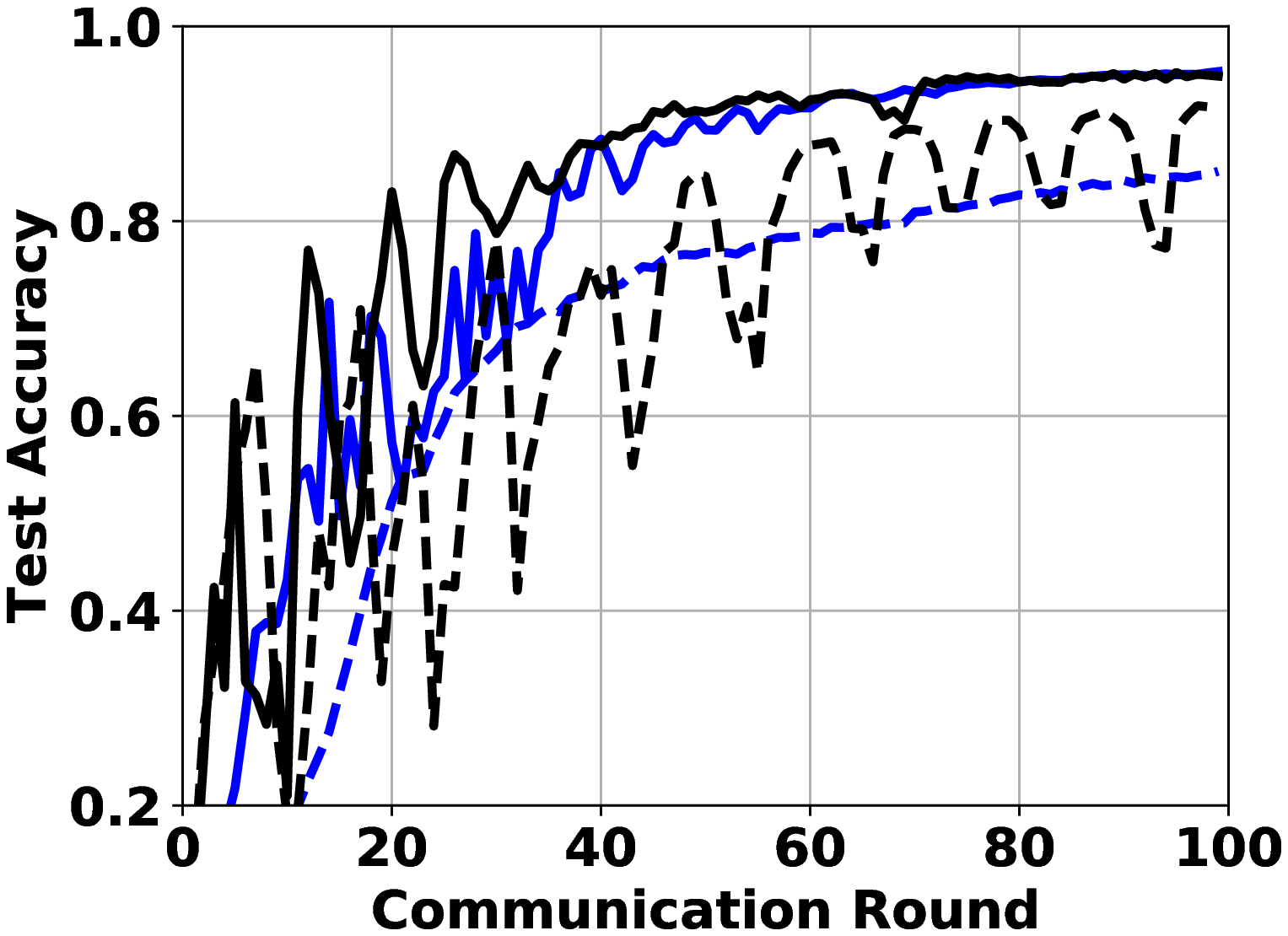}}
	\vspace{-.2in}
	\end{minipage}

\caption{Training loss (top) and test accuracy (bottom) for the CNN model for MNIST as comparison with respect to error feedback (EF). The non-i.i.d. levels are $p = 10, 5, 2, 1$ from top to bottom.}
\label{mnist_appdx_fig2}
\end{figure}%

\begin{figure}[t!]
	\begin{minipage}[t]{0.49\linewidth}
	\centering
	{\includegraphics[width=0.9\columnwidth]{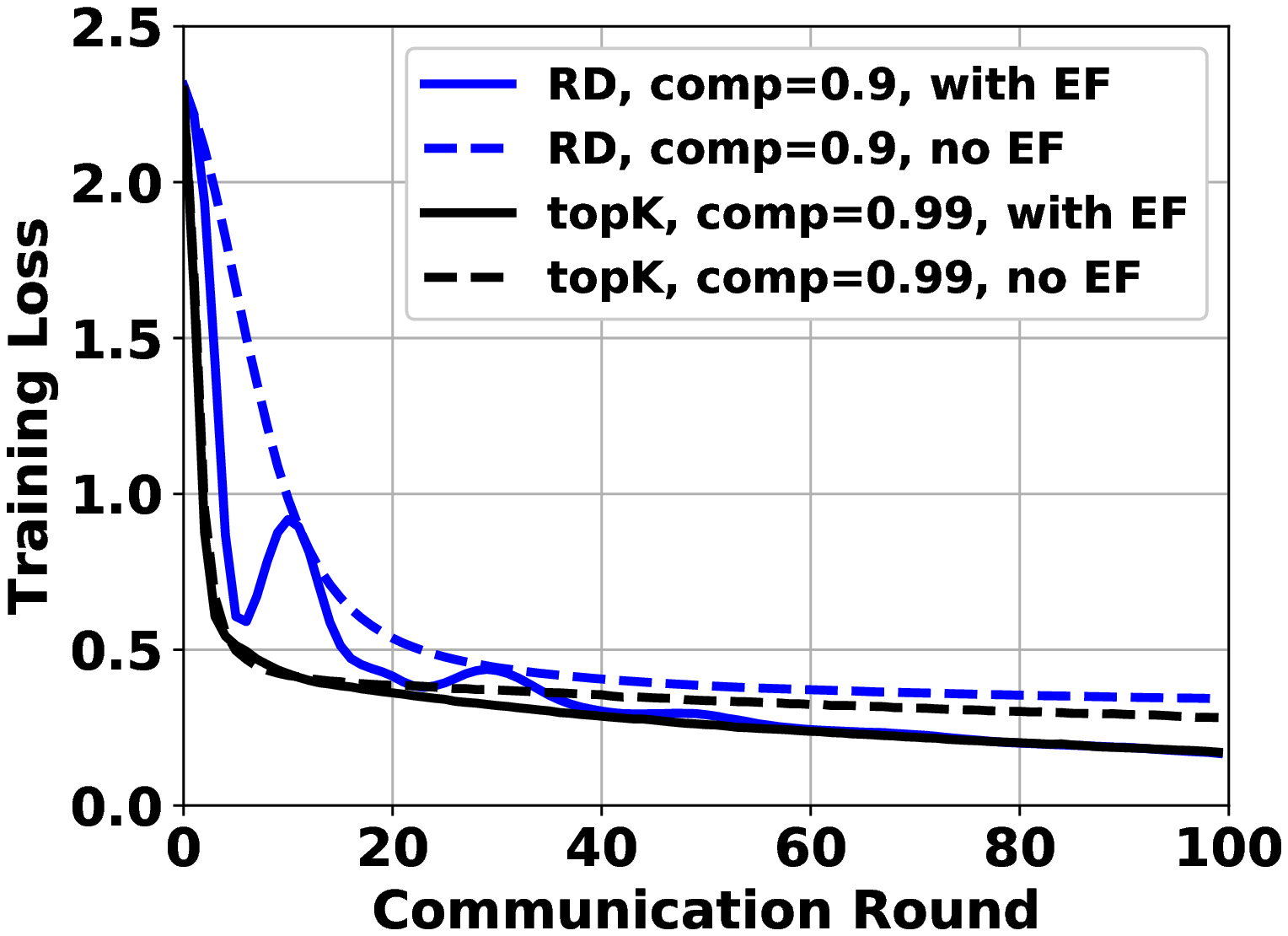}}
	\vspace{-.2in}
	\end{minipage}
	\hfill
	\begin{minipage}[t]{0.49\linewidth}
	\centering
	{\includegraphics[width=0.9\columnwidth]{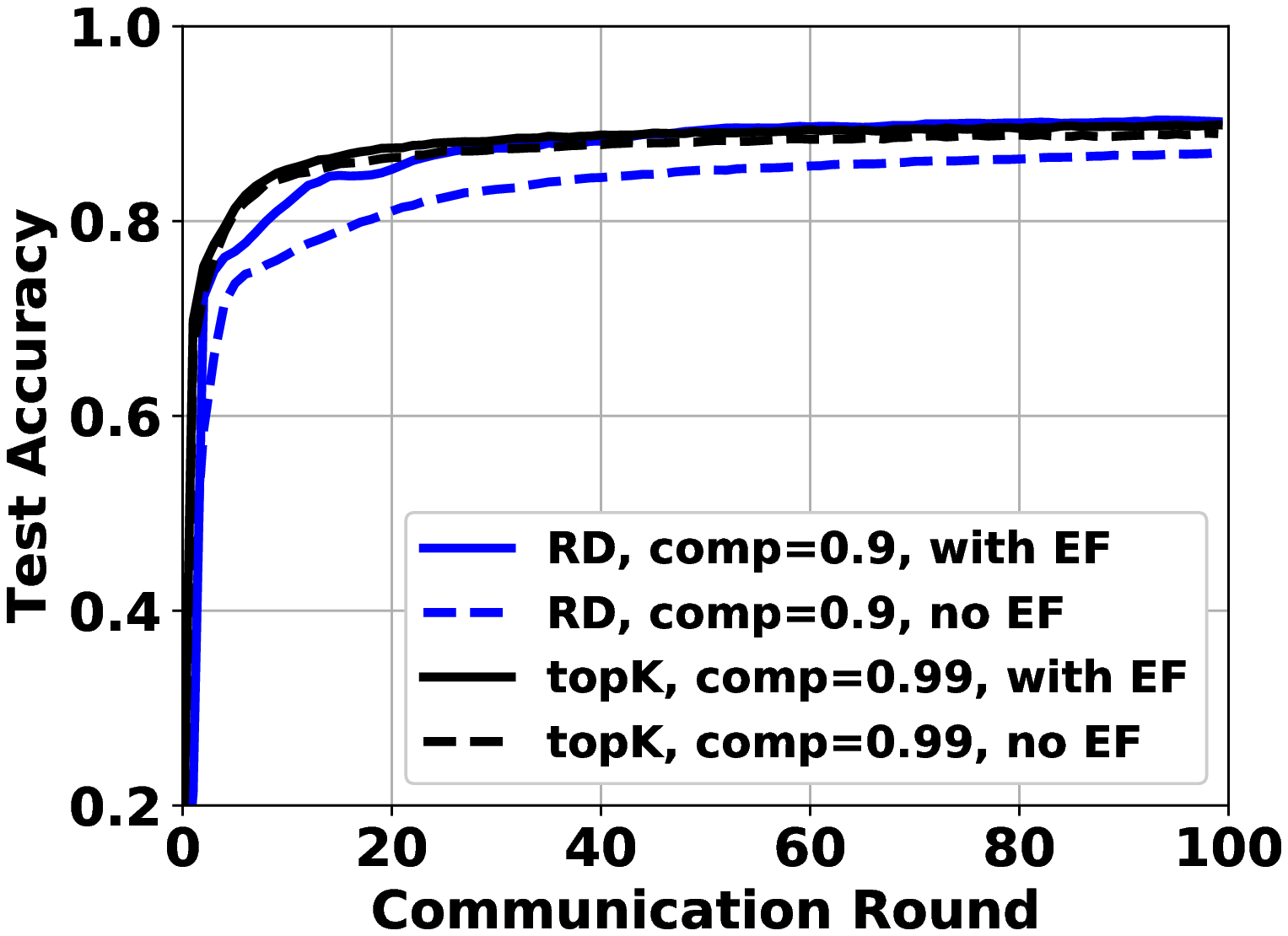}} 
	\vspace{-.2in}
	\end{minipage}

	\begin{minipage}[t]{0.49\linewidth}
	\centering
	{\includegraphics[width=0.9\columnwidth]{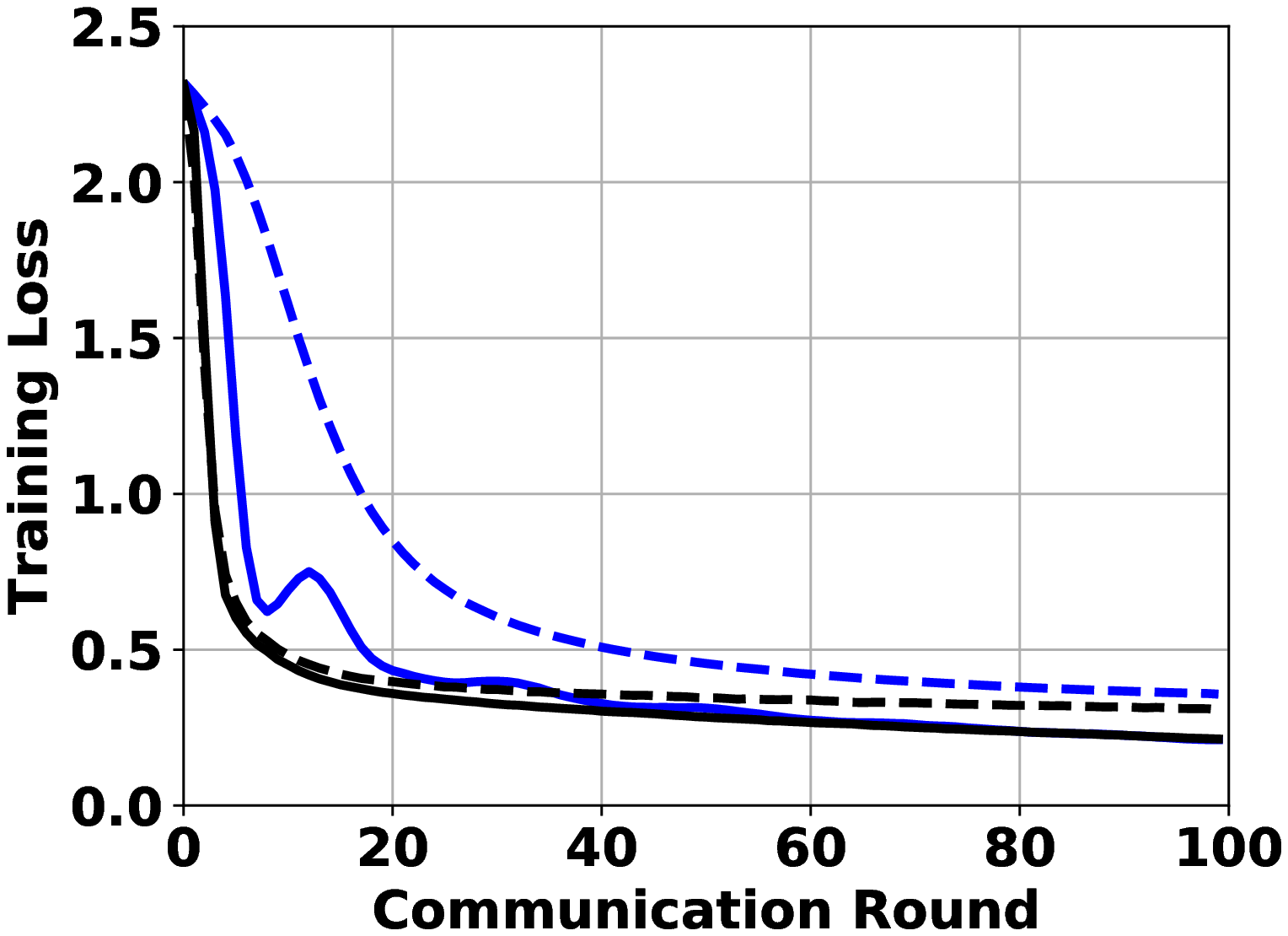}}
	\vspace{-.2in}
	\end{minipage}
	\hfill
	\begin{minipage}[t]{0.49\linewidth}
	\centering
	{\includegraphics[width=0.9\columnwidth]{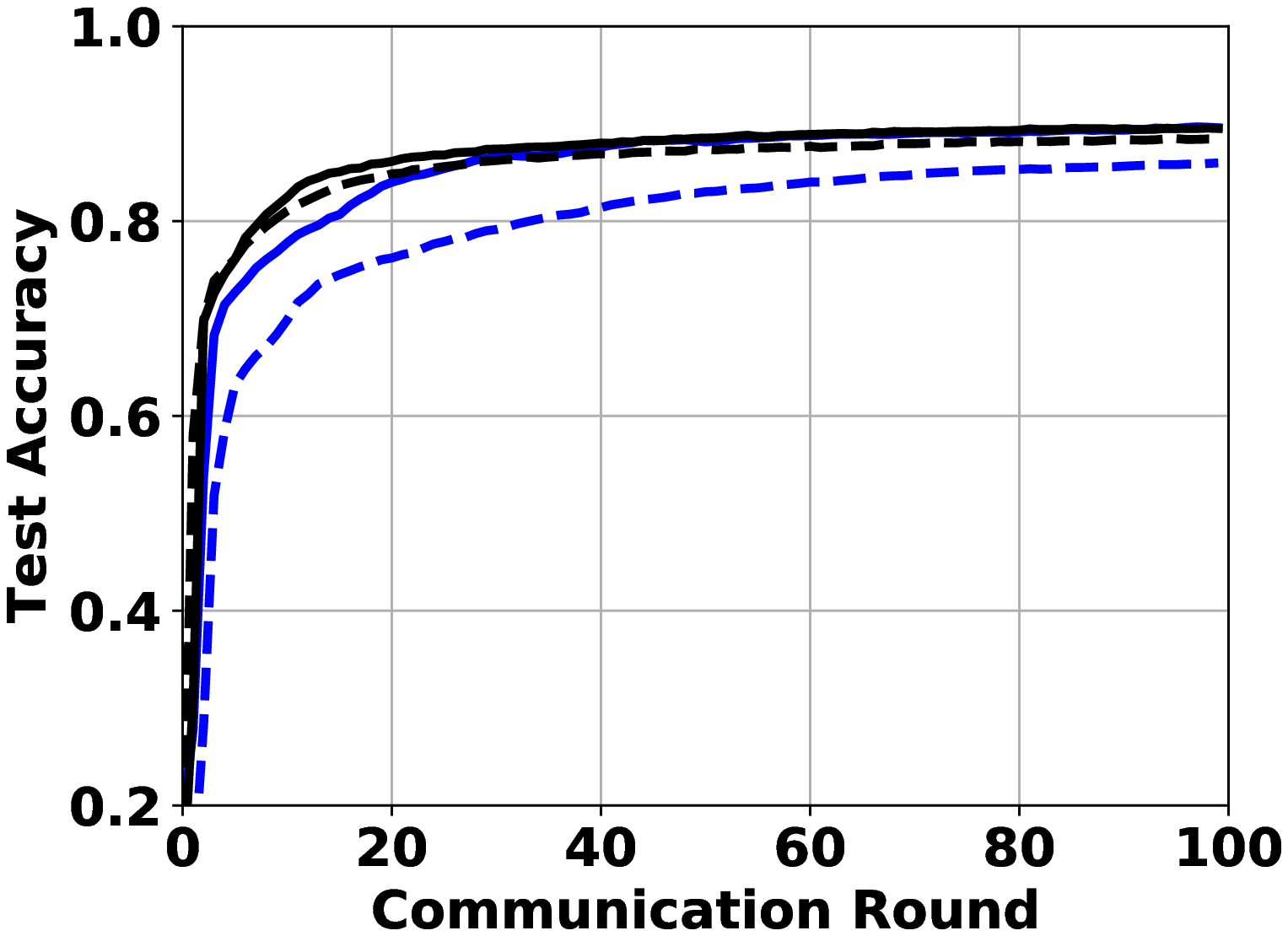}}
	\vspace{-.2in}
	\end{minipage}

	\begin{minipage}[t]{0.49\linewidth}
	\centering
	{\includegraphics[width=0.9\columnwidth]{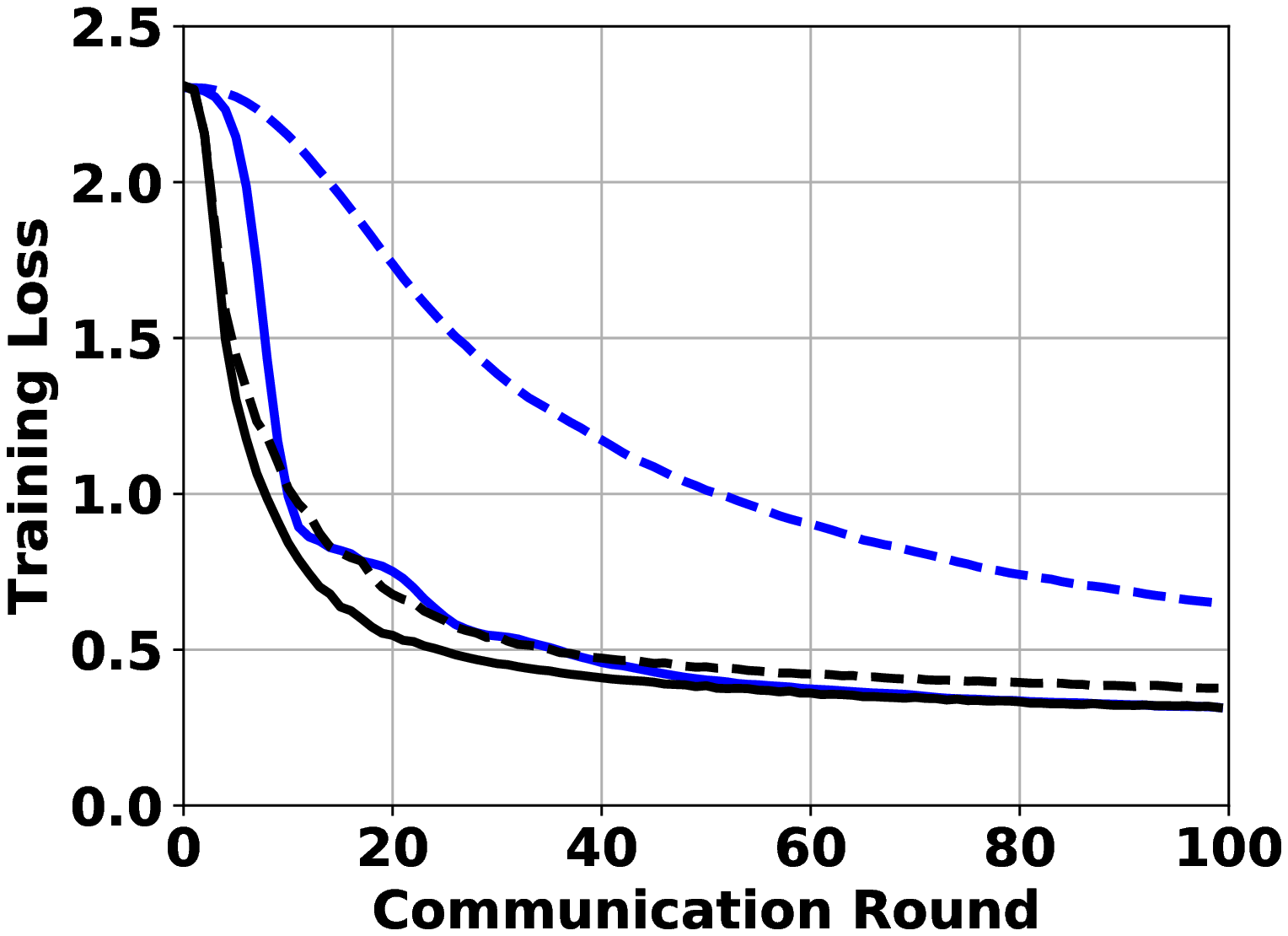}}
	\vspace{-.2in}
	\end{minipage}
	\hfill
	\begin{minipage}[t]{0.49\linewidth}
	\centering
	{\includegraphics[width=0.9\columnwidth]{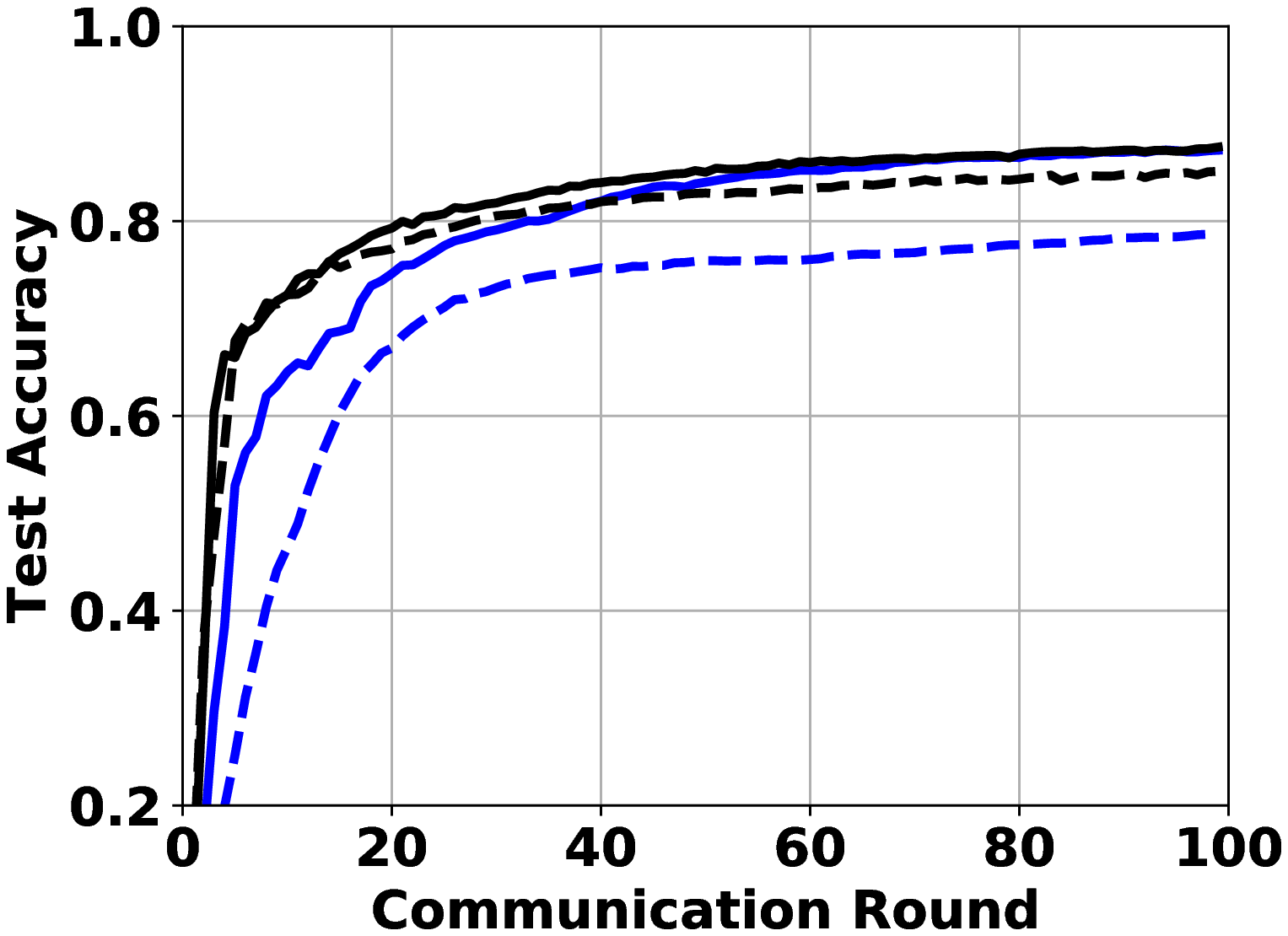}}
	\vspace{-.2in}
	\end{minipage}

	\begin{minipage}[t]{0.49\linewidth}
	\centering
	{\includegraphics[width=0.9\columnwidth]{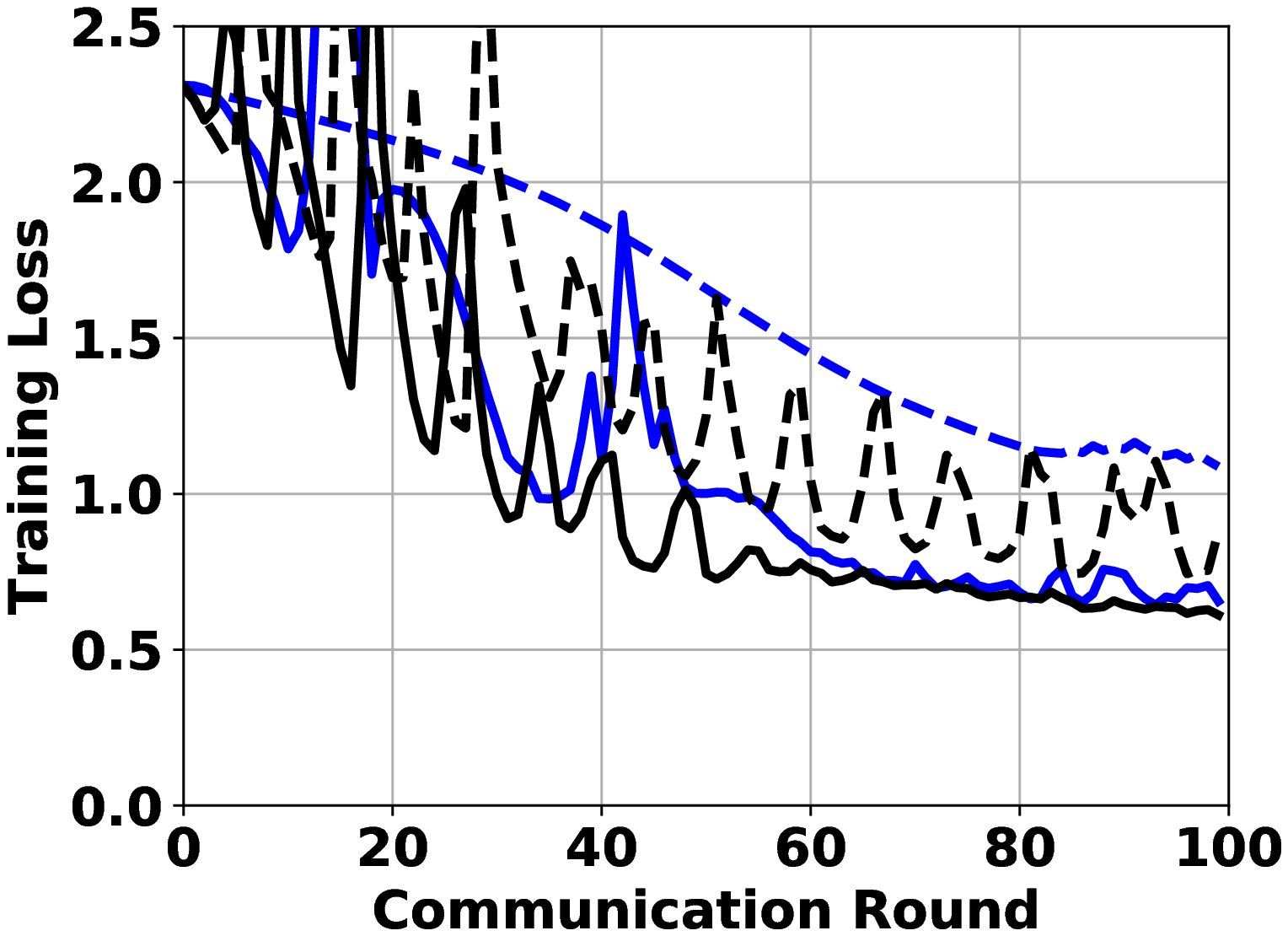}}
	\vspace{-.2in}
	\end{minipage}
	\begin{minipage}[t]{0.49\linewidth}
	\centering
	{\includegraphics[width=0.9\columnwidth]{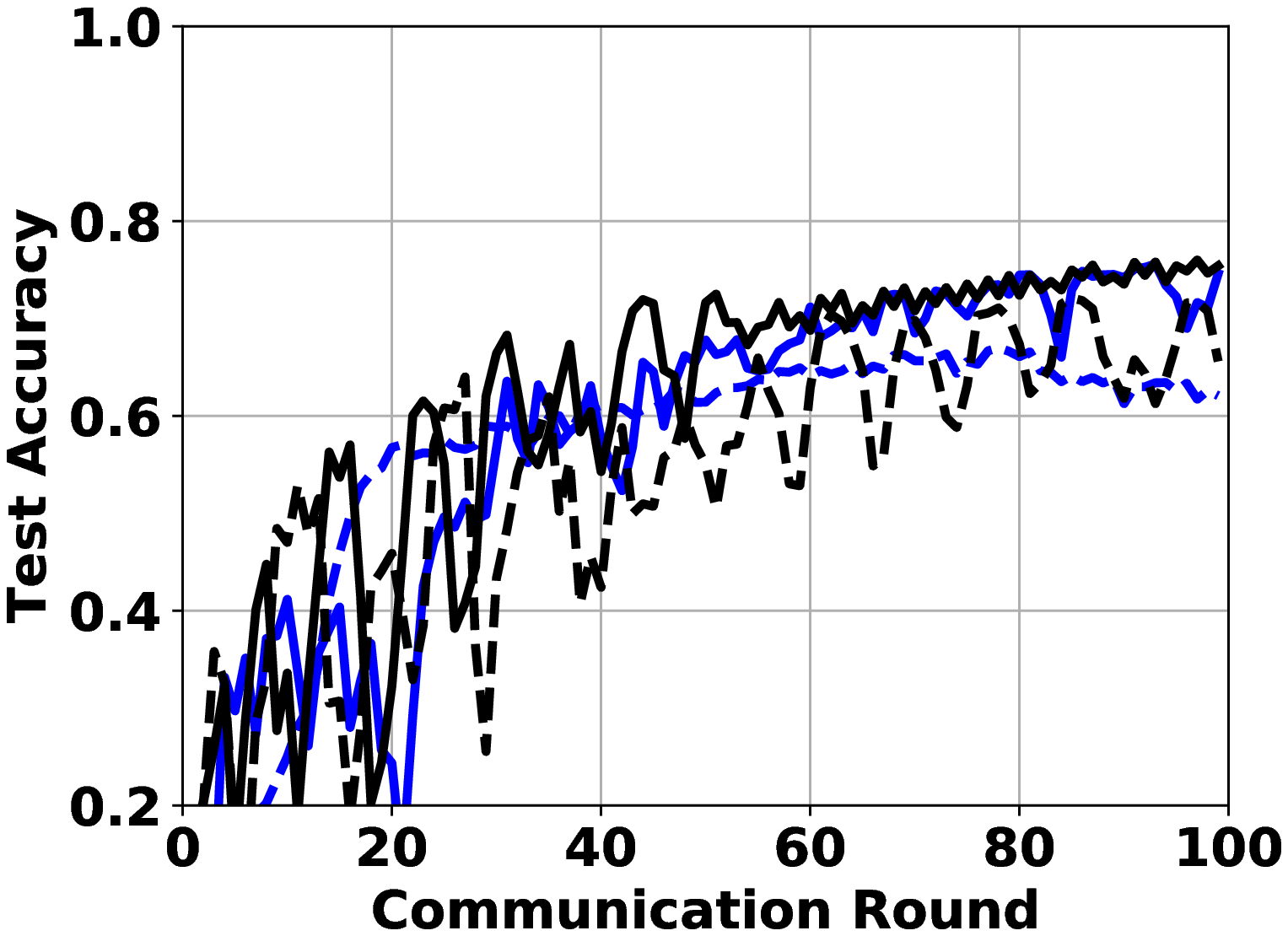}}
	\vspace{-.2in}
	\end{minipage}

\caption{Training loss (top) and test accuracy (bottom) for the CNN model for Fashion-MNIST as comparison with respect to error feedback (EF). The non-i.i.d. levels are $p = 10, 5, 2, 1$ from top to bottom.}
\label{appdx_fig2}
\end{figure}%
\subsection{Effectiveness of Gradient Compression:}
As shown in Figure~\ref{appdx_fig1} and ~\ref{appdx_cifar10_fig1} for CNN model and ResNet-18 on different non-i.i.d. Fashion-MNIST and CIFAR-10 datasets respectively, the top-row figures are for training loss versus communication round and the bottom-row are for test accuracy versus communication round.
Key observations are as follows:
\begin{list}{\labelitemi}{\leftmargin=1em \itemindent=0em \itemsep=.2em}
\item 
We can see that our algorithm with two gradient compression methods converges at different levels of non-i.i.d.-ness from left ($p=10$) to right ($p=1$).
In general, these cases preserving more information in each communication round usually have better results.
It is shown that top-k precedes random dropping with same compression rate $comp$ and that with low compression rate $comp$ for the same compression method have faster convergence speed.
\item
The training curve is twisted for highly non-i.i.d. case (e.g., $p=1$) compared to i.i.d. case.
As the non-i.i.d.-ness increases, this zigzagging curve is more obvious as shown in the figure, which we believe is a instinct feature of non-i.i.d. dataset in FL.
However, the learning curves are more smoothing with gradient compression and error feedback.
For example, for RD with $comp=0.9$ and top-K with $comp=0.99$, a better convergence curve than that of the original one without any compression is shown in Figure~\ref{appdx_fig1}(d) and ~\ref{appdx_cifar10_fig1} (d) ($p=1$).
The intuition is that gradient compression method might filter some noises that leads to the instability of the learning curve due to model asynchrony among workers originated from the non-i.i.d. datasets and local steps.
We believe it is worthing exploring further.
\item
As discussed in the paper, our algorithm is compatible with any gradient compression method in theory.
However, it does not mean that any gradient compression method with any degree of compression is feasible in practice.
Intuitively, when losing too much information, it is impossible to train a valid model.
There seems a unsuccessful training for random dropping with $comp=0.99$ in $p=1$ case.
So in practice, the trade-off between communication cost and convergence of the model always exists.
\end{list}

\subsection{Importance of Error Feedback:}
\begin{figure}[ht!]
	\begin{minipage}[t]{0.49\linewidth}
	\centering
	{\includegraphics[width=0.9\textwidth]{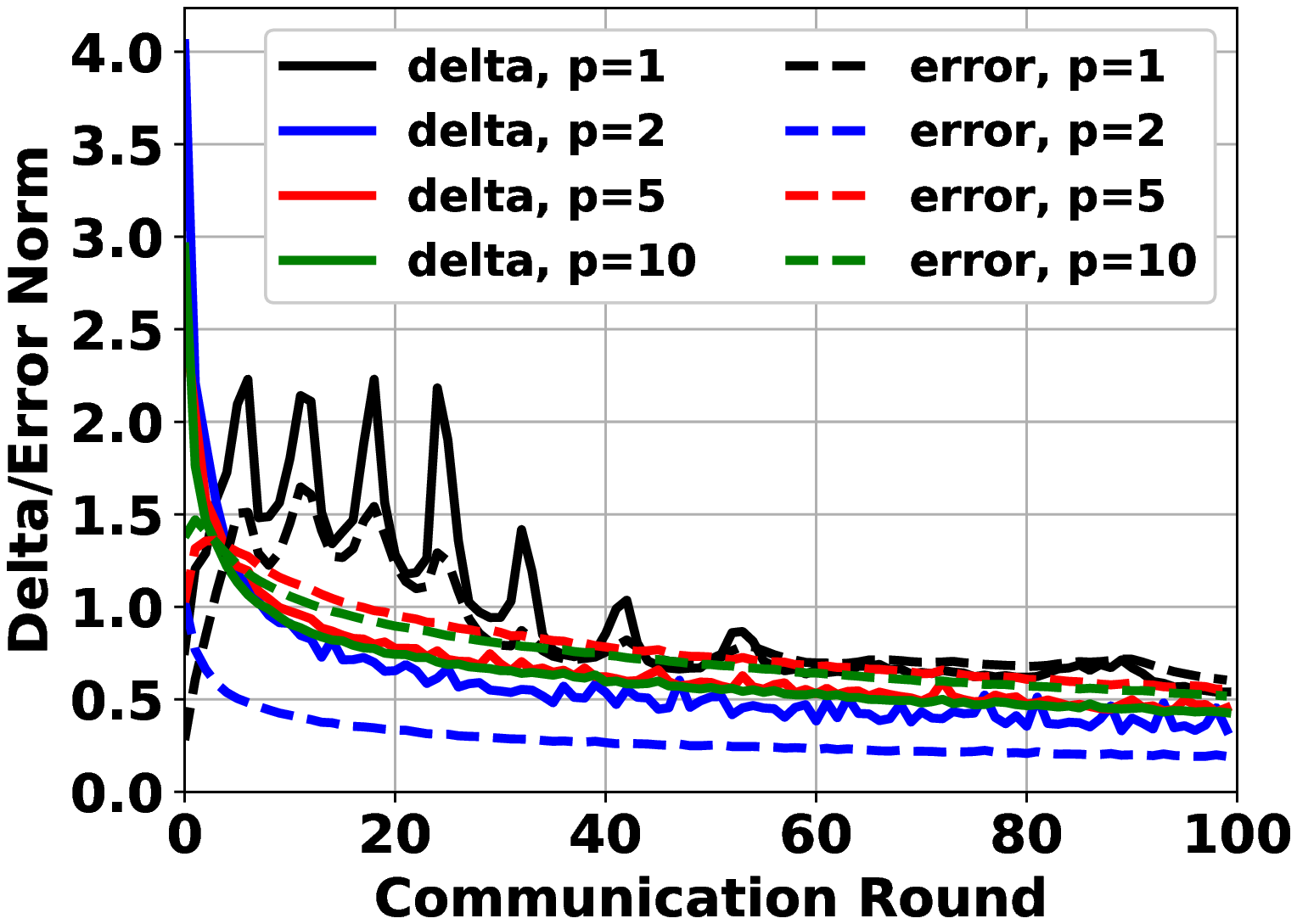}}
	\text{ (a) Top-K $comp=0.9$}
	\end{minipage}
	\begin{minipage}[t]{0.49\linewidth}
	\centering
	{\includegraphics[width=0.9\textwidth]{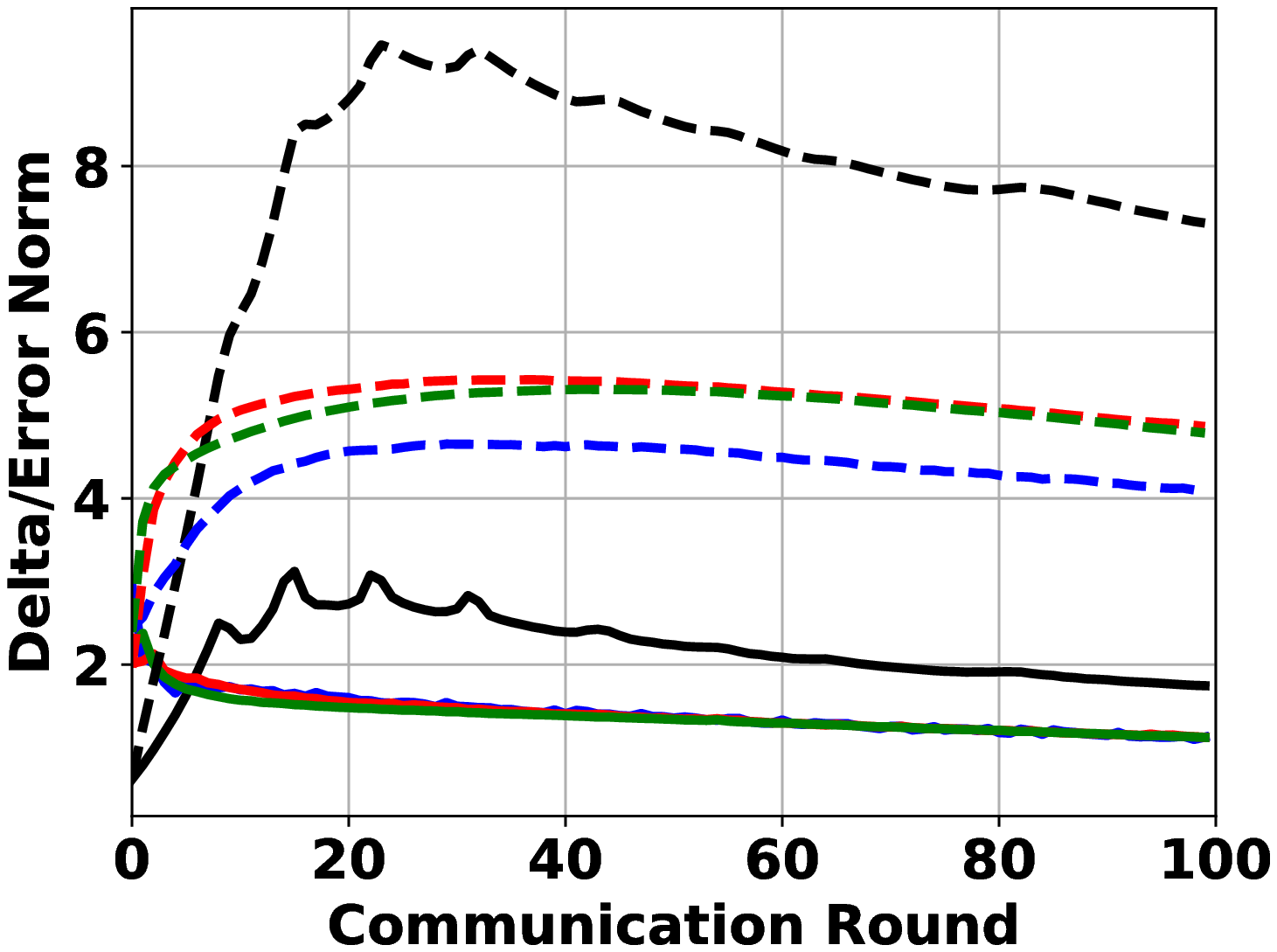}}
	\text{ (b) Top-K $comp=0.99$}
	\end{minipage}
	\begin{minipage}[t]{0.49\linewidth}
	\centering
	{\includegraphics[width=0.9\textwidth]{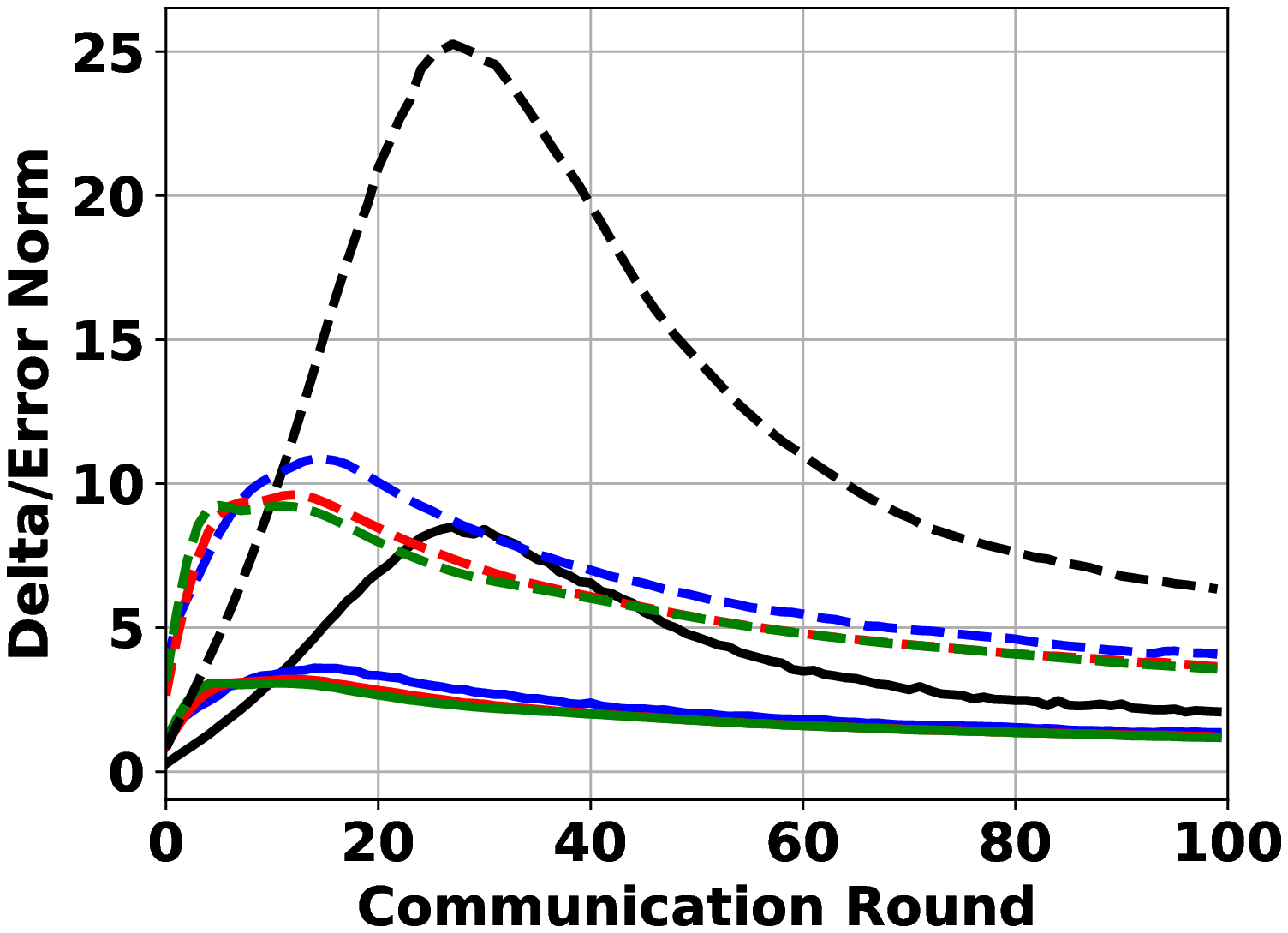}} 
	\text{ (c) RD $comp=0.9$}
	\end{minipage}
	\begin{minipage}[t]{0.49\linewidth}
	\centering
	{\includegraphics[width=0.9\textwidth]{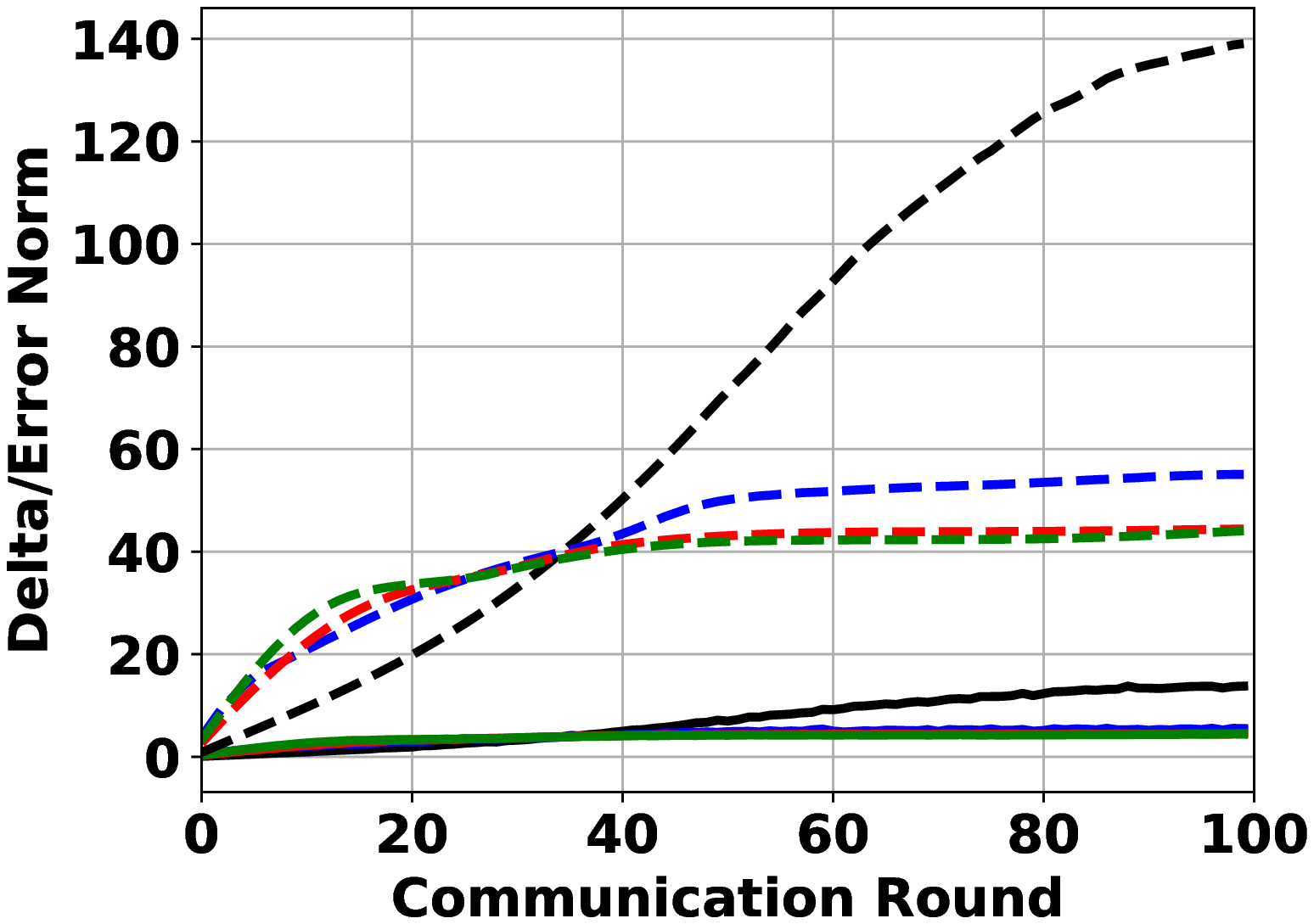}}
	\text{ (d) RD $comp=0.99$}
	\end{minipage}

\caption{Mean of the norms of $\frac{1}{m} \sum_{i=1}^{100} \|\Delta_t^i \|^2$ and the error term $\frac{1}{m} \sum_{i=1}^{100} \|\e_t^i \|^2$ for the CNN model on MNIST.}
\label{mnist_appdx_fig3}
\end{figure}

\begin{figure}[ht!]
	\begin{minipage}[t]{0.49\linewidth}
	\centering
	{\includegraphics[width=0.9\textwidth]{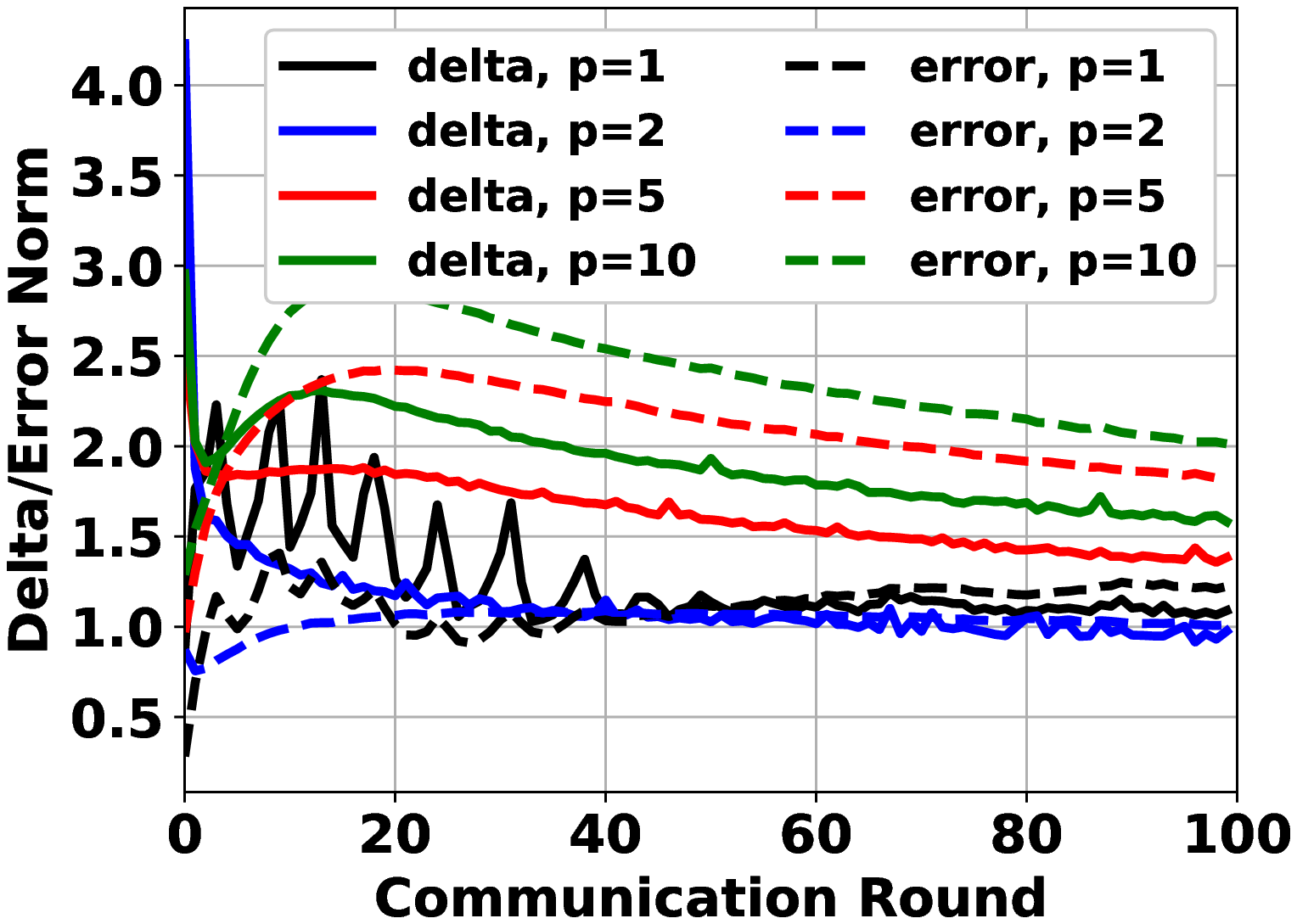}}
	\text{ (a) Top-K $comp=0.9$}
	\end{minipage}
	\begin{minipage}[t]{0.49\linewidth}
	\centering
	{\includegraphics[width=0.9\textwidth]{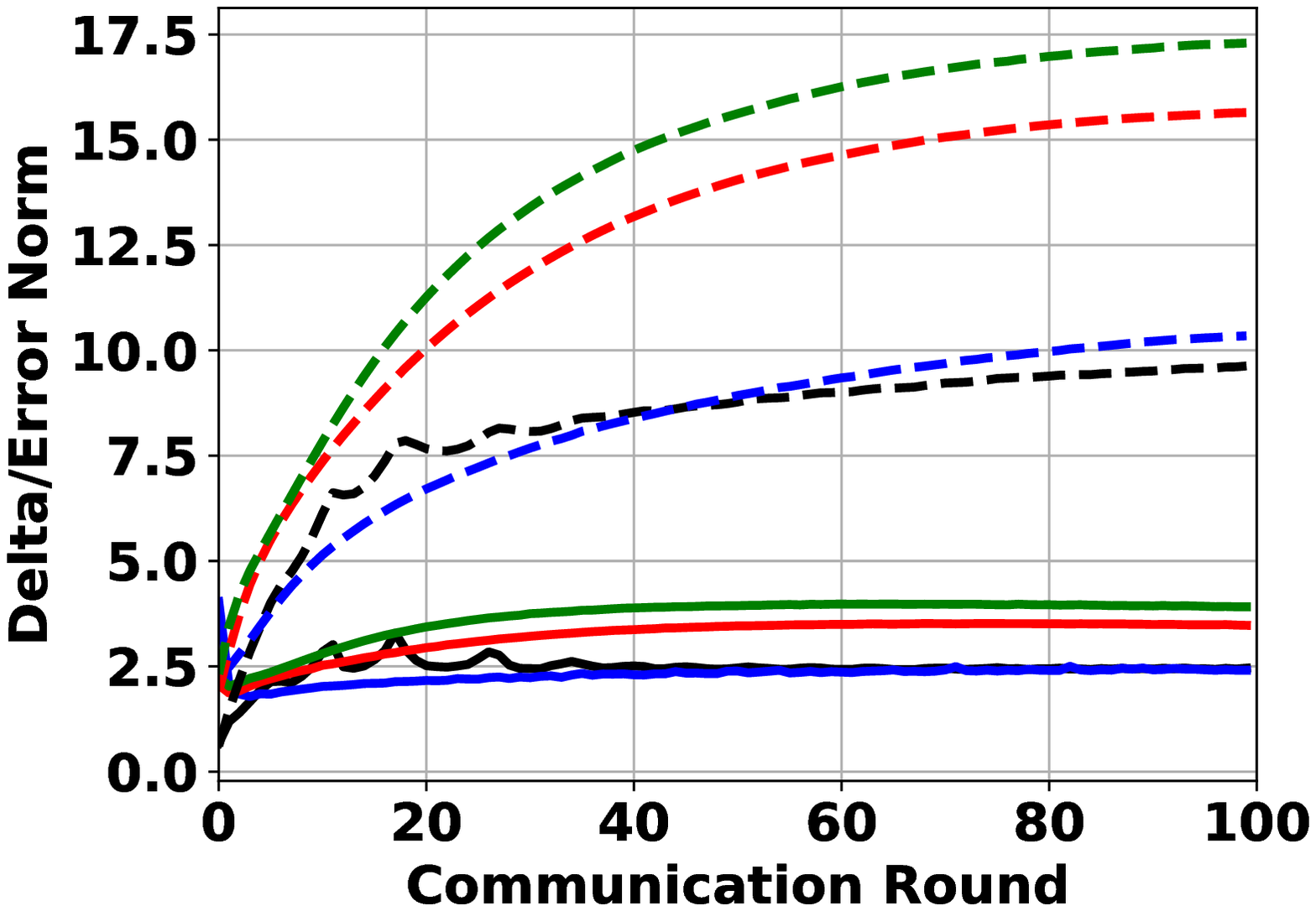}}
	\text{ (b) Top-K $comp=0.99$}
	\end{minipage}
	\begin{minipage}[t]{0.49\linewidth}
	\centering
	{\includegraphics[width=0.9\textwidth]{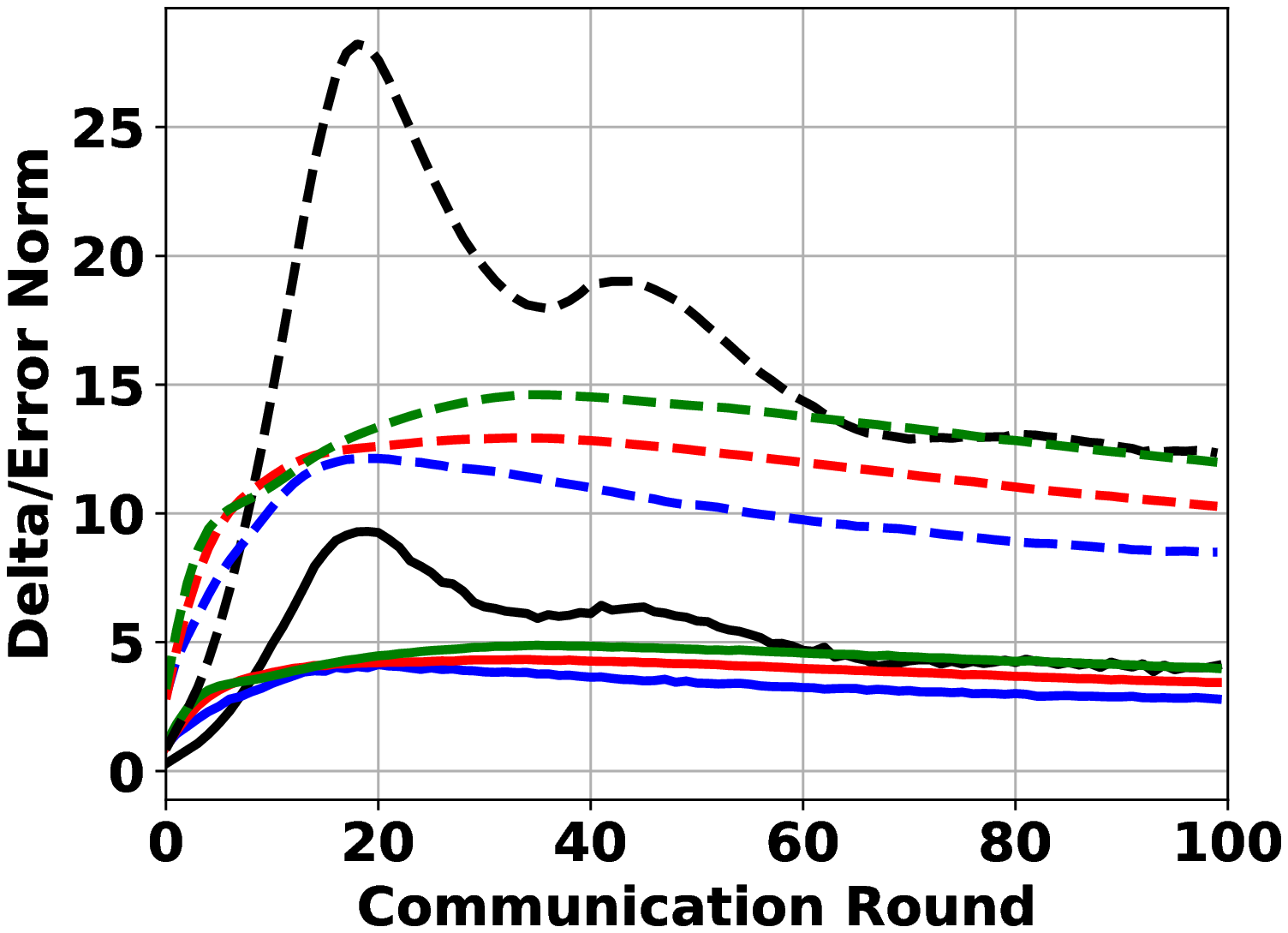}} 
	\text{ (c) RD $comp=0.9$}
	\end{minipage}
	\begin{minipage}[t]{0.49\linewidth}
	\centering
	{\includegraphics[width=0.9\textwidth]{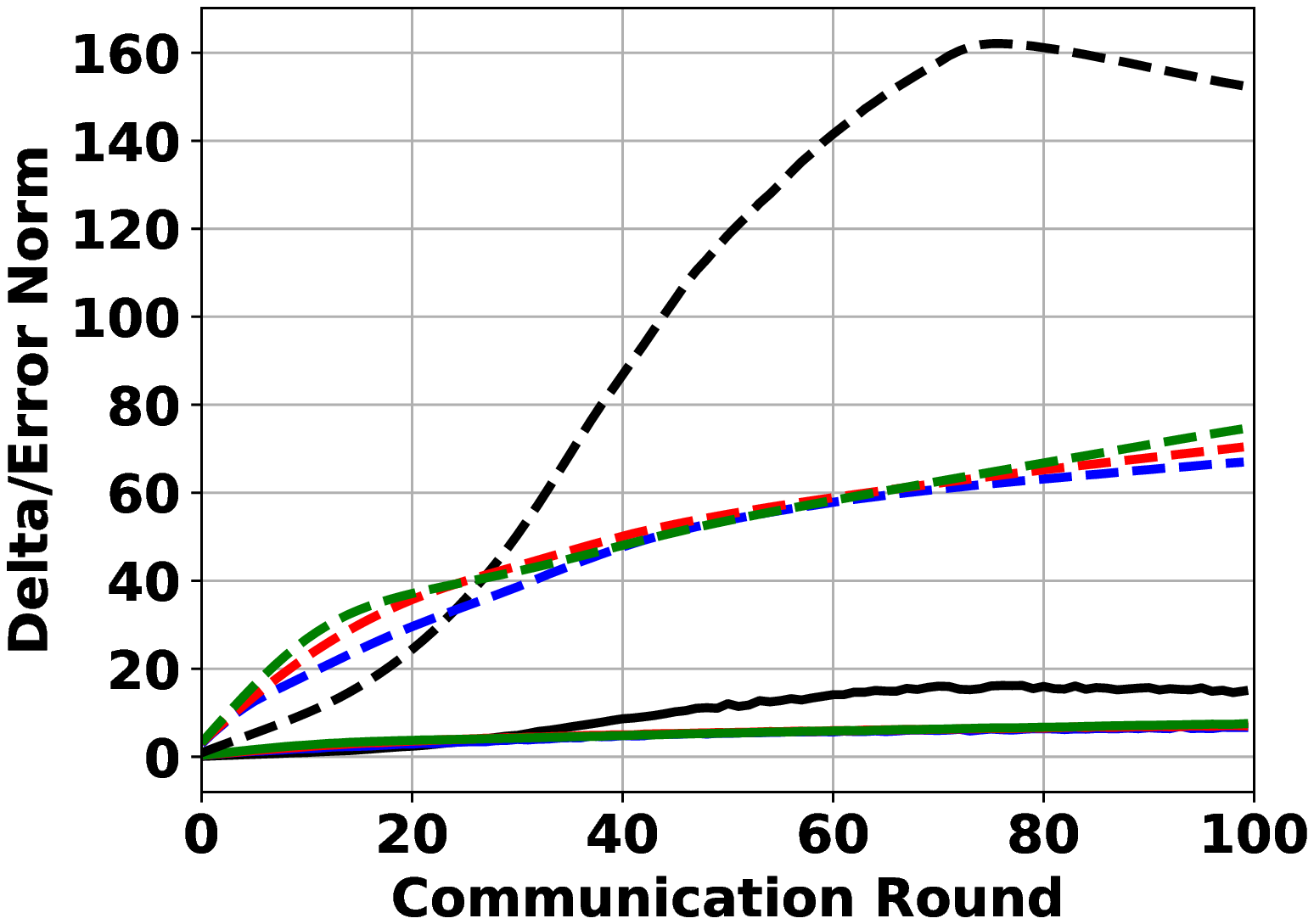}}
	\text{ (d) RD $comp=0.99$}
	\end{minipage}

\caption{Mean of the norms of $\frac{1}{m} \sum_{i=1}^{100} \|\Delta_t^i \|^2$ and the error term $\frac{1}{m} \sum_{i=1}^{100} \|\e_t^i \|^2$ for the CNN model on Fashion-MNIST.}
\label{appdx_fig3}
\end{figure}

As shown in Figure~\ref{mnist_appdx_fig2} and ~\ref{appdx_fig2}, error feedback helps the training in general.
If naively applying gradient compression, huge amount of information will be lost, thus leading to relatively poor performance. 
With error feedback, the error term accumulates the information that is not transmitted to the server in the current communication round and then compensates the gradients in the next communication round.
This is verified in Figure~\ref{mnist_appdx_fig3} and ~\ref{appdx_fig3}, which shows the mean of the norm of gradients changes $\frac{1}{m} \sum_{i=1}^{m} \|\triangle_t^i \|$ and the error term $\frac{1}{m} \sum_{i=1}^{m} \|\e_t^i \|$ for the total workers $m=100$.
One key observation is that the error term is usually several times of the gradients changes term in general.
Under the same condition, the error term is much larger in high non-i.i.d. case.
Another observation is that the error term is bounded under proper gradient compression methods and compression rate, which confirms our theoretical analysis.
However, with too ambitious compression, the error term continues growing as shown in Figure~\ref{mnist_appdx_fig3}(b) and ~\ref{appdx_fig3}(d).
This is in accordance with the analysis in Figure~\ref{appdx_fig1} that learning curve is unstable.
So in general, the error feedback could guarantee that not too much information is dropped due to the compression operator albeit with some delay.

\end{document}